\documentclass[10pt,twocolumn,letterpaper]{article}

\usepackage{cvpr}              %

\usepackage{array}
\usepackage[table]{xcolor}
\usepackage{graphicx}
\usepackage{amsmath}
\usepackage{amssymb}
\usepackage{booktabs}
\usepackage{float}
\usepackage{comment}
\usepackage[accsupp]{axessibility}
\usepackage{tikz}

\usepackage{tabularx}
\usepackage{multirow}
\usepackage{adjustbox}
\usepackage{makecell}
\usepackage{environ}
\usepackage{subcaption} %
\usepackage{placeins}
\usepackage{pifont}%

\newcolumntype{Y}{>{\centering\arraybackslash}X}

\newif\ifisdraft
\isdraftfalse
\isdrafttrue
\NewEnviron{drafty}{%
    \ifisdraft
        {\color{gray} \BODY}
    \fi
}

\NewEnviron{TODO}{%
    \ifisdraft
        {\color{red} \BODY}
    \fi
}

\newcommand{\jannik}[1]{
    \ifisdraft
        \textbf{{\color{orange}[Jannik: #1]}}
    \fi
}
\newcommand{\joh}[1]{
    \ifisdraft
        \textbf{{\color{blue}[Johannes: #1]}}
    \fi
}

\newcommand{\ven}[1]{\tiny\color{gray}(#1)}

\makeatletter
\renewcommand{\paragraph}{%
  \@startsection{paragraph}{4}{\z@}%
  {0.3em} %
  {-1em}  %
  {\normalfont\normalsize\bfseries}%
}
\makeatother

\definecolor{cvprblue}{rgb}{0.21,0.49,0.74}
\usepackage[pagebackref,breaklinks,colorlinks,allcolors=cvprblue]{hyperref}

\def\paperID{44123} %
\def\confName{CVPR}
\def\confYear{2026}

\title{Probabilistic Precipitation Nowcasting with Rectified Flow Transformers}

\author{
    Johannes Schusterbauer\thanks{Equal Contribution.} \qquad Jannik Wiese\footnotemark[1] \qquad Nick Stracke \qquad
    Timy Phan \qquad Bj\"orn Ommer\\[0.8em]
    CompVis @ LMU Munich \qquad Munich Center for Machine Learning (MCML)
}

\begin{document}
\maketitle

\begin{abstract}
Accurate weather forecasts are essential across various domains and are safety-critical in extreme weather conditions.
Compared to simulation-based forecasting, data-driven approaches show greater efficiency, enabling short-term, high-resolution nowcasting.
In particular, diffusion models proved effective in weather nowcasting due to their strong probabilistic foundation.
However, existing methods rely on deterministic compression to reduce the complexity of high-dimensional weather data, limiting their ability to capture uncertainty in the decoding process.
In this work, we introduce \textit{FREUD}, a \textbf{Fr}ame-wise \textbf{E}ncoder and \textbf{U}nited \textbf{D}ecoder model based on rectified flow transformers for efficient compression of spatio-temporal weather data. Frame-wise encoding enables continuous forecast updates, while the unified video decoder ensures temporal consistency. Our uncertainty-preserving first stage allows us to capture aleatoric uncertainty via ensembling, which is particularly beneficial for extreme weather events with high decoding variability. We achieve state-of-the-art performance in precipitation nowcasting with a compact latent-space rectified flow transformer on the SEVIR benchmark and show further performance gains by model and test-time scaling. Code available here: \small \url{https://github.com/CompVis/weather-rf}.

\end{abstract}

\begin{figure}
    \centering
    \definecolor{normal_color}{HTML}{131285}
    \definecolor{medium_color}{HTML}{ec6605}
    \definecolor{high_color}{HTML}{e10612}
    \includegraphics[width=0.75\linewidth]{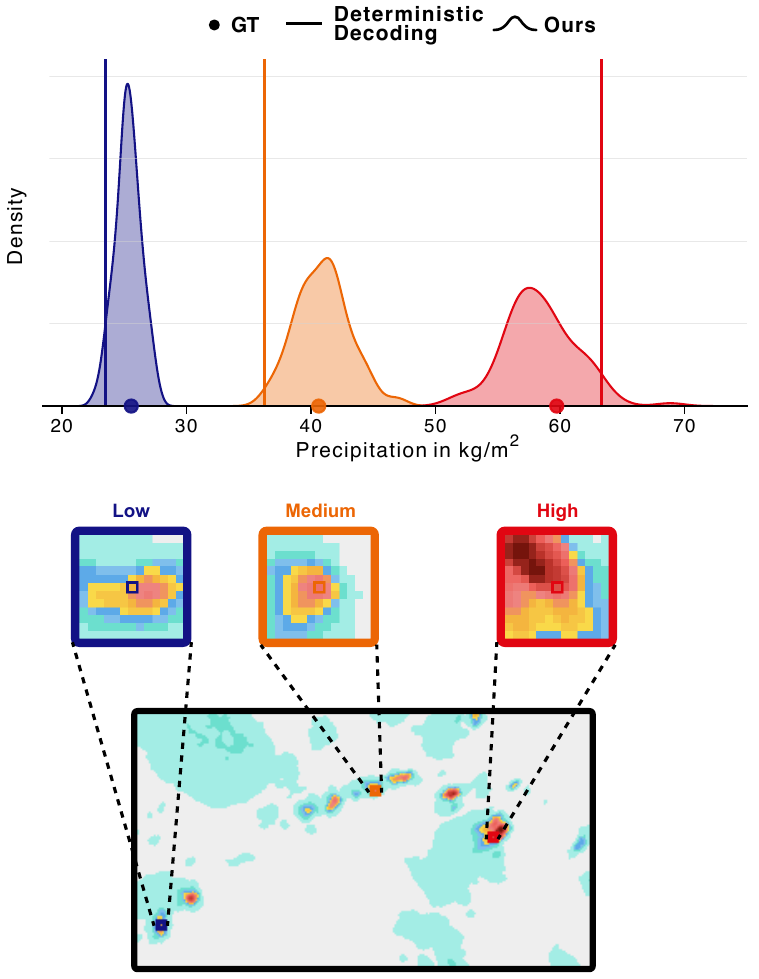}
    \caption{
        \textbf{Reconstruction ensemble distributions} from our generative compression stage for locations with medium ({\color{normal_color} blue}), heavy ({\color{medium_color} orange}), and extreme ({\color{high_color} red}) precipitation.
        In light-rain regions, the ensemble shows low variance, while under intense and chaotic rainfall, the spread increases and reliably covers the ground-truth values (dots),
        whereas the deterministic CasCast decoder~\cite{gong2024cascast} (lines) cannot quantify uncertainty and fails to cover ground truth.
    }
    \label{fig:front_page}
\end{figure}

\section{Introduction}
\label{sec:intro}
Accurate weather forecasting underpins many industries, including energy production~\cite{sperati_application_2016, wu2018Probabilistic, wang2017Deep}, agriculture~\cite{shen2024Analysis, raimundo2021Prediction, collins2024Evaluating}, and aviation~\cite{Enea2024Analysis, nunez-portillo2024Predicting, ramli2014Practical} and is safety-critical during extreme weather events.
Nowcasting refers to short-term, high-resolution forecasts over 30 minutes to 12 hours~\cite{trebing21:SmaAt-UNet, fernandez21:Broad-UNet, espeholt22:MetNet2} and requires rapid updates, rendering slow numerical weather prediction models less useful.
Data-driven machine learning (ML) models overcome this speed bottleneck~\cite{trebing21:SmaAt-UNet, agrawal19:UNet}, but deterministic formulations often blur predictions by averaging over multiple plausible futures~\cite{trebing21:SmaAt-UNet, fernandez21:Broad-UNet, espeholt22:MetNet2}.
Sampling from probabilistic generative models yields sharp, diverse forecasts, whose spread naturally captures predictive uncertainty~\cite{ravuri21:GAN, chan2024HyperDiffusion, shu2024ZeroShot, leinonen2023ldcast, gao2023prediff}.
Uncertainty quantification is of particular importance in safety-critical settings where the range of plausible outcomes matters more than the most likely one.
Early generative approaches use Generative Adversarial Networks (GANs)~\cite{goodfellow2014gan, ravuri21:GAN, liu2020MPLGAN, zhang2023skilful}.
However, GANs suffer from training instabilities and mode collapse~\cite{kossale2022Mode}, limiting coverage of rare but critical long-tail events.
Hence, effective nowcasting requires \emph{data-driven methods} that are \emph{accurate}, \emph{probabilistic}, \emph{fast}, and \emph{scalable}.

Diffusion and flow-based models~\cite{song2019scorebased, song2020ddim, ho2020ddpm, rombach2022ldm, liu2022Flow, albergo2022stochint, albergo2023stochastic, lipman2023flowmatching} overcome GAN instabilities and mode collapse, but iterative sampling in high-dimensional spaces is computationally expensive.
Commonly, a two-stage setup is employed to reduce overhead.
First, a compression stage abstracts imperceptible details. Second, a generative model operates in the learned latent space.
While effective for natural images and videos, the current \emph{two-stage design is ill-posed for weather forecasting}.
Compression is inherently lossy, and errors that are imperceptible in images may correspond to substantial shifts in predicted precipitation. Two-stage modeling thus undermines reliability in safety-critical settings.
Moreover, training such autoencoders requires balancing regularization~\cite{kingma2022AutoEncoding}, perceptual~\cite{zhang2018lpips}, and adversarial losses~\cite{goodfellow2014gan, rombach2022ldm}, where the adversarial components can reintroduce instability and degrade coverage of extreme events.
Despite these drawbacks, recent nowcasting models~\cite{gao2023prediff, gong2024cascast, leinonen2023ldcast} continue to rely on this architecture to manage the high dimensionality of weather data. %

In this work, we introduce a novel and simple first stage design, tailored to weather nowcasting.
Replacing the standard decoder with a rectified flow decoder ensures that predictions are sharp and realistic, without adversarial or perceptual losses. Therefore, the compression stage becomes (1) easy to train, (2) scalable, and (3) able to quantify uncertainty in the decoding process.
To ensure a smooth and well-behaved latent space, we introduce a novel stochastic $\tanh$ regularization, which restricts latents to the range $[-1, 1]$. Unlike KL regularization~\cite{kingma2022AutoEncoding}, it requires no loss weighting or architectural changes~\cite{rombach2022ldm, kingma2022AutoEncoding}.
We further propose an asymmetric architecture for our first stage, tailored for forecasting.
In line with recent image synthesis works~\cite{gui2025reptok, zheng2025rea}, we utilize transformers for the first-stage encoder and decoder, thereby avoiding the inefficient convolutional design~\cite{zheng2025rea}.
For latent-space forecasting, we must avoid \textit{information leakage from future to past} frames, making sequence-level encoders~\cite{wang2024OmniTokenizer, tang2024VidTok, li2025WFVAE, zhang25:REGEN, yang2025Rethinking} unsuitable. At the same time, we require frame-wise encoding~\cite{gao2023prediff, gong2024cascast, li2024Latent} to support incremental updates and robustness to frame drops. However, purely frame-wise designs introduce flickering and \textit{temporal inconsistency}~\cite{xing2024Large}.

We therefore propose \emph{FREUD}, a \textbf{FR}amewise \textbf{E}ncoder and \textbf{U}nited \textbf{D}ecoder architecture. Each frame is encoded independently by a lightweight transformer, while a hierarchical rectified flow transformer~\cite{crowson2024Scalable, peebles2023Scalable, ma2024SiT} decodes them jointly. FREUD provides a fast, scalable first stage for latent-space generation and yields temporally consistent reconstructions without adversarial training or loss weighting.

Prior diffusion-based nowcasting models either use a \emph{fixed conditioning window}~\cite{gong2024cascast, gao2023prediff, leinonen2023ldcast}, which is vulnerable to missing or corrupted frames~\cite{kanellopoulos1990Rain, das2010Rain, al-saegh2021Rainfall}, or generate future frames auto-regressively~\cite{price2024GenCast, jiang2024TCDiffusion}, which accumulates errors over time. Robust nowcasting therefore requires flexible, adaptive conditioning.

We achieve this through masking-based diffusion forcing~\cite{hoppe2022ramvid, chen2024diffusionforcing}. This allows inference with any number of conditioning frames and maintains strong skill even with only two past frames.
Recent methods cascade a deterministic forecast with a generative model~\cite{yu2024DiffCast, gong2024cascast, pathak2024KilometerScale}, improving localization but yielding overconfident predictions. We condition solely on observed precipitation, achieving better calibration.
Combining this insight with our novel FREUD compression stage and masking-based rectified flow training yields state-of-the-art precipitation nowcasting on the SEVIR benchmark~\cite{veillette2020SEVIR}. 
Our compact model outperforms prior work in Continuous Ranked Probability Score (CRPS) and perceptual metrics, while maintaining competitive localization without biasing the generation. We formulate our contributions as follows:

\begin{itemize}
    \item \textbf{A simple, scalable first stage tailored for weather nowcasting.} We introduce a rectified-flow encoder-decoder with novel stochastic $\tanh$-regularization that achieves higher reconstruction quality without perceptual or adversarial losses, ensuring a smooth, bounded latent space.
    \item \textbf{Frame-wise encoder and united video decoder.} We propose an asymmetric design that encodes frames independently to prevent information leakage and support variable-length inputs, while jointly decoding all frames simultaneously for temporal consistency.
    \item \textbf{Decoding uncertainty quantification.} Our probabilistic decoder design enables multiple reconstructions from the same latent, providing natural estimates of decoding uncertainty, crucial for safety-critical applications.
    \item \textbf{Scalable and robust latent-space nowcasting.} We achieve state-of-the-art precipitation forecasting through masking-based rectified-flow training that handles variable-length conditioning and frame drops, while scaling with model size, ensemble count, and function evaluations to trade latency for accuracy.
\end{itemize}

\section{Related Works}
\label{sec:related}
\begin{figure}[t]
    \centering
    \includegraphics[width=\linewidth]{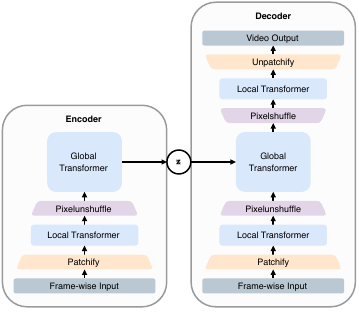}
    \vspace{-0.6cm}
    \caption{\textbf{FREUD architecture.} \textit{Left}: Frame-wise encoder. \textit{Right}: Generative decoder conditioned on encoder latents $\mathbf{z}$.}
    \label{fig:architecture}
    \vspace{-0.2cm}
\end{figure}

\paragraph{Precipitation Nowcasting}
Classical Numerical Weather Prediction (NWP) is computationally expensive, making short-term nowcasts difficult~\cite{shi15:ConvLSTM, shi17:TrajectoryGRU, sonderby20:MetNet, leinonen2023ldcast, ravuri21:GAN, chen2023Foundation}.
Optical-flow extrapolation methods~\cite{woo2017Operational, cheung2012Application, germann2002ScaleDependence, sakaino2013SpatioTemporal, woo2014Application} are faster but cannot capture the formation of new weather patterns~\cite{hur2019Iterative, shi15:ConvLSTM}. These issues motivate deep-learning approaches, categorized as deterministic or probabilistic. Deterministic models learn spatio-temporal mappings from past to future precipitation~\cite{agrawal19:UNet, shi15:ConvLSTM, wang2017predrnn, gao2022simvp, guen2020disentangling, gao2022earthformer}.
However, they often yield blurry forecasts due to regression losses and also lack uncertainty quantification~\cite{ravuri21:GAN, leinonen2023ldcast, gao2023prediff, nai2024Reliable}
Probabilistic approaches use GANs~\cite{ravuri21:GAN, zhang2023skilful, liu2020MPLGAN} or diffusion models (DMs)~\cite{leinonen2023ldcast, gao2023prediff, gong2024cascast, nai2024Reliable, ling2024SRNDiff, pathak2024KilometerScale} to generate weather predictions, producing sharp samples and estimating uncertainty via Monte-Carlo (MC) sampling.
Mode collapse and training instabilities limit GANs~\cite{kossale2022Mode}, making diffusion models the state of the art~\cite{leinonen2023ldcast, gao2023prediff, yu2024DiffCast, gong2024cascast}.
Many methods operate in the latent space of a compression autoencoder~\cite{leinonen2023ldcast, gao2023prediff, gong2024cascast}, but ignore compression uncertainty~\cite{ling2024SRNDiff}. Moreover, deterministic conditioning~\cite{gong2024cascast, yu2024DiffCast, pathak2022FourCastNet}, often biases the generation~\cite{pathak2024KilometerScale}. We address these issues with a first stage that quantifies compression uncertainty and enables purely data-driven nowcasting. Further discussion of related works is provided in the Appendix.

\paragraph{Uncertainty Estimation}
in ML differentiates uncertainty from underdeterminedness of the problem (aleatoric uncertainty) or underspecification of the model (epistemic uncertainty)~\cite{hullermeier2021Aleatoric}. Aleatoric uncertainty is important in weather nowcasting due to non-linearities in the atmospheric system~\cite{leinonen2023ldcast, nai2024Reliable, shi2024CoDiCast, hong2006Uncertainty}. Traditionally, Bayesian Neural Networks~\cite{mackay_practical_1992, mackay_bayesian_1992}, MC Dropout~\cite{srivastava_dropout_2014, gal_dropout_2016}, and model ensembling~\cite{lakshminarayanan2017Simple} are used to estimate uncertainty but are either impractical for large generative models or degrade performance~\cite{valdenegro-toro_deeper_2022}. DMs allow zero-shot aleatoric uncertainty quantification~\cite{chan2024HyperDiffusion, shu2024ZeroShot} by sampling multiple times given the same conditioning. The variance of this ensemble serves as an estimate of aleatoric uncertainty~\cite{chan2024HyperDiffusion, leinonen2023ldcast, gao2023prediff, nai2024Reliable}.

\paragraph{Diffusion Autoencoders}
(DiffAEs) replace the deterministic decoder with a generative model conditioned on encoder outputs~\cite{preechakul2022Diffusion}.
Previous work uses discretized~\cite{shi2022DiVAE, bachmann20244M21, yu2024Image, bachmann2025FlexTok} as well as continuous~\cite{zhao2025EpsilonVAE, chen2025Diffusion, betker2023Improving, hudson2024SODA} latent encodings to enable downstream generation in the latent space~\cite{zhao2025EpsilonVAE, chen2025Diffusion, betker2023Improving, hudson2024SODA, shi2022DiVAE, bachmann2025FlexTok}. Compared to standard VAEs, DiffAEs yield better detail reconstruction for images and more stable training~\cite{zhao2025EpsilonVAE, chen2025Diffusion}.
Earlier video generation methods use frame-wise compression~\cite{an2023LatentShift, ge2023Preserve, blattmann2023Align, he2023Latenta, luo2023videofusion, singer2022MakeAVideo, wang2024Swap}, causing temporal inconsistencies~\cite{zhang25:REGEN, xing2024Large}, or 3D encoder-decoder models~\cite{themoviegenteam@meta2024Movie, yu2023Languagea, wang2024OmniTokenizer, tang2024VidTok, li2025WFVAE}, with limited capacity~\cite{zhang25:REGEN}. 
Concurrent DiffAE-based video compressors~\cite{yang2025Rethinking, zhang25:REGEN} achieve superior reconstructions over VAE baselines with few sampling steps and strong generative downstream performance. We improve upon these approaches by using a hierarchical Transformer-based architecture, unlocking scalable performance, and introducing a novel regularization.  Moreover, our design better suits forecasting by avoiding perceptual losses, encoding frames independently, and assessing reconstruction uncertainty.

\section{Method}
\label{sec:method}
In the following, we consider sequences of precipitation maps $\mathbf{x}^t \in \mathbb{R}^{C \times H \times W}$, where $C$, $H$, and $W$ denote channels, height, and width, respectively. The superscript $t$ indexes time frames, and the subscript $i$ denotes diffusion steps $\mathbf{x}_i$. Given the $L^{\text{in}}$ previous precipitation maps, the task is to predict the next $L^{\text{out}}$ frames.
As we cannot model the chaotic nature of precipitation deterministically, we sample from the conditional distribution $p(\mathbf{x}^{\text{out}}|\mathbf{c})$ of possible future precipitation using a generative model.
We follow previous work and cast the precipitation nowcasting problem as a probabilistic spatio-temporal prediction task~\cite{gao2023prediff, yu2024DiffCast, gao2022earthformer}.
Hence, future frames $\mathbf{x}^{\text{out}}$ are sequences of $L^{\text{out}}$ precipitation maps while the conditioning $\mathbf{c}$ are $L^{\text{in}}$ previous frames.

\subsection{Preliminary}\label{sec:preliminaries}

\paragraph{Rectified Flow}
\label{sec:rectified-flow}
Generative models that transform noise into data have recently gained significant attention, particularly diffusion-~\cite{song2019scorebased, song2020ddim, ho2020ddpm, rombach2022ldm} and flow-based models~\cite{dinh2017density, chen2018neural, liu2022Flow, albergo2022stochint, albergo2023stochastic, lipman2023flowmatching}.
\textit{Rectified Flows} (RF) \cite{liu2022Flow, esser2024scaling} or \textit{Stochastic Interpolants} \cite{albergo2022stochint, albergo2023stochastic, ma2024SiT} generalize these frameworks by mapping samples $\mathbf{x}_0 \sim p_0$ from a simple prior (typically $\mathcal{N}(0,\mathbf{I})$) to data samples $\mathbf{x}_1 \sim p_1$ through the continuous interpolation $\mathbf{x}_i = \alpha_i \mathbf{x}_1 + \sigma_i \mathbf{x}_0$, where $\alpha_i$ increases and $\sigma_i$ decreases over time, controlling the transition from noise to data. 
The velocity field is defined as
\begin{equation}
\label{eq:velocity}
    \mathbf{v}(\mathbf{x}_i, i)
    = \frac{d \mathbf{x}_i}{di}
    = \dot{\alpha}_i \mathbf{x}_1 + \dot{\sigma}_i \mathbf{x}_0
\end{equation}
where $\dot{\alpha}_i$ and $\dot{\sigma}_i$ are time derivatives of $\alpha_i$ and $\sigma_i$. This induces a probability flow ODE whose marginal distribution at time $i$ equals $p_i(\mathbf{x})$.
Following prior work~\cite{esser2024scaling, ma2024SiT, lipman2023flowmatching}, we adopt the linear interpolant $\alpha_i=i$, $\sigma_i=1-i$.
A neural network $\mathbf{v}_\theta(\mathbf{x}_i, i)$ with parameters $\theta$ is trained to predict the velocity field by minimizing
\begin{equation}\label{eq:flow-loss}
    \mathcal{L}_\mathbf{v}(\theta)
    =
    \int \mathbb{E}[\lVert \mathbf{v}_\theta(\mathbf{x}_i, i) - (\dot{\alpha}_i \mathbf{x}_1 + \dot{\sigma}_i \mathbf{x}_0) \rVert^2] di .
\end{equation}
After training, samples from the prior are transported to the data distribution by numerically integrating the ODE with $\mathbf{v}_\theta(\mathbf{x}_i, i)$ using standard solvers, e.g., Euler integration.

\begin{figure}
    \centering
    \includegraphics[width=\linewidth]{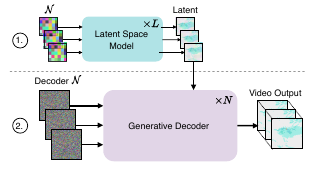}
    \vspace{-0.4cm}
    \caption{\textbf{Inference pipeline}. We first generate weather forecasts in the latent space of our FREUD encoder and then decode the predictions with our generative FREUD decoder.}
    \label{fig:inference}
\end{figure}

\paragraph{Uncertainty Quantification}

Following Chan et al.~\cite{chan2024HyperDiffusion}, we consider the atmosphere to be a dynamic system $\mathbf{x}^{\text{out}} = \mathcal{F}(\mathbf{c}) + \mathbf{\eta}$ with deterministic dynamics $\mathcal{F}(\mathbf{c})$ and irreducible or unmodeled noise $\mathbf{\eta}$. Even if we can perfectly capture the deterministic dynamics of the system $\mathcal{F}(\mathbf{c})$, there remains an error because of the stochastic component. Therefore, if we can draw Monte-Carlo samples $\mathbf{x}^{\text{out}}_e$ from the distribution, the sample variance 
\begin{equation}
    \text{Var}(\mathbf{\tilde{x}}^{\text{out}}),\quad \mathbf{\tilde x}^{\text{out}} = [\mathbf{x}_e^{\text{out}}]_{e=0}^N, \quad \mathbf{x}^{\text{out}}_i \sim p(\mathbf{x}^{\text{out}} | \mathbf{c}) = \mathcal{F}(\mathbf{c}) + \eta
\end{equation}
where the subscript $\cdot_e$ indexes independently sampled ensemble members, will converge to the irreducible aleatoric uncertainty $\text{Var}(\eta)$ for $N \rightarrow \infty$~\cite{chan2024HyperDiffusion}.
As we can approximate $p(\mathbf{x}^{\text{out}} | \mathbf{c})$ using a generative model, we can estimate the uncertainty by computing $N$ probabilistic predictions and calculating the ensemble variance \cite{chan2024HyperDiffusion, leinonen2023ldcast, nai2024Reliable, shi2024CoDiCast}.

\begin{figure*}[tb]
    \centering \small
    \setlength\tabcolsep{1pt}

    \newcommand{\imagepng}[3]{%
        \includegraphics[width=0.1\linewidth]{fig/lsm/forecast/T-L-seed_42/#1_#2_#3.png}%
    }

    \newcommand{\rowpng}[2]{%
        \imagepng{#1}{#2}{1} & %
        \imagepng{#1}{#2}{3} & %
        \imagepng{#1}{#2}{5} & %
        \imagepng{#1}{#2}{7} & %
        \imagepng{#1}{#2}{9} & %
        \imagepng{#1}{#2}{11}   %
    }

        \begin{subfigure}{1\textwidth}
        \centering
        \begin{tabular}{lcccccccc}
            & $+10$\,min & $+20$\,min & $+30$\,min & $+40$\,min & $+50$\,min & $+60$\,min 
            & \hspace{0.4cm} \multirow{3}{*}{%
              \raisebox{-1.05\height}{%
                \includegraphics[width=0.1\linewidth]{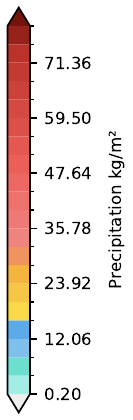}%
              }%
            } \\
            \rotatebox{90}{\hspace{2em} E1}
                & \rowpng{ens}{0}  \\
            \rotatebox{90}{\hspace{2em} E2}
                & \rowpng{ens}{1}\\
            \arrayrulecolor{gray!50!white}
            \cmidrule{1-7}
            \arrayrulecolor{black}\\
            \rotatebox{90}{\hspace{2em} GT} & \rowpng{gt}{0}\\
        \end{tabular}
    \end{subfigure}
    \vspace{-0.3cm}
    \caption{\textbf{Forecast ensemble} from our L model. Red rectangles show zoom-ins to highlight differences. Best viewed zoomed in.
    }
    \label{fig:lsm:qualitative}
\end{figure*}

\subsection{FREUD}
Diffusion- and flow-based generative methods iteratively sample in data space, which is particularly costly for high-resolution weather data. Therefore, following prior work on image and video synthesis, we perform generative modeling in a compressed latent space, which substantially reduces computational complexity \cite{rombach2022ldm, themoviegenteam@meta2024Movie, rombach2022ldm, gong2024cascast, leinonen2023ldcast}.

For operational weather nowcasting, a first-stage encoder–decoder should fulfill several key properties. 
First, \textbf{training should be simple and stable}, avoiding adversarial or perceptual objectives. 
Second, the model must offer \textbf{efficient compression} of high-dimensional radar. 
Third, its latent space should be \textbf{smooth and well-structured}, amenable to generative modeling.
Fourth, the model should enable \textbf{uncertainty estimation} in the decoding process, an essential property for safety-critical applications. 
Fifth, \textbf{frame-wise encoding} is desirable to ensure robustness to sensor faults or dropped frames, common in radar networks. 
Finally, outputs must be \textbf{temporally coherent}, since flickering reconstructions across frames would lead to inconsistencies in subsequent forecasts.
They motivate our design of the FREUD framework, which features an architecture that provides efficient compression, robustness to sensor faults, and temporally coherent reconstructions.

\paragraph{Architecture}
\cref{fig:architecture} shows our FREUD architecture. To ensure efficiency, we follow recent works \cite{zheng2025rea, gui2025reptok} and use a fully transformer-based encoder-decoder design.

\noindent\textit{Encoder}.
The encoder is implemented as a lightweight transformer that operates frame-wise, allowing each frame to be processed independently. This makes the model robust to missing or corrupted radar frames, which are common in operational weather sensing, and enables online updates as new frames become available without requiring re-encoding of the entire sequence. The frame-wise design further prevents information leakage from future to past frames, which sequence-level encoders would suffer from, thereby preserving the causal structure essential for forecasting.

\noindent\textit{Decoder}.
For decoding, we use a transformer-based video decoder that jointly reconstructs all frames to ensure temporal consistency. To cope with the high computational cost of attention in large video tensors, we follow \textit{Hourglass diffusion transformers}~\cite{crowson2024Scalable}, and adopt a hierarchical transformer architecture. Here, the spatial resolution is progressively reduced and later restored through pixel-unshuffle and pixel-shuffle operations, which rearrange spatial dimensions into channel dimensions (and vice versa) \cite{shi2016RealTime}.
We condition the decoder on the encoder latents by concatenating them channel-wise in the bottleneck, guiding the video reconstruction process through the encoded spatial representations.
We further improve efficiency through space-time factorized attention~\cite{harvey2022flexible}. Each block alternates between spatial and temporal attention, followed by a feed-forward network. Spatial attention attends to all tokens within each frame, and temporal attention attends to all tokens in the same spatial positions across all frames.

\noindent Finally, we further reduce complexity in high-resolution stages by integrating neighborhood attention \cite{hassani2023Neighborhood}, which restricts self-attention to local spatial neighborhoods rather than full global context.

\paragraph{Training Paradigm}
Previous works train autoencoders with a small KL regularization to encourage a smooth latent space, and additional perceptual and adversarial losses to boost sharpness~\cite{rombach2022ldm, gong2024cascast, leinonen2023ldcast}. However, these objectives require careful loss balancing~\cite{chen2025Diffusion, zhao2025EpsilonVAE, zhang25:REGEN, yang2025Rethinking}, and adversarial training often leads to training instability and mode collapse, while both perceptual and adversarial losses can suppress subtle yet safety-critical details~\cite{ling2024SRNDiff, kossale2022Mode}. Moreover, their decoders are deterministic at inference, providing no estimate of decoding uncertainty.
We address these limitations by training our first stage with the rectified flow (RF) loss from~\cref{eq:flow-loss}. Optimization with this loss is stable and makes perceptual and adversarial objectives obsolete, greatly simplifying training. It further allows us to sample from our decoder during inference, which naturally provides an empirical way to quantify decoding uncertainty through multiple reconstructions.
We condition the decoder on the frame-wise latents and train both the encoder and decoder jointly with the RF loss from~\cref{eq:flow-loss}.

\paragraph{Latent Space Regularization}
A well-structured latent space is essential for latent space generative modeling~\cite{gui2025reptok, yao2025Reconstruction, zheng2025rea}. Current autoencoders typically include a small KL regularization term to encourage a smooth latent space~\cite{rombach2022ldm}. In contrast, our framework does not inherently constrain the latent space, leading to high variance and low density. Thus, in addition to the \textit{unregularized} (\textit{unreg.}) case, we introduce and ablate two additional regularization schemes to ensure robust latent representations.

\noindent\textit{KL-regularization} (\textit{KL-reg.}), as used in latent diffusion autoencoders~\cite{rombach2022ldm}, penalizes the divergence between the latent distribution and a standard normal. This encourages smoother and more robust latent representations but introduces a trade-off: stronger KL weights improve regularization at the cost of reconstruction fidelity, requiring loss balancing. Moreover, it requires architectural modifications, as the encoder must predict both the mean and the standard deviation per latent dimension.

\noindent\textit{Stochastic $\tanh$ regularization} (\textit{T-reg.}).
We propose \textit{T-reg.} as a simple yet effective alternative to KL-based regularization that achieves all desired properties of a well-behaved latent space: bounded, smooth, and amenable to generation. We constrain the latent values to $[-1, 1]$ via a $\tanh$ activation and add a small Gaussian perturbation $\epsilon \sim \mathcal{N}(0, \sigma^2 \mathbf{I})$ with variance $\sigma^2$ to the latent representation, so that
\begin{equation}
\mathbf{\tilde z}^t = \tanh(\text{Enc}_\theta(\mathbf{x}^t)) + \sigma \epsilon
\end{equation}
where $\text{Enc}_\theta$ denotes the frame-wise FREUD encoder and $\mathbf{x}^t$ the input frame. This stochastic perturbation makes nearby latents decode into similar pixel-space videos, encouraging smoothness and robustness to small changes in the latent space. Unlike KL regularization, \textit{T-reg.} acts purely as an architectural constraint rather than an additional loss, eliminating the need for loss balancing. The final loss for training the FREUD first stage with \textit{T-reg.} is thus just the flow loss from \cref{eq:flow-loss}, simplified with the linear schedule to
\begin{equation}
\mathcal{L} = \lVert \mathbf{v}_\theta(\mathbf{x}_i, i) - (\mathbf{x}_1 - \mathbf{x}_0) \rVert^2
\end{equation}
where $\mathbf{x}_1$ is a data sample and $\mathbf{x}_0 \sim \mathcal{N}(0, \mathbf{I})$.
Similar to \cite{themoviegenteam@meta2024Movie}, we find that a simple outlier punishment is helpful in early training (for details see Appendix).

\begin{figure}[t]
    \centering
    \includegraphics[width=0.99\linewidth]{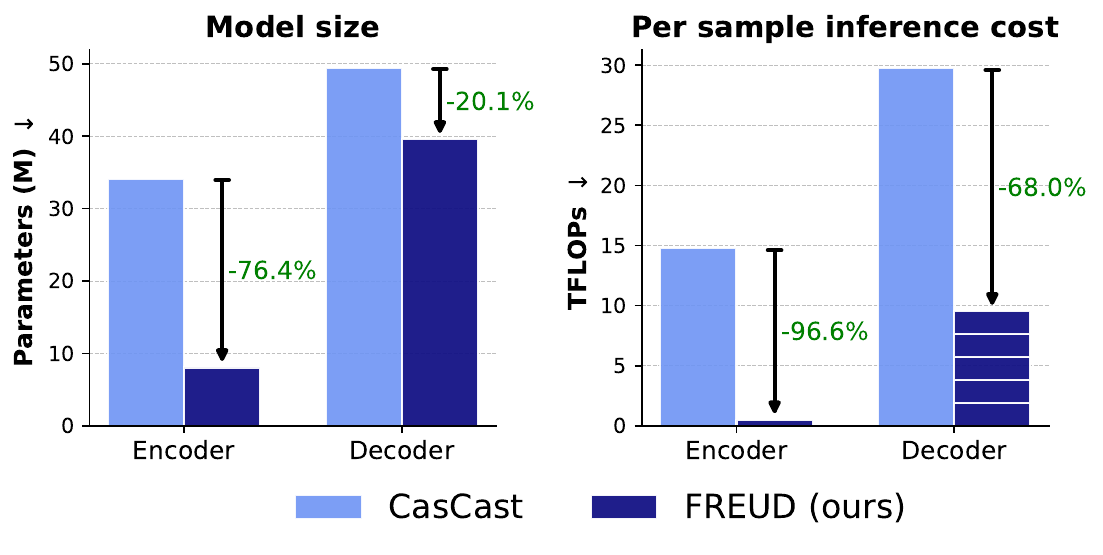}
    \vspace{-0.4cm}
    \caption{\textbf{FREUD efficiency.} Our transformer-based autoencoder has fewer parameters and uses fewer FLOPs for encoding and decoding (5 NFE). This allows for faster training and inference.
    }
    \label{fig:freud-efficiency}
\end{figure}

\subsection{Latent Space Nowcasting}\label{sec:latent_space_nowcasting}
With FREUD as our first stage, we perform nowcasting directly in its latent space. Its frame-wise encoder tolerates missing frames, so our latent model must likewise support variable-length conditioning and remain robust under limited past observations.
For this we adopt a masking-based training strategy, which teaches the model to infer future dynamics from arbitrary subsets of past frames and allows flexible conditioning lengths during inference, ensuring resilience to missing or corrupted frames.

\paragraph{Training}
We follow the simple training paradigm of RaMViD~\cite{hoppe2022ramvid}, where we randomly mask part of each clip and learn to denoise the remaining part.
More formally, we split a video of length \mbox{$T = L^{\text{in}} + L^{\text{out}}$} into conditioning frames \mbox{$\mathbf{x}_1^{\mathcal{C}}$} and generation frames \mbox{$\mathbf{x}_1^{\mathcal{G}}$} where \mbox{$\mathcal{C} \cup \mathcal{G} = \{0,\dots,T-1\}$} and \mbox{$\mathcal{C} \cap \mathcal{G} = \emptyset$}.
The final video is given by \mbox{$\mathbf{x}_{1} = \mathbf{x}_1^\mathcal{C} \oplus \mathbf{x}_1^\mathcal{G}$} where $\oplus$ is a frame selection operation.
For every sample we draw $k$ conditioning indices uniformly from $\{1, ..., K\}$, where the hyperparameter $K<T$ defines the maximum number of conditioning frames.
Only the remaining frames with indices in $\mathcal{G} = \{0, \dots T-1\} \setminus \mathcal{C}$, are noised, so the model input is  $\mathbf{x}_i = \mathbf{x}_1^{\mathcal{C}} \oplus \mathbf{x}_i^{\mathcal{G}}$. Randomising the amount of conditioning $|\mathcal{C}| = k$ during training teaches the model to leverage any variable subsets of information to make its predictions. We compute the loss from \cref{eq:flow-loss} only on noisy frames $\mathbf{x}_i^{\mathcal{G}}$.
Similar to \cite{hoppe2022ramvid} we define probability $p_\mathcal{U}$ with which we train fully unconditional, i.e., set $\mathcal{C} = \emptyset$. This enables classifier-free guidance~\cite{ho2022ClassifierFree}.

\paragraph{Inference}
At inference we encode the latest $L^{\text{in}}$ available observations with the frame-wise encoder and append $L^{\text{out}}$ Gaussian latent frames $\mathbf{x}_0^t \in \mathbb{R}^{C_l \times H_l \times W_l}$. This yields a sequence of \mbox{$T = L^{\text{in}} + L^{\text{out}}$} latent frames, consisting of conditioning latents and noisy targets. The latent rectified flow model then denoises this sequence, and the FREUD decoder maps the denoised latents back to pixel space, producing a forecast for the $L^{\text{out}}$ future frames. See \cref{fig:inference} for an overview of the full pipeline.
To estimate predictive uncertainty, we form latent-space ensembles by resampling the $L^{\text{out}}$ with fixed conditioning. Additionally, we can generate decoder ensembles by rerunning the decoding process with different noise initializations for the same latent sequence.

\section{Experiments}
\label{sec:experiments}
Following prior work~\cite{gao2022earthformer, gong2024cascast}, we evaluate our method on the \textit{SEVIR} benchmark~\cite{veillette2020SEVIR}, which consists of 20,393 (extreme) weather events, collected from 2017 to 2019. Each event covers a $384\times384$\,km region for a $4$\,h timespan. Vertically Integrated Liquid (VIL) from the NEXRAD radar mosaic indicates precipitation and is recorded with $1$\,km spatial and $5$\,min temporal resolution. We provide more details on the SEVIR dataset, the computational resources used for experiments, as well as additional experiments on the MeteoNet benchmark~\cite{larvor2020meteonet} in the Appendix.

\begin{table}[tb]
    \centering
    \caption{\textbf{Reconstruction performance} of the framewise CasCast autoencoder, a framewise Diffusion Autoencoder, and FREUD variants with different regularization.}
    \label{tab:first_stage_comp}
    \footnotesize
\centering
\setlength{\tabcolsep}{3pt}
\begin{tabularx}{\linewidth}{lccccccc}
    \toprule
    \textbf{Model}
    & \makecell{\textbf{Ens.}}
    & \makecell{\textbf{RMSE} $\downarrow$}
    & \makecell{\textbf{MAE} $\downarrow$}
    & \makecell{\textbf{SSIM} $\uparrow$}
    & \makecell{\textbf{PSNR} $\uparrow$}
    & \makecell{\textbf{dMAE} $\downarrow$}
    \\
    \midrule
    \multirow{1}{*}{CasCast \small \cite{gong2024cascast}}
      & --  & 0.022 & 0.008 & 0.976 & 39.153 & 0.012 \\
    \arrayrulecolor{gray!50!white}
    \midrule
    \arrayrulecolor{black}
    \multirow{2}{*}{\makecell{T-reg.\\DiffAE}}
  & 1  & 0.027 & 0.011 & 0.970 & 37.415 & 0.015 \\                     
  & 10 & 0.025 & 0.010 & 0.974 & 38.13 & 0.012 \\
    \arrayrulecolor{gray!50!white}
    \midrule
    \arrayrulecolor{black}
    \multirow{2}{*}{\makecell{unreg.\\FREUD}}
  & 1  & 0.027 & 0.009 & 0.989 & 37.550 & 0.013 \\                     
  & 10 & 0.023 & 0.008 & 0.987 & 38.915 & 0.012 \\
    \arrayrulecolor{gray!50!white}
    \midrule
    \arrayrulecolor{black}
    \multirow{2}{*}{\makecell{KL-reg.\\FREUD}}
      & 1  & 0.025 & 0.009 & 0.992 & 37.927 & 0.013 \\
      & 10 & 0.022 & 0.008 & 0.987 & 39.029 & 0.011 \\
    \arrayrulecolor{gray!50!white}
    \midrule
    \arrayrulecolor{black}
    \multirow{2}{*}{\makecell{T-reg.\\FREUD}}
      & 1  & 0.019 & 0.008 & 0.998 & 40.224 & 0.011 \\
      & 10 & \textbf{0.018} & \textbf{0.007} & \textbf{0.999} & \textbf{41.085} & \textbf{0.010} \\
    \bottomrule
\end{tabularx}

\end{table}

\subsection{FREUD First Stage}
First, we evaluate our frame-wise encoder and joint decoder architecture. We follow prior work~\cite{chen2025Diffusion, zhao2025EpsilonVAE, zhang25:REGEN, yang2025Rethinking} and report RMSE, MAE, SSIM, and PSNR. We capture temporal smoothness with the MAE of discrete time derivatives (dMAE), and overall predictive uncertainty by the variance (Var) over ensemble members, as defined in~\cref{sec:preliminaries}.

\paragraph{Reconstruction Performance}
We report reconstruction metrics for FREUD and the prior state-of-the-art frame-wise autoencoder CasCast~\cite{gong2024cascast} in \cref{tab:first_stage_comp}.
Our \textit{T-reg.}\thinspace FREUD achieves superior results across all metrics, with a significant gain in SSIM, indicating perceptually accurate reconstructions without adversarial or perceptual losses.
It further needs fewer parameters and fewer Floating Point Operations (FLOPs) than the CasCast autoencoder, as shown in~\cref{fig:freud-efficiency}. Encoding is ~96\%, while decoding is 68\% faster with five NFE. This efficiency pays off twice: fast encoding accelerates latent-space model training, while efficient encoding and decoding enable rapid forecast updates during inference, which is crucial for operational nowcasting.
In summary, FREUD is not only easier to train, but also more compute-efficient, and delivers higher-fidelity reconstructions, even with only one ensemble member.

\begin{figure}[bt]
    \centering
    \includegraphics[width=0.85\linewidth]{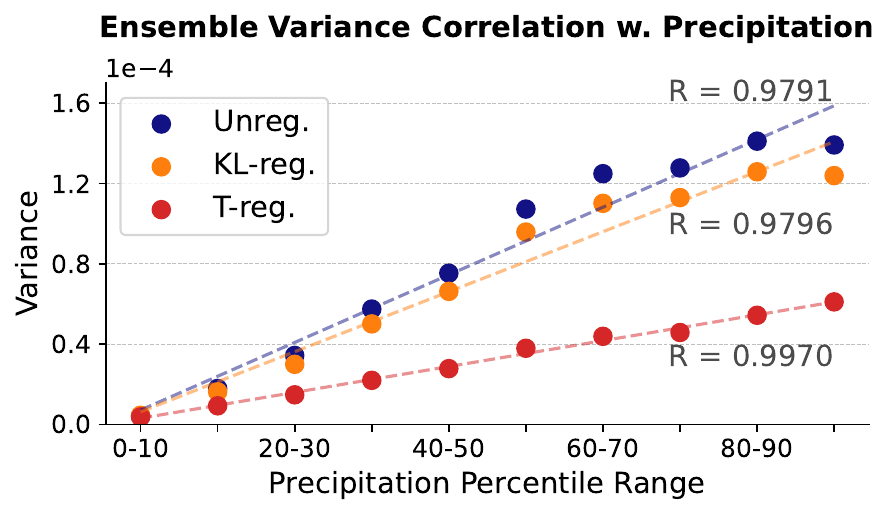}
    \vspace{-0.2cm}
    \caption{
    \textbf{Correlation between precipitation intensity and decoding variance.} All FREUD variants show a strong linear relationship, with \textit{T-reg.} achieving the highest correlation.}
    \label{fig:extreme_v_var_main}
    \vspace{-0.2cm}
\end{figure}

\begin{figure}[tb]
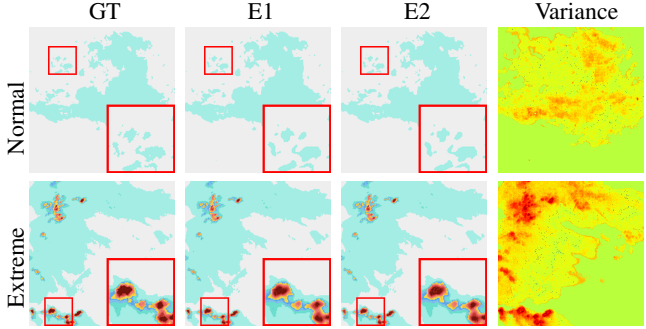

    \centering
    \small
    \setlength\tabcolsep{1pt}
    \newcommand{\imagepng}[3]{
        \includegraphics[width=0.23\linewidth]{fig/freud/ensembles/T-seed_42_2/#1_precip/#2_#3.png}
    }
    \newcommand{\rowpng}[1]{
        \imagepng{#1}{gt}{0} &
        \imagepng{#1}{ens}{0} &
        \imagepng{#1}{ens}{1} &
        \imagepng{#1}{var}{0}
    }

    \begin{subfigure}{1\linewidth}
        \centering
        \begin{tabular}{ccccc}
        & GT & E1 & E2 & Variance \\
        \rotatebox{90}{\hspace{.5em} Normal} & \rowpng{low} \\
        \rotatebox{90}{\hspace{.5em} Extreme} & \rowpng{high} \\
        \end{tabular}
    \end{subfigure}
    \vspace{-0.2cm}
    \caption{\textbf{Qualitative reconstructions} of a normal and an extreme weather event, with corresponding ensemble members and ensemble variance. The red rectangle highlights a zoomed-in area for visualizing differences in details. Best viewed zoomed in.
    }
    \label{fig:freud:qualitative}
    \vspace{-0.2cm}
\end{figure}

\paragraph{Decoding Uncertainty}
Through the generative decoder, FREUD can sample multiple realizations given the same conditioning. As described in \cref{sec:preliminaries}, we estimate decoding uncertainty from the variance across reconstructions. \cref{fig:freud:qualitative} shows a five-member ensemble of FREUD for both normal and extreme weather events along with corresponding ensemble variance maps.
While reconstructions are highly similar overall, they differ in small-scale structures, particularly in regions of intense precipitation. Unlike natural image synthesis, where such variations are imperceptible, in weather forecasting they reflect genuine differences in high-impact regions.
The variance maps reveal a clear correlation between precipitation intensity and reconstruction variance, consistent with the non-linearity and chaotic dynamics of severe weather. This shows that ensemble-based reconstructions capture meaningful variability precisely in regions of heightened uncertainty, providing valuable insight for safety-critical extreme events.
\cref{fig:extreme_v_var_main} further confirms a strong linear relationship between precipitation percentile and intra-ensemble variance across all FREUD variants, with the \textit{T-reg.} model showing the highest correlation ($r = 0.997$).
In summary, these findings confirm that first-stage ensembling effectively identifies dynamically complex weather, providing meaningful and spatially localized uncertainty estimates.
Additional experiments inserting abnormal patterns, similar to~\cite{chan2024HyperDiffusion}, and more qualitative results are presented in the Appendix.

\subsection{Forecasting performance}
Given the FREUD first stage, we train a latent space model for forecasting. We follow prior works~\cite{gao2022earthformer, gong2024cascast} and use $L^{\text{in}} = 13$ frames ($65$\,min) of previous precipitation to predict the next $L^{\text{out}} = 12$ frames ($60$\,min).
Unless specified otherwise, we use ten forecasting ensemble members similar to prior work~\cite{gong2024cascast, gao2023prediff, leinonen2023ldcast, yu2024DiffCast}.
We use the evaluation pipeline provided by Gong et al. \cite{gong2024cascast} and evaluate probabilistic nowcasting performance with the Continuous Ranked Probability Score (CRPS), perceptual realism with the Structural Similarity Index Measure (SSIM), and report Heidke Skill Score (HSS) and Critical Success Index (CSI). HSS expresses improvement over random chance, with $1$ indicating a perfect forecast and $0$ indicating no improvement over guessing. CSI quantifies the hit rate while ignoring true negatives, making it robust when no events dominate. Both scores are averaged across six precipitation/no precipitation thresholds. We provide more details in the Appendix.

\paragraph{State-of-the-Art Comparison}
\begin{table}[t]
\centering
\caption{\textbf{SEVIR benchmark} comparison of our method with various deterministic and probabilistic models. Baselines are sourced from CasCast \cite{gong2024cascast}. For $\dag$ the model is trained on a downsampled dataset with a size of 128 since they found that training on the original dataset does not yield high-quality predictions.}
\label{tab:sota_sevir}
\vspace{-0.2cm}
\footnotesize
\setlength{\tabcolsep}{5.1pt}
\centering
\newcommand{\cfgresult}[1]{\textcolor{black!60}{#1}}
\begin{tabular}{
lccccc
} 

\toprule

\textbf{Method} &
\textbf{CRPS}$\downarrow$ &
\textbf{SSIM}$\uparrow$ &
\textbf{HSS}$\uparrow$ &
\textbf{CSI}$\uparrow$
\\

\midrule

ConvLSTM~\cite{shi15:ConvLSTM}~\ven{NeurIPS '15}
& 0.0264        %
& 0.7749        %
& 0.5232        %
& 0.4102        %
\\
PredRNN~\cite{wang2017predrnn}~\ven{TPAMI '22}
& 0.0271        %
& 0.7497        %
& 0.5192        %
& 0.4045        %
\\
PhyDNet~\cite{guen2020disentangling}~\ven{CVPR '20}
& 0.0253        %
& 0.7649        %
& 0.5311        %
& 0.4198        %
\\
SimVP~\cite{gao2022simvp}~\ven{CVPR '22}
& 0.0259        %
& 0.7772        %
& 0.5280        %
& 0.4153        %
\\
EarthFormer~\cite{gao2022earthformer}~\ven{NeurIPS '22}
& 0.0251        %
& 0.7756        %
& 0.5411        %
& 0.4310        %
\\

\arrayrulecolor{gray!50!white}
\midrule
\arrayrulecolor{black}

NowcastNet~\cite{zhang2023skilful}~\ven{Nature '23}
& 0.0283        %
& 0.5696        %
& 0.5365        %
& 0.4152        %
\\
PreDiff$^\dag$~\cite{gao2023prediff}~\ven{NeurIPS '23}
& 0.0202        %
& 0.7648        %
& 0.4914        %
& 0.3875        %
\\
CasCast~\cite{gong2024cascast}~\ven{ICML '24}
& 0.0202 %
& 0.7797 %
& \textbf{0.5602} %
& \textbf{0.4401} %
\\

\arrayrulecolor{gray!50!white}
\midrule
\arrayrulecolor{black}

FREUD + LSM-L (\textit{ours})
& \textbf{0.0190} %
& 0.7841 %
& 0.5011 %
& 0.3864 %
\\
\cfgresult{\rotatebox[origin=c]{180}{$\Lsh$} with cfg} %
& \cfgresult{0.0192} %
& \textbf{0.7937} %
& \cfgresult{0.5537} %
& \cfgresult{0.4277} %
\\

\bottomrule
\end{tabular}

\end{table}
\cref{tab:sota_sevir} quantitatively compares our method to current state-of-the-art weather nowcasting methods on the SEVIR benchmark.
Our Large Latent Space Model (LSM-L) combined with the \textit{T-reg.} FREUD first stage achieves state-of-the-art CRPS and SSIM, outperforming all existing methods in both distribution coverage and perceptual quality.
Our method provides significant gains, given that improvements on SEVIR are typically small. For instance, CasCast~\cite{gong2024cascast} improves over PreDiff~\cite{gao2023prediff} by $0$\% CRPS and $+1.95$\% SSIM, whereas our method improves over CasCast by $+5.94$\% CRPS and $+1.80$\% SSIM.
CasCast reports its state-of-the-art results using Classifier-Free Guidance (CFG)~\cite{ho2022ClassifierFree}, which also improves localization metrics in our model. With CFG, our method becomes competitive on HSS and CSI, being only outperformed by CasCast on HSS. However, we observe that CFG systematically increases predicted precipitation for both CasCast and our method, suggesting that CFG-based improvements in localization may stem from this simple shift rather than better modeling. We therefore consider CFG to be flawed for nowcasting and further ablate its effects in the Appendix.
Qualitative results in \cref{fig:lsm:qualitative} show that our predictions are visually consistent with the ground truth and capture realistic variations in localized precipitation patterns. We provide more samples in the Appendix.

\begin{figure}
    \centering
    \includegraphics[width=0.85\linewidth]{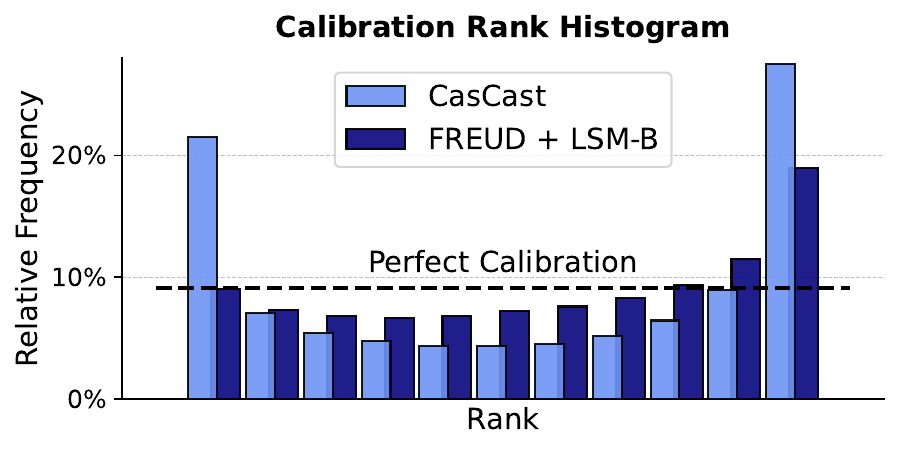}
    \vspace{-0.3cm}
    \caption{\textbf{Calibration Rank Histogram} for our pipeline. Our rank histogram is flatter, indicating improved calibration compared to CasCast~\cite{gong2024cascast}.
    }
    \label{fig:lsm:rank-histogram}
\end{figure}

\paragraph{Calibration}
Similar to previous work~\cite{leinonen2023ldcast, li2024Generativea, brecht2024Replacing}, we use ensemble rank histograms to assess calibration: a uniform shape indicates well-calibrated uncertainty, whereas a U-shape reflects overconfidence. As shown in \cref{fig:lsm:rank-histogram}, our ensemble is substantially closer to uniform, indicating better calibration. In contrast, CasCasts rank histogram indicates strong overconfidence. Quantitative assessment via the reliability index (RI)~\cite{dellemonache2006Probabilistic, wilks2019Indices}, with 1,000 bootstrap resamples to compute a 95\% confidence interval, confirms our approach achieves a significantly better reliability error ($RI = 0.135 \pm 0.001$) than CasCast ($RI = 0.312 \pm 0.001$).

\paragraph{Scalablity}
We show the scalability of our approach by training models of different sizes, with quantitative results reported in \cref{tab:lsm:scaling}. Similar to trends observed in image synthesis, we find that larger latent-space models consistently improve performance across all metrics. Notably, even our smallest model (LSM-S) achieves competitive performance despite using only a fraction of the parameters of prior work. Our largest model (LSM-L) achieves the best overall results, confirming that our approach scales favorably with model capacity.

\begin{table}[tb]
    \caption{\textbf{Scaling Analysis} of the forecasting latent space flow model. Localization metrics and CRPS consistently improve with larger model sizes. All models outperform CasCast for CRPS.}
    \label{tab:lsm:scaling}
    \vspace{-0.2cm}
    \footnotesize
\setlength{\tabcolsep}{4pt}
\centering
\newcommand{\cfgresult}[1]{\textcolor{black!60}{#1}}
\begin{tabular}{lccccc} 
\toprule
Model & Num. Params & CRPS $\downarrow$ & SSIM$ \uparrow$ & HSS $\uparrow$ & CSI $\uparrow$\\
\midrule
CasCast~\cite{gong2024cascast}
& 309M
& 0.0202 %
& 0.7797 %
& \textbf{0.5602} %
& \textbf{0.4401} %
\\

\arrayrulecolor{gray!50!white}
\midrule
\arrayrulecolor{black}

Ours LSM-S
& 44M
& 0.0200 %
& 0.7742 %
& 0.4514 %
& 0.3485 %
\\

\cfgresult{\rotatebox[origin=c]{180}{$\Lsh$} with cfg} %
& 
& \cfgresult{0.0201} %
& \cfgresult{0.7887} %
& \cfgresult{0.5155} %
& \cfgresult{0.3962} %
\\

Ours LSM-B
& 141M
& 0.0196 %
& 0.7828 %
& 0.4923 %
& 0.3805 %
\\

\cfgresult{\rotatebox[origin=c]{180}{$\Lsh$} with cfg} %
& 
& \cfgresult{0.0204} %
& \textbf{0.7987} %
& \cfgresult{0.5256} %
& \cfgresult{0.4041} %
\\

Ours LSM-L
& 473M
& \textbf{0.0190} %
& 0.7841 %
& 0.5011 %
& 0.3864 %
\\
\cfgresult{\rotatebox[origin=c]{180}{$\Lsh$} with cfg} %
& 
& \cfgresult{0.0192} %
& \cfgresult{0.7937} %
& \cfgresult{0.5537} %
& \cfgresult{0.4277} %
\\

\bottomrule
\end{tabular}

\end{table}

\begin{table}[t]
\centering
\newcommand{\cmark}{\ding{51}}%
\newcommand{\xmark}{\ding{55}}%
\footnotesize
\setlength{\tabcolsep}{3pt}
\caption{\textbf{Training-free deterministic conditioning.} Localization performance on SEVIR improves when introducing deterministic priors while distribution coverage deteriorates.  $i=1$ corresponds to a fully deterministic and $i=0$ to a fully generative prediction. }
\label{tab:detcond}
\vspace{-0.2cm}
\begin{tabular}{l ccccc}
\toprule
Model  & det. prior & $i$  & CRPS $\downarrow$  &  HSS $\uparrow$ & CSI-M $\uparrow$\\
\midrule
\color{gray} Earthformer &  \color{gray} na & \color{gray} na & \color{gray} 0.0251 & \color{gray} 0.5411 & \color{gray} 0.4310 \\
CasCast
    & \cmark & na  & 0.0202  & 0.5602    & 0.4401 \\
\arrayrulecolor{gray!50!white}
\midrule
\arrayrulecolor{black}
\multirow{4}{*}{Ours}
    & \xmark & 0  & \textbf{0.0190}  & 0.5011	& 0.3864 \\

    & \cmark  & 0.2 & 0.0198  & 0.5714 & 0.4444 \\

    & \cmark  & 0.5 & 0.0219 & \textbf{0.5735} &\textbf{ 0.4455} \\
    & \cmark & 0.75 & 0.0233 & 0.5598 & 0.4338 \\
\bottomrule
\end{tabular}
\end{table}

\paragraph{Deterministic Prior}
We achieve top-3 CSI and 2nd-best HSS without deterministic priors. However, our training paradigm allows zero-shot deterministic conditioning by initializing future frames with noisy Earthformer~\cite{gao2022earthformer} predictions, where the noise level allows to express confidence in the deterministic prediction. Conditioning on these reduces diversity by lowering conditional variance, which improves localization of mean predictions if correct but collapses the distribution to incorrect modes otherwise.
\cref{tab:detcond} confirms this trade-off: relying on deterministic predictions allows better localization but worse coverage.

\subsection{Ablations}
\label{sec:ablations}

\paragraph{Regularization}
As shown in \cref{tab:first_stage_comp}, \textit{T-reg.} FREUD delivers the best reconstruction quality and the most meaningful uncertainty estimates (\cref{fig:extreme_v_var_main}). \textit{T-reg.} also produces a more compact and structured latent space (see Appendix), whereas other variants exhibit higher variance and lower density. \cref{tab:comp_reg} shows that this improved structure translates into better downstream performance: B-LSMs trained on \textit{T-reg.} latents achieve superior CRPS and SSIM compared to those trained on other latent spaces.

\begin{table}[tb]
    \centering
    \caption{\textbf{Comparison of B-LSMs} in our regularized latent spaces. \textit{T-reg.}\thinspace outperforms \textit{unreg.}\thinspace and \textit{KL-reg.}\thinspace in terms of CRPS and SSIM but performs worse for localization metrics.}
    \vspace{-0.2cm}
    \footnotesize
    \label{tab:comp_reg}
    \setlength{\tabcolsep}{6pt}
    \centering
    \begin{tabular}{lcccc}
    \toprule
    \textbf{Regularization} & \textbf{CRPS}$\downarrow$ & \textbf{SSIM}$\uparrow$ & \textbf{HSS}$\uparrow$ & \textbf{CSI}$\uparrow$ \\
    \midrule
    Unreg.   & 0.0222 & 0.7630 & \textbf{0.5139} & \textbf{0.3956} \\
    KL-reg.  & 0.0201 & 0.7790 & 0.4993 & 0.3892  \\
    T-reg.   & \textbf{0.0196} & \textbf{0.7828} & 0.4923 & 0.3805 \\
    \bottomrule
    \end{tabular}
    \vspace{-0.4cm}
\end{table}

\paragraph{Joint Video Decoding}
We ablate the effect of united (joint) decoding in \cref{tab:first_stage_comp}. While a frame-wise DiffAE is competitive with the frame-wise CasCast autoencoder~\cite{gong2024cascast}, it performs noticeably worse than FREUD. Using the same regularization settings, FREUD reduces dMAE by 33\%, indicating improved temporal consistency. Qualitative video samples in the supplementary material further confirm this effect: frame-wise decoding exhibits small fluctuations and flickering artifacts, whereas united decoding produces smoother and more temporally stable reconstructions.

\section{Conclusion}
\label{sec:conclusion}
We introduce FREUD, a simple and scalable first-stage compression framework. FREUD is built on a frame-wise transformer encoder and a probabilistic rectified-flow video decoder. This design eliminates the need for perceptual or adversarial losses, produces sharper and more stable reconstructions, and, most importantly, allows for direct uncertainty estimation at the decoding stage. When combined with a latent-space rectified-flow transformer, our pipeline achieves state-of-the-art precipitation nowcasting on SEVIR. It offers strong calibration, flexible conditioning, and favorable scaling with model size, ensemble size, and test-time computation. By capturing first- and second-stage uncertainties while remaining purely data-driven, our approach paves the way for reliable short-term precipitation forecasting, without relying on deterministic physics-based predictors. We discuss remaining limitations and potential societal impacts in the Appendix.

\onecolumn
\twocolumn
\section*{Acknowledgments}
We would like to thank Carla Sagebiel for proposing the framework name, Dominik Lorenz for initiating the project, Owen Vincent for the technical support, and Marius Jacobs \& Nadja Sauter for their preliminary experiments. 
This project has been supported by the bidt project KLIMA-MEMES, the Horizon Europe project ELLIOT (GA No.\ 101214398), the project ``GeniusRobot'' (01IS24083) funded by the Federal Ministry of Research, Technology and Space (BMFTR), the BMWE ZIM-project (No.\ KK5785001LO4) ``conIDitional LoRA'', and the German Federal Ministry for Economic Affairs and Energy within the project ``NXT GEN AI METHODS - Generative Methoden für Perzeption, Prädiktion und Planung''. The authors gratefully acknowledge the Gauss Center for Supercomputing for providing compute through the NIC on JUWELS/JUPITER at JSC and the HPC resources supplied by the NHR@FAU Erlangen.

\section*{Author Contributions}

JS and JW jointly lead the project, conceptualized the core methodology, and preprocessed the data.
JS conceived the initial idea and devised initial prototypes.
JW developed, optimized, and evaluated the final models.
NS provided an initial implementation of Diffusion Autoencoders and contributed to figure design.
TP conceptualized the stochastic $\tanh$ regularization method and conducted initial comparisons of regularization.
All authors contributed to writing.
BO supervised the project and reviewed the manuscript.

{
    \small
    \bibliographystyle{ieeenat_fullname}
    \bibliography{main}
}´

\appendix
\clearpage
\setcounter{page}{1}
\maketitlesupplementary

\setcounter{figure}{0}
\renewcommand{\thefigure}{S\arabic{figure}}

\section{Extended Results}\label{sec:add_experiments}
\subsection{Extreme Precipitation}

We further evaluate our method under extreme weather conditions. We define extreme weather as the top $20\,\%$ of events with the highest average precipitation, and the rest as ``normal''.

\paragraph{Forecasting Performance} We provide a quantitative comparison between our method and the state-of-the-art CasCast method~\cite{gong2024cascast} in extreme weather in \cref{tab:metrics_extreme}.
Under extreme conditions, the CRPS and SSIM deteriorate for all models, indicating that prediction is more difficult in chaotic extreme events.
Interestingly, localization-centric metrics improve for all methods. Therefore, determining the movement of extreme patterns seems to be easier than predicting the onset of light to medium rain. Notably, the performance gap for these metrics between our method and CasCast further narrows. Hence, our approach achieves comparable localization performance to state-of-the-art methods in critical extreme weather scenarios while maintaining superior distribution coverage. 
Furthermore, the scaling properties discussed in \cref{sec:ablations} hold under extreme conditions. The L-LSM achieves the best CRPS and SSIM scores, and HSS and CSI match the results from CasCast~\cite{gong2024cascast} closely with classifier-free guidance.
Therefore, our L variant is particularly well-suited to determine localized risk at improved distribution coverage compared to prior methods.

\begin{table}[b]
\centering
\caption{Comparison of CasCast~\cite{gong2024cascast} with our method for extreme weather events. Extreme weather is defined as the 80th percentile of all events with the highest precipitation.}\label{tab:metrics_extreme}
\footnotesize
\setlength{\tabcolsep}{8pt}
\newcommand{\cfgresult}[1]{\textcolor{black!60}{#1}}
\centering
\begin{tabular}{lcccc} 
\toprule
\textbf{Model} & \textbf{CRPS}$\downarrow$ & \textbf{SSIM}$\uparrow$ & \textbf{HSS}$\uparrow$ & \textbf{CSI}$\uparrow$ \\
\midrule
CasCast~\cite{gong2024cascast}
& 0.0404
& 0.6354
& \textbf{0.5726}
& \textbf{0.4728}
\\
Our B-LSM 
& 0.0384
& 0.6514
& 0.4982
& 0.4162 
\\
\cfgresult{\rotatebox[origin=c]{180}{$\Lsh$} with cfg} %
& \cfgresult{0.0400}
& \cfgresult{0.6634}
& \cfgresult{0.5321}
& \cfgresult{0.4678}
\\
Our L-LSM 
& \textbf{0.0357}
& 0.6602
& 0.5107
& 0.4233 
\\
\cfgresult{\rotatebox[origin=c]{180}{$\Lsh$} with cfg} %
& \cfgresult{0.0360}
& \textbf{0.6731}
& \cfgresult{0.5659}
& \cfgresult{0.4677}
\\
\bottomrule
\end{tabular}

\end{table}

\begin{figure}[tbh]
    \centering

    \begin{subfigure}[c]{0.3\textwidth}
        \centering
        \includegraphics[width=\linewidth]{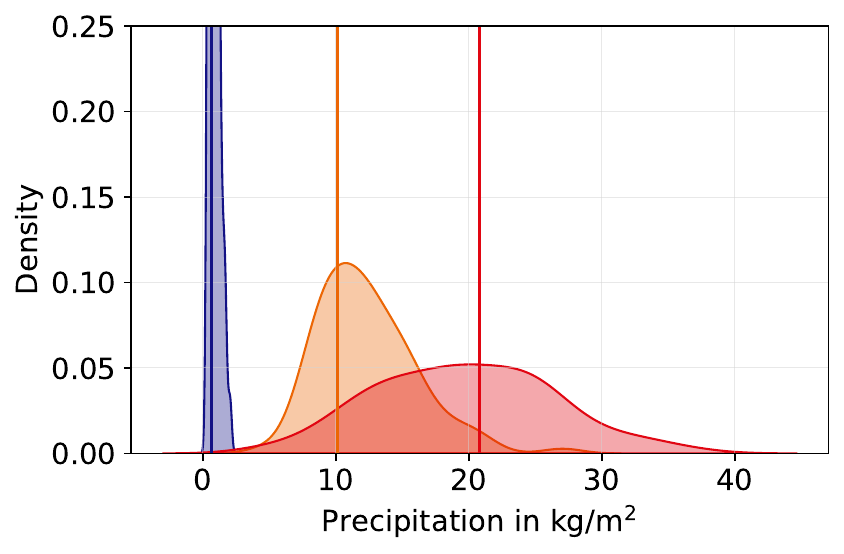}
    \end{subfigure}
    \hspace{3cm}
    \begin{subfigure}[c]{0.05\textwidth}
        \centering
        \begin{tabular}{c}
             \includegraphics[width=\linewidth]{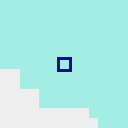}\\
             \includegraphics[width=\linewidth]{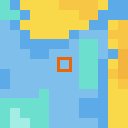}\\
             \includegraphics[width=\linewidth]{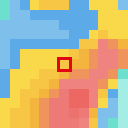}
        \end{tabular}
    \end{subfigure}
    \hspace{1em}
    \begin{subfigure}[c]{0.18\textwidth}
        \centering
        \includegraphics[width=\linewidth]{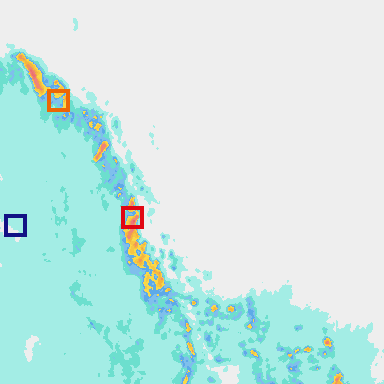}
    \end{subfigure}
    \caption{Forecasting ensemble distributions for three pixels of low, medium, and high precipitation regions. Vertical lines show ground truth precipitation. The more extreme the weather, the larger the distribution spread, indicating higher uncertainty.}
    \label{fig:ensemble_extreme_normal}
\end{figure}

\paragraph{Reconstruction Performance}
In addition to the regression experiment in \cref{fig:extreme_v_var_main} in the main paper, we quantitatively analyze the differences between reconstruction ensemble members for extreme and normal weather in \cref{tab:var_extreme_weather}.
We find $2.4 \times$ higher variance with extreme weather, and intra-ensemble differences, as measured by RMSE, are $1.5 \times$ higher for extreme weather.
This result confirms that reconstructions of chaotic extreme weather events show larger ensemble differences, indicating higher uncertainty.

\begin{table}[tbh]
    \centering
    \caption{Differences over reconstructions from a 10-member \textit{T-reg.}\thinspace FREUD ensemble. Extreme weather events are defined as the 80th percentile of all events in the validation dataset with the highest average precipitation.}
    \label{tab:var_extreme_weather}
    \footnotesize
\setlength{\tabcolsep}{3pt}
\centering
\begin{tabularx}{\linewidth}{l c c c c}
    \toprule
    \textbf{Weather Data}
    & \makecell{\textbf{Var} {\tiny $\times 10^{-5}$}}
    & \makecell{\textbf{SD} {\tiny $\times 10^{-3}$}}
    & \makecell{\textbf{RMSE} {\tiny $\times 10^{-2}$}}
    & \makecell{\textbf{MAE} {\tiny $\times 10^{-3}$}} \\
    \midrule
    
    Extreme %
    & 6.866     %
    & 5.066      %
    & 1.172     %
    & 5.737  \\ %
    
    Normal  %
    & 2.849     %
    & 2.003      %
    & 0.769    %
    & 2.884 \\ %
    \bottomrule
\end{tabularx}

\end{table}

\paragraph{Extreme Weather Phenomena}

\begin{figure*}[h]
    \centering \small
    \setlength\tabcolsep{1pt}

    \newcommand{\imagepng}[3]{
        \includegraphics[width=0.13\linewidth]{fig/lsm/forecast/hurricane_2/#1_#2_#3.png}
    }

    \newcommand{\rowpng}[2]{
        \imagepng{#1}{#2}{1} & %
        \imagepng{#1}{#2}{3} & %
        \imagepng{#1}{#2}{5} & %
        \imagepng{#1}{#2}{7} & %
        \imagepng{#1}{#2}{9} & %
        \imagepng{#1}{#2}{11} %
    }

    \begin{minipage}{0.9\textwidth}
         \begin{subfigure}{1\textwidth}
            \centering
            \begin{tabular}{lcccccc}
                & $+10$\,min & $+20$\,min & $+30$\,min & $+40$\,min & $+50$\,min & $+60$\,min \\
                \rotatebox{90}{\hspace{1.8em} E1} & \rowpng{ens}{0} \\
                \rotatebox{90}{\hspace{1.8em} E2} & \rowpng{ens}{1} \\
                \arrayrulecolor{gray!50!white} \cmidrule{1-7} \arrayrulecolor{black} \\
                \rotatebox{90}{\hspace{1.75em} GT} & \rowpng{gt}{0} \\
            \end{tabular}
        \end{subfigure}
    \end{minipage}
    \hspace{-0.8cm}
    \begin{minipage}{0.09\textwidth}
        \centering
        \includegraphics[width=\linewidth]{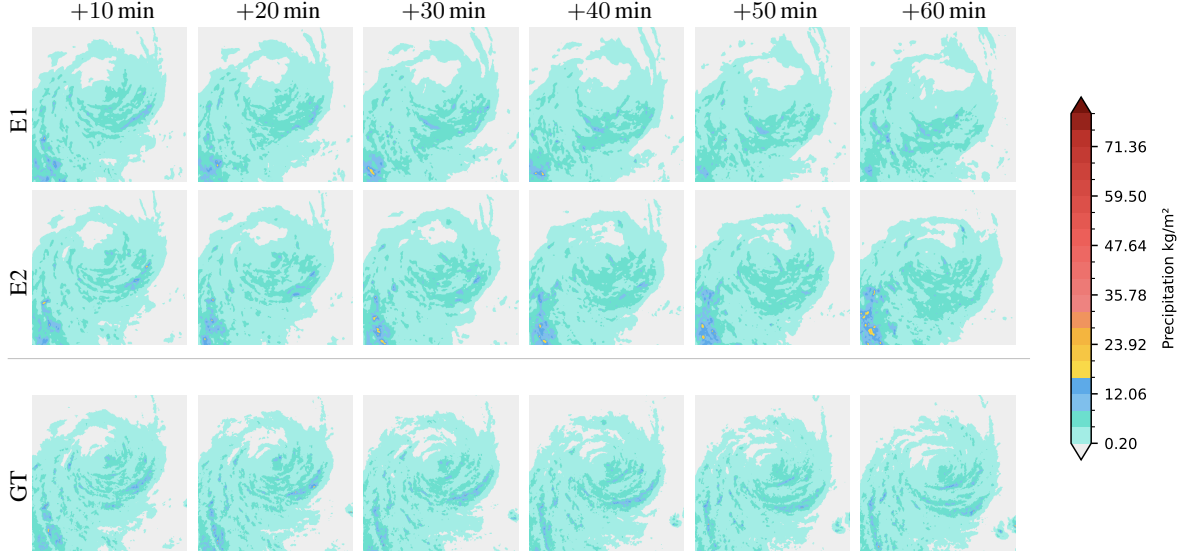}
    \end{minipage}
    \caption{Qualitative sample from a known hurricane event. Our method is able to capture the circular motion observed in the ground truth. For a better visualization of circular motion, please refer to the supplemented video visualizations.}
    \label{fig:lsm:qualitative_hurricane_2}
\end{figure*}

We further assess how well our method generalizes to rare and extreme weather phenomena by evaluating the few labeled severe events in the SEVIR train and test sets. As shown in \cref{tab:sup:rare-events}, our approach consistently outperforms CasCast across tornado, flood, and flash-flood cases, demonstrating clear advantages. Beyond these quantitative results, we show a qualitative example in \cref{fig:lsm:qualitative_hurricane_2}, which shows that our model can capture the large-scale rotational dynamics and global motion patterns characteristic of a hurricane. We further provide video visualizations that illustrate the circular motion more clearly.

\begin{table*}[t]
\centering
\footnotesize
\caption{Reconstruction and forecasting performance on rare and severe SEVIR weather events. We compare CasCast~\cite{gong2024cascast} with our method across reconstruction metrics (RMSE, PSNR, SSIM) and forecasting metrics (CRPS, HSS) for tornado, flood, and flash-flood subsets.
}
\label{tab:sup:rare-events}
\begin{tabular}{l l l l l l l}
\toprule
& & \multicolumn{3}{c}{\textbf{Reconstruction}} & \multicolumn{2}{c}{\textbf{Forecast}}\\
\cmidrule(lr){3-5} \cmidrule(lr){6-7}
\textbf{Event} & \textbf{Model} & RMSE & PSNR & SSIM & CRPS & HSS \\
\midrule
Tornado (194 events) & CasCast 
    & 0.104 
    & 22.800 
    & 0.957 
    & 0.0627 
    & 0.4675 \\
& \textbf{Ours} 
    & 0.0120 $\color{green}{-0.092}$ 
    & 38.434 $\color{green}{+15.634}$ 
    & 0.970 $\color{green}{+0.013}$ 
    & 0.0374 $\color{green}{-0.0253}$ 
    & 0.4882 $\color{green}{+0.0207}$ \\
\arrayrulecolor{gray!50!white}
\midrule
\arrayrulecolor{black}
Flood (177 events) & CasCast 
    & 0.080 
    & 25.316 
    & 0.960 
    & 0.0488 
    & 0.4493 \\
& \textbf{Ours} 
    & 0.011 $\color{green}{-0.069}$ 
    & 39.236 $\color{green}{+13.920}$ 
    & 0.971 $\color{green}{+0.011}$ 
    & 0.0280 $\color{green}{-0.0208}$ 
    & 0.4106 $\color{red}{-0.0387}$ \\
\arrayrulecolor{gray!50!white}
\midrule
\arrayrulecolor{black}
Flash Flood (385 events) & CasCast 
    & 0.072 
    & 25.619 
    & 0.961 
    & 0.0545 
    & 0.4660 \\
& \textbf{Ours} 
    & 0.012 $\color{green}{-0.060}$ 
    & 38.378 $\color{green}{+12.759}$ 
    & 0.971 $\color{green}{+0.010}$ 
    & 0.0307 $\color{green}{-0.0238}$ 
    & 0.5140 $\color{green}{+0.048}$ \\
\bottomrule
\end{tabular}
\end{table*}

\subsection{Ensemble Distributions}
\begin{figure*}[tb]
    \centering
    \begin{subfigure}[t]{0.45\textwidth}
        \centering
        \includegraphics[width=\linewidth]{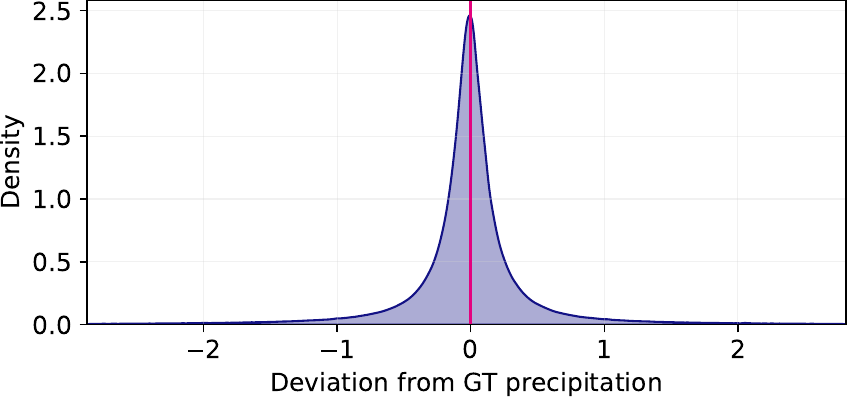}
        \caption{FREUD}
        \label{fig:freud:deviation_dist}
    \end{subfigure}
    \hfill
    \begin{subfigure}[t]{0.45\textwidth}
        \centering
        \includegraphics[width=\linewidth]{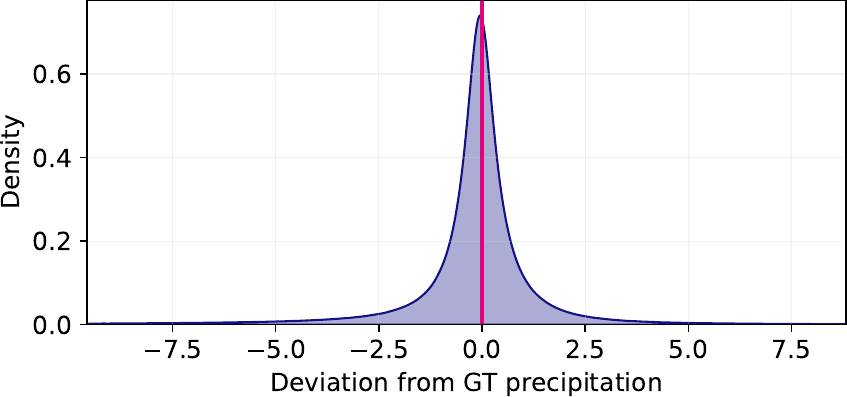}
        \caption{B-LSM}
        \label{fig:lsm:deviation_dist}
    \end{subfigure}
    \caption{Deviations from ground truth precipitation, whereby 0 denotes a perfect forecast. Only pixels with more than $1\,kg/m^2$ precipitation are considered. The x-axis range indicates three standard deviations from the mean to remove outliers. Both distributions are almost zero-centered and show similar likelihoods of over- and underestimating precipitation.}
    \label{fig:deviation_dist}
\end{figure*}

\paragraph{Error Distributions}
\cref{fig:deviation_dist} shows the distribution of deviations from ground-truth precipitation, for FREUD and our forecasting model (B-LSM), where 0 denotes a perfect forecast. We only consider pixels where precipitation is observed. Both produce nearly zero-centered error distributions with similar likelihoods of over- and underestimation. The figure reflects a slight tendency to underestimate high-precipitation regions, consistent with average weather conditions; however, it overall indicates that ensemble members are well-spread around the ground truth.

\begin{figure}[tb]
    \centering
    \includegraphics[width=0.98\linewidth]{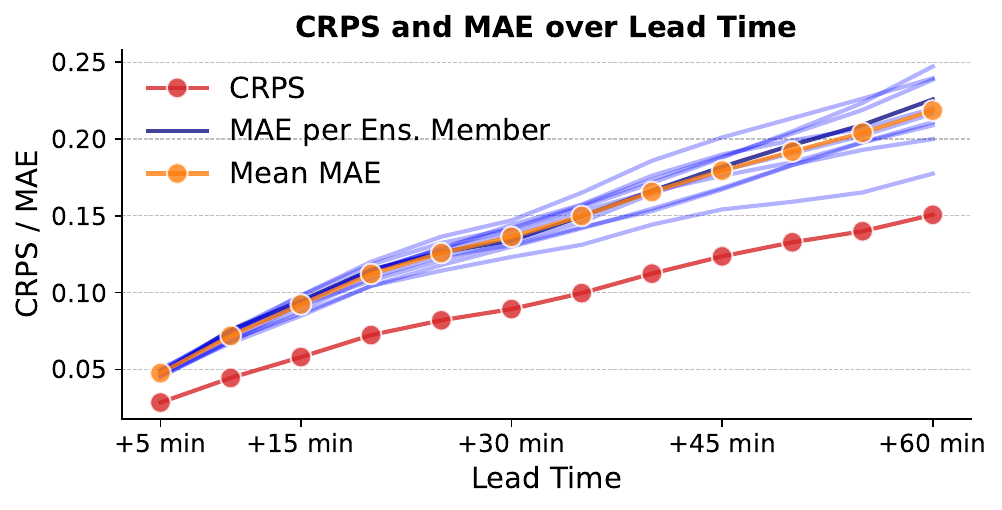}
    \caption{Forecast skill of the individual ensemble members and of their aggregated prediction for the example in \cref{fig:lsm:qualitative}. Each member constitutes a plausible realisation of the future weather, yet the expectation over ensemble members attains lower error.}
    \label{fig:per_ensemble_v_ensemble}
\end{figure}

\paragraph{Ensemble Performance}
\cref{fig:per_ensemble_v_ensemble} shows the per ensemble member MAE as well as the corresponding ensemble CRPS and MAE of the mean over ensemble members for the qualitative sample of \cref{fig:lsm:qualitative} in the main paper. 
Across all lead times, the ensemble consistently outperforms the average performance of its individual members. This indicates that individual errors tend to spread around a common mean, and discrepancies cancel out when aggregated, yielding a mean forecast that closely aligns with the ground truth. This behavior is consistent with the error distributions discussed earlier.

\paragraph{Pixel Value Distributions}
\cref{fig:ensemble_extreme_normal} shows the distribution of LSM ensemble forecasts for pixels in low, medium, and high precipitation regimes. The mode shifts with the true intensity, and the variance increases for heavier rainfall, reflecting higher uncertainty in chaotic, high-precipitation regions. All regimes exhibit a long tail toward larger values, capturing the possibility of intensifying rain; this tail becomes more pronounced for high precipitation. In this regime, however, the ensemble mean tends to underestimate the true value, consistent with climatology, where a decrease in intensity is more common than further escalation. Overall, the ensemble behavior aligns with known precipitation dynamics, indicating that our model has learned realistic weather statistics.

\subsection{Performance on MeteoNet}

\begin{figure}
    \centering
    \includegraphics[width=0.5\linewidth]{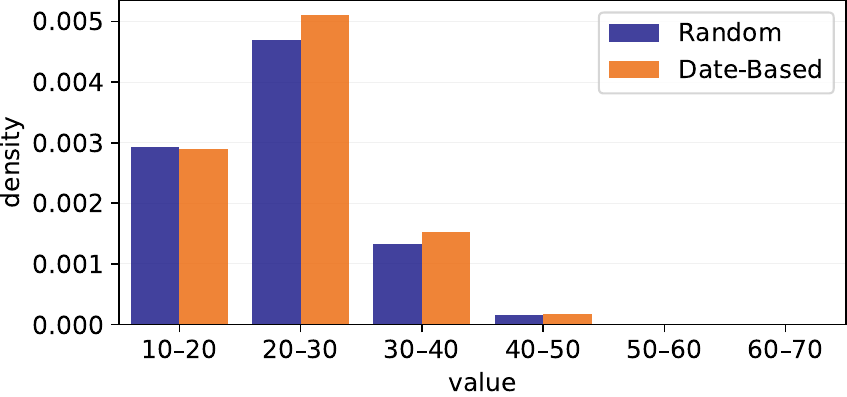}
    \vspace{-3mm}
    \caption{Training data distribution of MeteoNet splits.}
    \label{fig:meteonet_split_hist}
\vspace{-5mm}
\end{figure}

In addition to the results obtained for the SEVIR~\cite{veillette2020SEVIR} benchmark, we validate our model's applicability to other datasets by applying our model to the MeteoNet~\cite{larvor2020meteonet} benchmark.
\cref{tab:meteonet_results} shows that our model performs similarly on MeteoNet and SEVIR. Further, our model outperforms all baselines but CasCast in terms of CRPS and achieves comparable CSI.
However, we find that results on MeteoNet are highly sensitive to the chosen train–test split.
\cref{fig:meteonet_split_hist} shows statistical differences between MeteoNet training splits.  While differences are small, the date-based split shows a slightly higher mean, resulting in lower CSI for the random split as true negatives have no effect on CSI computation.
Since CasCast~\cite{gong2024cascast} provides limited details on their experimental setup and no public MeteoNet checkpoint, we cannot fully validate comparability under their protocol. For this reason, we focus our main analysis on SEVIR.

\begin{table}[tbh]
    \centering
    \caption{\textbf{Performance on MeteoNet:} Our approach achieves competitive performance on MeteoNet and performs similarly to the model trained on the SEVIR dataset. Yet, we observe a strong influence of the train-test split on downstream performance.}
    \label{tab:meteonet_results}
    \adjustbox{width=\columnwidth}{
        \footnotesize
\setlength{\tabcolsep}{5.1pt}
\centering
\newcommand{\cfgresult}[1]{\textcolor{black!60}{#1}}
\begin{tabular}{
llccccc
} 

\toprule

\textbf{Method} &
\textbf{Split} &
\textbf{CRPS}$\downarrow$ &
\textbf{SSIM}$\uparrow$ &
\textbf{HSS}$\uparrow$ &
\textbf{CSI}$\uparrow$
\\

\midrule

EarthFormer~\cite{gao2022earthformer}~\ven{NeurIPS '22}
& unknown
& 0.0224
& --
& --
& 0.2831
\\

NowcastNet~\cite{zhang2023skilful}~\ven{Nature '23}
& unknown
& 0.0277
& --
& --
& 0.2955
\\

PreDiff~\cite{gao2023prediff}~\ven{NeurIPS '23}
& unknown
& 0.0197
& --
& --
& 0.2546
\\

CasCast~\cite{gong2024cascast}~\ven{ICML '24}
& unknown
& 0.0180 %
& -- %
& -- %
& 0.3156 %
\\

\arrayrulecolor{gray!50!white}
\midrule
\arrayrulecolor{black}

FREUD + LSM-L (\textit{ours})
& Random
& 0.0224 %
& 0.7212 %
& 0.0876 %
& 0.1117 %
\\
\cfgresult{\rotatebox[origin=c]{180}{$\Lsh$} with cfg} %
& 
& \cfgresult{0.0231} %
& \cfgresult{0.7133} %
& \cfgresult{0.1368} %
& \cfgresult{0.0876} %
\\

FREUD + LSM-L (\textit{ours})
& Date-based
& 0.0193 %
& 0.7312 %
& 0.2082 %
& 0.1417 %
\\
\cfgresult{\rotatebox[origin=c]{180}{$\Lsh$} with cfg} %
& 
& \cfgresult{0.0194} %
& \cfgresult{0.7405} %
& \cfgresult{0.3150} %
& \cfgresult{0.2191} %
\\

\bottomrule
\end{tabular}

    }
\end{table}

\subsection{Effect of Classifier-free Guidance}
\label{sec:cfg_ablation}

\begin{figure}[tbh]
    \centering
    \begin{subfigure}{0.47\linewidth}
        \centering
        \includegraphics[width=\linewidth]{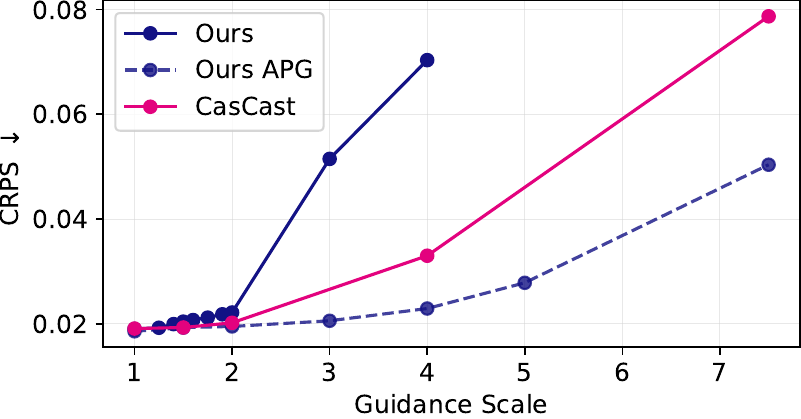}
        \caption{CRPS}
    \end{subfigure}
    \hspace{0.1cm}
    \begin{subfigure}{0.47\linewidth}
        \centering
        \includegraphics[width=\linewidth]{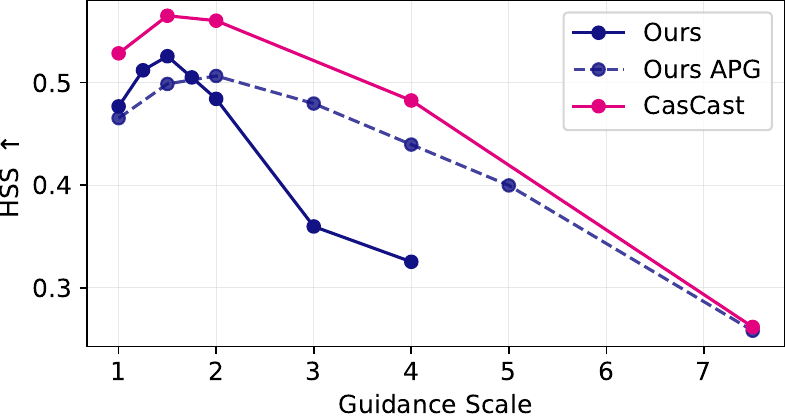}
        \caption{HSS}
    \end{subfigure}
    \caption{
    Performance of our B-LSM and CasCast with increasing CFG~\cite{ho2022ClassifierFree} and APG~\cite{sadat2024Eliminating} guidance strength. For all approaches, CRPS continuously worsens while HSS improves up to an optimal value and deteriorates afterwards.
    }
    \label{fig:lsm:cfg_performance}
\end{figure}

\paragraph{Impact on Performance}
We analyze the impact of classifier-free guidance (CFG)~\cite{ho2022ClassifierFree} on forecasting performance in~\cref{fig:lsm:cfg_performance}, comparing our B-LSM to the state-of-the-art CasCast method~\cite{gong2024cascast}.
As guidance strength increases, CRPS rises monotonically, reflecting reduced distributional coverage, consistent with observations in image generation~\cite{nichol2021improved, ho2022ClassifierFree}. HSS, however, improves up to an optimal guidance level ($1.5$ for our model) before declining. We observe the same behavior with Adaptive Projected Guidance (APG)~\cite{sadat2024Eliminating}, despite its design to counteract over-saturation in image synthesis at high guidance strengths.

\paragraph{Qualitative Effect}
We visualize two qualitative samples at different guidance scales with our method (\cref{fig:lsm:cfg_qual_1} and \cref{fig:lsm:cfg_qual_2}) and CasCast (\cref{fig:cascast:qual_1} and \cref{fig:cascast:qual_2}).
For both methods, we observe that the extremeness of weather conditions increases with higher guidance. Samples degenerate into unrealistic extreme weather beyond a method-specific threshold. As APG does not solve this problem, we hypothesize that this behavior is distinct from over-saturation.
Analyzing unconditional samples provided in \cref{fig:lsm:unconditional_qualitative}, we find notably low precipitation intensity, as low precipitation is more common than strong rain. Hence, by applying guidance, we push samples from low to high precipitation.

\begin{figure*}[tb]
    \centering
    \begin{subfigure}{0.4\linewidth}
        \centering
        \includegraphics[width=\linewidth]{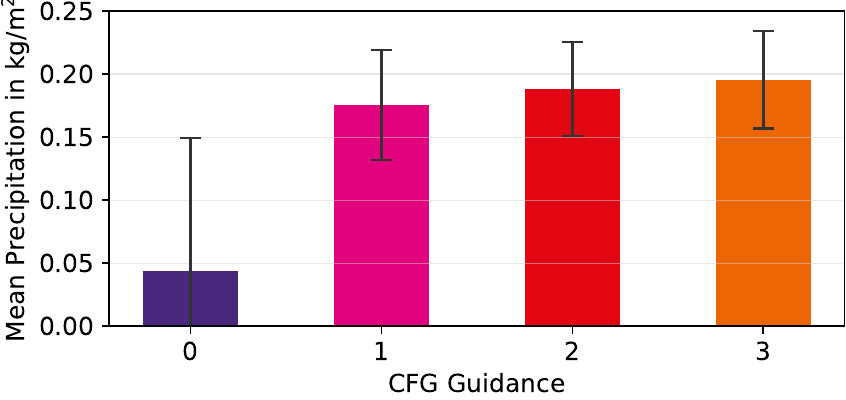}
        \caption{Ours}
    \end{subfigure}
    \hspace{1cm}
    \begin{subfigure}{0.4\linewidth}
        \centering
        \includegraphics[width=\linewidth]{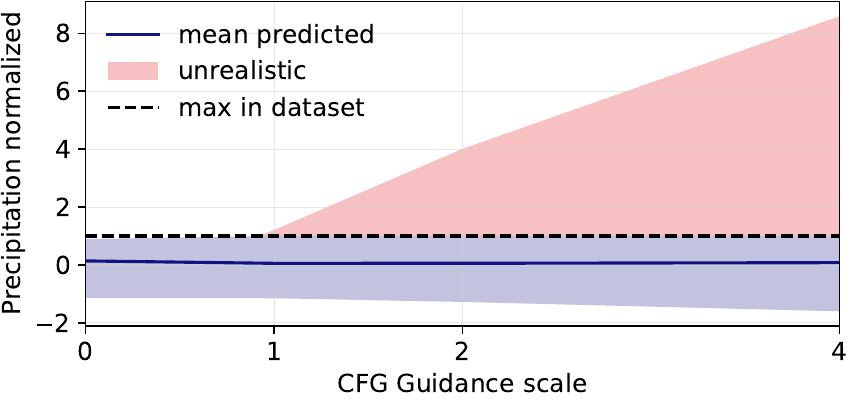}
        \caption{CasCast}
    \end{subfigure}
    \caption{
    The effect of cfg~\cite{ho2022ClassifierFree} on our method and CasCast~\cite{gong2024cascast}. For our approach (a), the mean precipitation consistently increases while the ensemble variance (error bars) decreases. For CasCast (b), the mean remains largely unaffected, but the min-max range of predicted values (shaded area) explodes. We keep the encoded $[0, 1]$ range in the right plot instead of applying the non-linear mapping to precipitation for visualization.
    }
    \label{fig:lsm:cfg_intensity}
\end{figure*}

\paragraph{Impact on Descriptive Statistics}
Inspired by these qualitative insights, we investigate the descriptive statistics of forecasts with different guidance scales in \cref{fig:lsm:cfg_intensity}. For our method, we find that guidance consistently increases the average precipitation intensity in forecasts.
For CasCast, we observe a different yet related behavior: while the mean intensity remains largely unaffected, extreme values increase dramatically, leading to unrealistically high precipitation that quickly exceeds the strongest precipitation found in the SEVIR dataset~\cite {veillette2020SEVIR}.
Therefore, both methods produce samples with unrealistically high precipitation with strong guidance, which is likely not an over-saturation artefact.

\paragraph{Unsuitability of CFG for Forecasting}
Based on our previous findings, we believe the problem with guidance in generative precipitation nowcasting is related to the dominance of low precipitation in the datasets. Generative models that learn the data distribution will tend to produce samples with low precipitation, resulting in unconditional samples with less intense precipitation than conditional samples.
Therefore, when applying classifier-free guidance, we push samples towards higher precipitation, which leads to the observed unrealistic samples.
Therefore, guidance is flawed in nowcasting, as the potential improved alignment with conditions is confounded by a distribution shift. Furthermore, guidance induces reduced distribution coverage and stronger overconfidence, which is problematic for uncertainty-aware forecasting of chaotic weather systems.
This result is not limited to our method and applies to state-of-the-art approaches, such as CasCast~\cite{gong2024cascast}, as well. Therefore, it is a general flaw of diffusion-based nowcasting. Notably, our method performs well without guidance, achieving superior distribution coverage and perceptual similarity at competitive localization compared to methods that rely on guidance (see \cref{tab:sota_sevir} in main paper).

\subsection{Further Ablations}
\begin{figure*}[tb]
    \centering
    \begin{subfigure}{0.24\textwidth}
        \includegraphics[width=\linewidth]{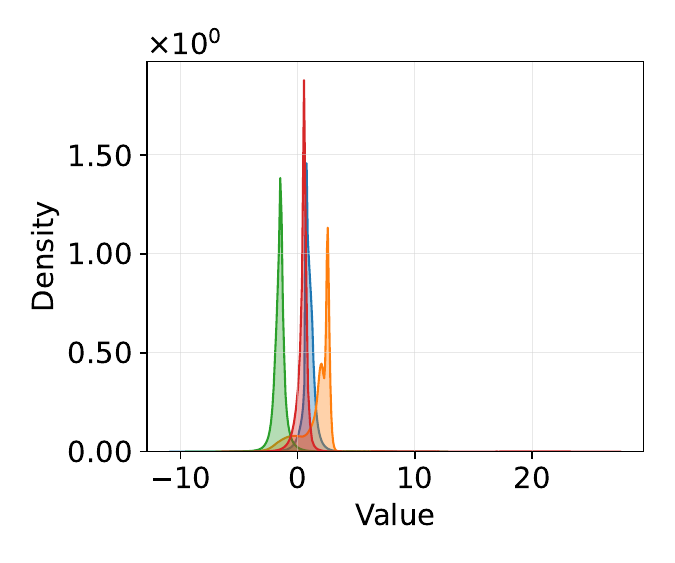}
        \caption{
            \centering
            CasCast\\
            $M = 0.367$,\\
            $\sigma = 1.481$
        }
    \end{subfigure}
    \begin{subfigure}{0.24\textwidth}
        \includegraphics[width=\linewidth]{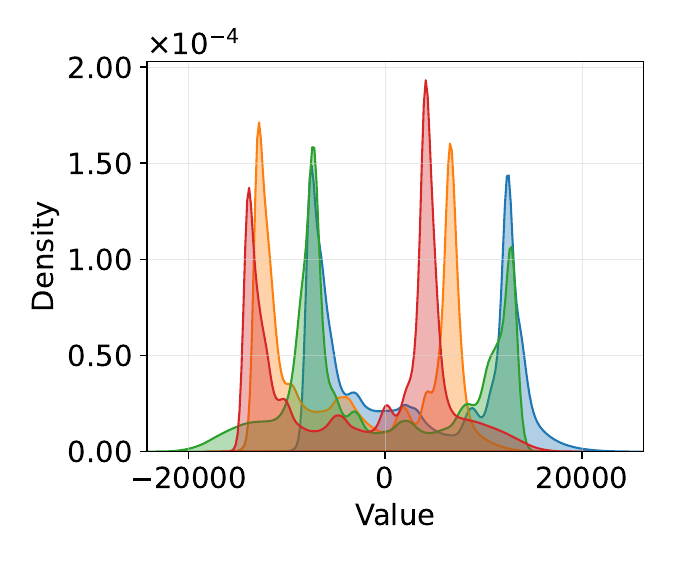}
        \caption{
          \centering
          Unreg. FREUD\\
          $M = -741.841$,\\
          $\sigma > 1 \times 10^3$
        }
    \end{subfigure}
    \begin{subfigure}{0.24\textwidth}
        \includegraphics[width=\linewidth]{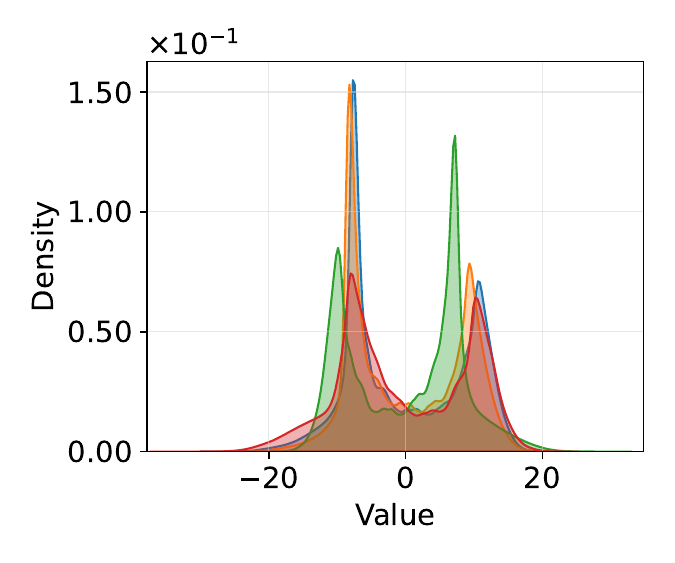}
        \caption{
            \centering
            KL-reg. FREUD\\
            $M = 0.040$,\\
            $\sigma = 8.779$
        }
    \end{subfigure}
    \begin{subfigure}{0.24\textwidth}
        \includegraphics[width=\linewidth]{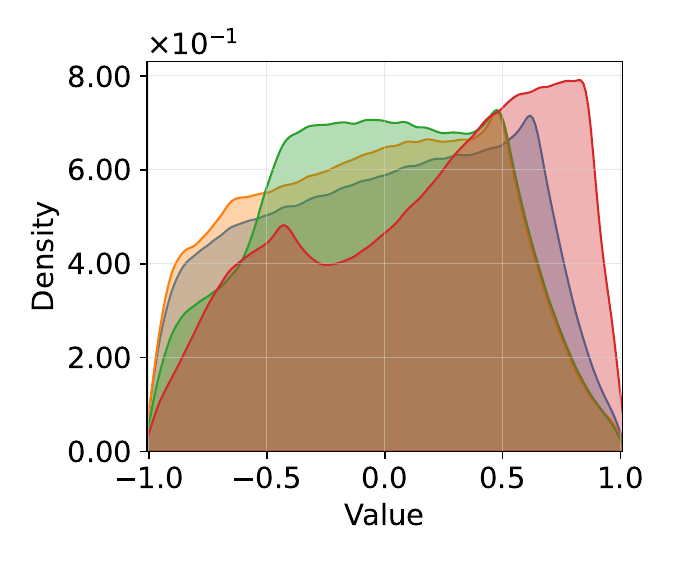}
        \caption{
            \centering
            T-reg. FREUD\\
            $M = 0.0261$,\\
            $\sigma = 0.499$
        }
    \end{subfigure}
    \caption{Latent space distributions for different regularization schemes and the CasCast encoder~\cite{gong2024cascast}. Colors indicate the latent space channel. The x-axis range denotes the min-max range of latent values.}
    \label{fig:freud:latent_space}
\end{figure*}

\paragraph{Latent Space Distributions}
\cref{fig:freud:latent_space} shows the density distribution of the latent space for all regularization variants. Without regularization, we observe high norms and variance, resulting in low density and long tails. Moreover, we observe a bimodal distribution for all latent channels.
The same bimodal pattern is observed with \textit{KL-reg.}\thinspace FREUD, but the variance is substantially reduced, leading to a higher density throughout the value range, and the distribution is almost zero-centered. However, we still observe long tails in the distributions.
Similarly, the CasCast latent value distribution exhibits reduced variance and an almost zero-centered latent space; however, we still observe very long tails due to the weak KL regularization.
In comparison, the \textit{T-reg.}\thinspace latent space exhibits the most zero-centered distribution with the lowest variance. We do not observe heavy tails and a less pronounced bimodal pattern; thus, the density remains high across the $[-1, 1]$ value range. This indicates that the \textit{T-reg.}\thinspace FREUD encoder produces a latent distribution that is easier to learn and sample for both the generative decoder and downstream latent space model (see \cref{tab:comp_reg} in main paper).

\begin{figure}[tb]
    \centering
    \includegraphics[width=0.9\linewidth]{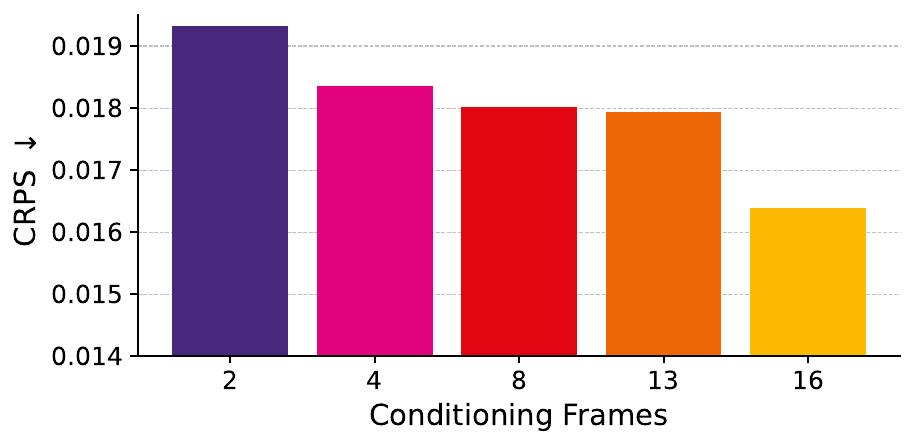}
    \caption{Performance of our forecasting pipeline for different numbers of conditioning frames.}
    \label{fig:lsm:n-cond-frames}
\end{figure}

\paragraph{Number of Conditioning Frames}
The masking-based training paradigm (RaMViD)~\cite{hoppe2022ramvid} allows variable conditioning frames during inference. We find improved performance when using more conditioning frames in \cref{fig:lsm:n-cond-frames}.

\paragraph{Full vs Factorized Attention}
We further train a latent space model with full attention over space and time, and compare it to our spatio-temporally factorized attention model in \cref{tab:full_attention}. Full attention provides slightly better performance, but its computational cost is prohibitive: The complexity of full self-attention is given by $(T \times H \times W)^2$, whereas the complexity of factorized attention is given as $(H \times W)^2 + T^2$. For our use case, we have $T = 25$, $H = 24$, and $W = 24$ after latent embedding and patching (see \cref{sec:implementation}), therefore, we require $207.4$ GFLOPs for full attention and $0.3$ GFLOPs for factorized attention. We believe that for operational nowcasting, such marginal improvements do not justify the substantial computational overhead; therefore, we only report results from the factorized model throughout the paper.

\begin{table}[t]
\centering
\caption{Comparison of forecasting skill with our B-LSM using factorized spatio-temporal attention and full self-attention. We find slightly improved performance with the full attention variant.}
\label{tab:full_attention}
\footnotesize
\setlength{\tabcolsep}{4pt}
\begin{tabular}{lcccc}
    \toprule
    \textbf{Attention} & \textbf{CRPS}$\downarrow$ & \textbf{SSIM}$\uparrow$ & \textbf{HSS}$\uparrow$ & \textbf{CSI}$\uparrow$ \\
    \midrule
    Factorized          %
    & 0.0196            %
    & 0.7828            %
    & 0.4923            %
    & 0.3805            %
    \\
    Full            %
    & 0.0187            %
    & 0.7897            %
    & 0.4968            %
    & 0.3848            %
    \\
    \bottomrule
\end{tabular}
\end{table}

\paragraph{Blob Toy Experiment}

\begin{figure}[tbh]
    \centering
    \includegraphics[width=0.9\linewidth]{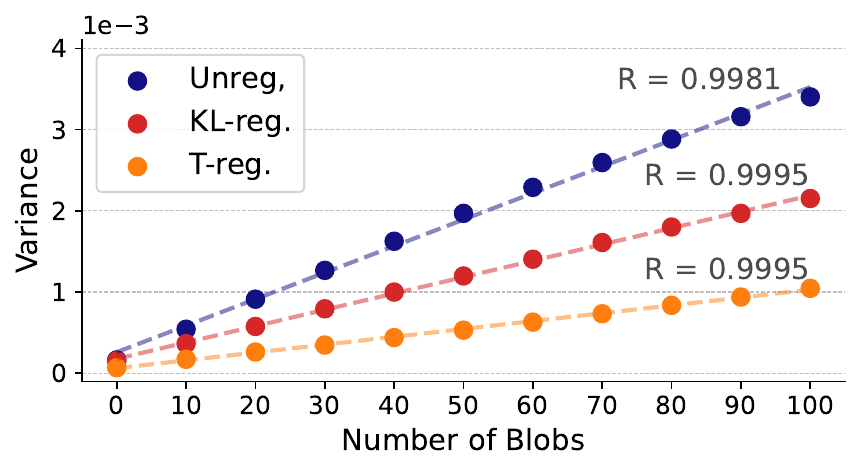}
    \caption{Intra-ensemble variance of first-stage reconstructions in the presence of increasing abnormal features (blobs). All linear regressions show significant correlations.}
    \label{fig:freud_var_v_blobs}
\end{figure}

We emulate an experiment from prior work~\cite{chan2024HyperDiffusion} to evaluate uncertainty quantification by inserting areas of unmoving extreme precipitation (blobs) into the data and assessing the variance of reconstruction ensembles.
\cref{fig:freud_var_v_blobs} shows a linear regression of the number of blobs against the ensemble variance. We find a significant linear correlation between ensemble variance and the blob count, confirming that reconstruction variance can detect deviations from the training distribution.

We show qualitative samples of a reconstruction ensemble with blobs together with corresponding variance maps in \cref{fig:freud:qualitative_blobs_42} and \cref{fig:freud:qualitative_blobs741}. The variance maps display an outline of high variance surrounding blobs and lower variance within them. Therefore, variance is high where abnormal blobs interact with normal precipitation, as it is unclear how the precipitation will alter the blob shape. However, in the blob center, pixels are surrounded by extreme precipitation, leading to confidence in their extreme values.
Due to the sharp outline, we can isolate abnormal patterns using variance maps. Therefore, the FREUD reconstruction ensembles can successfully detect and localize abnormal patches.

\paragraph{Scaling Test-time Compute}

\begin{figure}[tb]
    \centering
    \begin{subfigure}[t]{0.95\linewidth}
        \centering
        \includegraphics[width=\linewidth]{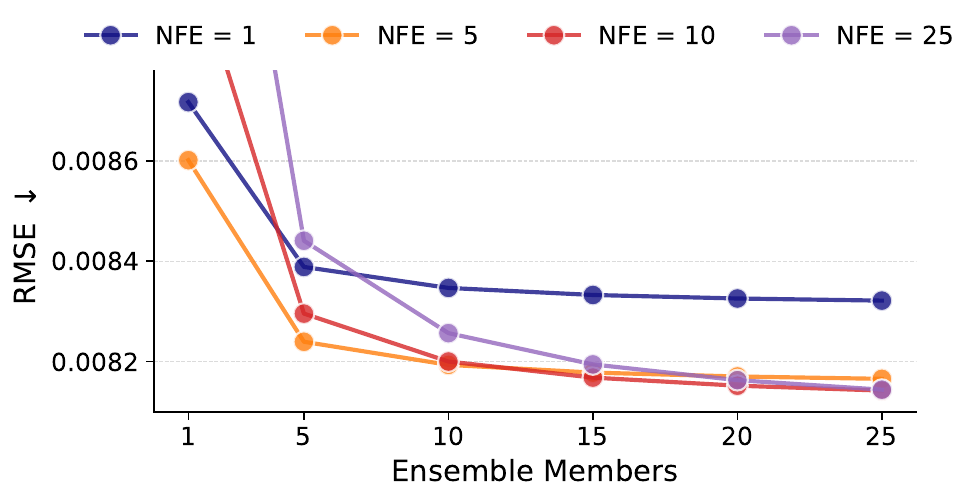}
        \caption{FREUD}
        \label{fig:freud:nfe-ens-rmse}
    \end{subfigure}
    \vfill
    \begin{subfigure}[t]{0.95\linewidth}
        \centering
        \includegraphics[width=\linewidth]{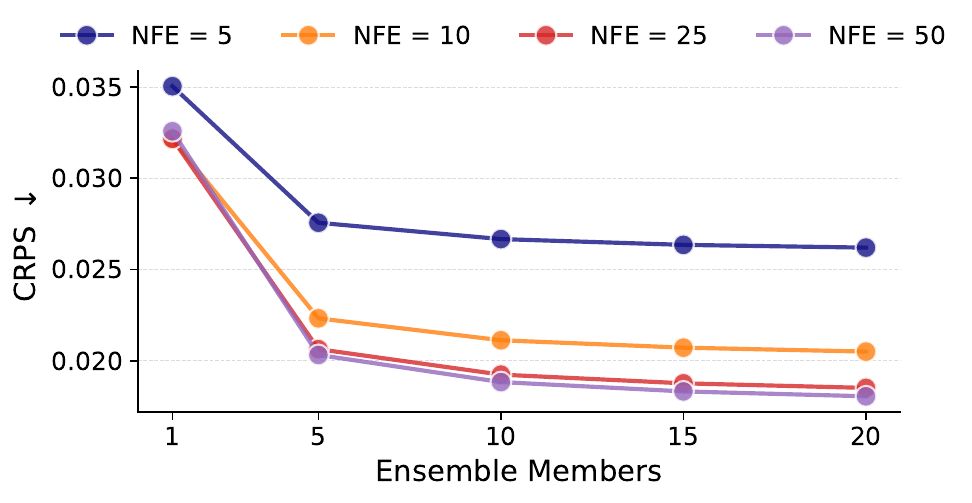}
        \caption{B-LSM}
        \label{fig:lsm:nfe-ens-crps}
    \end{subfigure}
    \caption{
        Ablation of scaling test time compute by increasing the number of function evaluations (NFE) and the ensemble size for the compression stage and B-LSM. Larger ensembles consistently improve the performance, and more function evaluations improve performance for the LSM.
    }
    \label{fig:nfe_ens_comp}
\end{figure}

\cref{fig:freud:nfe-ens-rmse} shows the influence of the ensemble size and the number of sampling steps for each ensemble member on the reconstruction performance of FREUD.
We observe a clear benefit from larger ensemble sizes, which plateau after 10 ensemble members.
The sampling steps do not show a consistent effect, and many-step runs are outperformed by few-step sampling for small ensemble sizes. Due to the strong conditioning from the encoder, the generative decoding task is simple, and the increased generative capabilities of expensive sampling runs are outweighed by integration errors, explaining the observed effect. Therefore, we can use efficient few-step sampling, which limits the overhead from using a generative decoder.
However, ensemble size and function evaluations seem to be entangled, as runs with many function evaluations benefit more from larger ensembles. We find a good trade-off with five function evaluations, which performs best up to 20 ensemble members.

Similarly, we evaluate the effect of ensemble size and sampling steps on the latent space model in \cref{fig:lsm:nfe-ens-crps}. CRPS consistently improves with a larger ensemble size and more function evaluations. Yet, the benefit of using more than 25 sampling steps and 15 ensemble members is marginal, revealing diminishing returns.
Still, these results highlight that we can improve forecasting performance by scaling test time compute, enabling flexibility in deployment depending on the requirements for accuracy and latency.

\begin{table}[t]
\centering
\footnotesize
\setlength{\tabcolsep}{3pt}
\caption{Inference time per decoder ensemble size on a single H200 GPU.}
\label{tab:decoderensemblingcost}
\begin{tabular}{l cccccc}
\toprule
Ensemble size &
    1     & 5       & 10     & 15    & 20   & 25 \\
Inference time (H200) &
    0.63s & 2.49s   & 4.83s  & 7.18s & 9.56s & 11.93s \\
\bottomrule
\end{tabular}
\end{table}

\paragraph{Ensembling inference cost}
Our method outperforms CasCast with a single decoder ensemble (cf. \cref{tab:sota_sevir}), yet \cref{fig:freud:nfe-ens-rmse} shows larger ensembles improve performance. Since decoding is highly parallelizable, ensembling incurs limited wallclock overhead.
\cref{tab:decoderensemblingcost} shows the evolution of wallclock latency with ensemble size. Even producing 25 decoder ensemble members requires less than 12\,s inference time, remaining compatible with operational 5\,min nowcasting.

\paragraph{Comparison with Diffusion Forcing}

\begin{table}[tb]
    \centering
    \footnotesize
    \caption{Comparison of RaMViD to Diffusion Forcing training. RaMViD training yields better downstream forecasting skill across all metrics.}
    \label{tab:diffusion_forcing}
    \begin{tabular}{lccccc}
        \toprule
         \textbf{Training Paradigm} & \textbf{CRPS} & \textbf{SSIM} & \textbf{HSS} & \textbf{CSI} \\
         \midrule
         \multirow{1}{*}{Diffusion Forcing} & 0.0218 & 0.7759 & 0.4627 & 0.3544 \\
         RaMViD & 0.0196 & 0.7828 & 0.4923 & 0.3805 \\
         \midrule
    \end{tabular}
\end{table}

Our masking-based training paradigm (RaMViD~\cite{hoppe2022ramvid}, see \cref{sec:latent_space_nowcasting}) can be interpreted as a special case of \textit{Diffusion Forcing}~\cite{chen2024diffusion}. In Diffusion Forcing, each frame is assigned an independent diffusion timestep during training. This offers more flexibility at inference time, as frames can be denoised using standard full-sequence, autoregressive, or rolling denoising. RaMViD can be interpreted as a variant of Diffusion Forcing, where during training, each frame is assigned either timestep $i = 1$ or timestep $i = \tau$, and during inference, each frame is initialized with a diffusion timestep $i = 1$ (data without noise) or $i = 0$ (pure noise). 

We implement a variant of our model trained using Diffusion Forcing. Since our use case is forecasting, we further bias the sampling of per-frame timesteps so that temporally later frames tend to be assigned lower timesteps. Thus, on average, early frames are less noisy than later frames. This resembles the forecasting task at inference time.
\cref{tab:diffusion_forcing} compares RaMViD inference with the more flexible Diffusion Forcing approach using a B-LSM. We find superior performance with our masking-based training compared to the Diffusion Forcing setup across all metrics.

\newpage

\section{Implementation Details}\label{sec:implementation}
In the following,€ we provide additional implementation details and hyperparameter settings. 

\subsection{Architecture}

\paragraph{FREUD} The FREUD encoder uses a $4 \times 4$ spatial patching followed by a downsampling layer, which is implemented as two transformer blocks with a hidden dimension of 96 and three attention heads, followed by a PixelShuffle~\cite{shi2016RealTime} operation to halve the resolution. Thus, we achieve a total of $8 \times$ downsampling along both spatial dimensions. The downsampling layers use neighborhood attention, where each pixel can only attend to a $7 \times 7$ spatial neighborhood. 2D Axial RoPE is used for positional embeddings.
The encoder processes the downsampled inputs with four additional transformer blocks~\cite{dosovitskiy2020vit} with 384-dimensional tokens, six attention heads, and full self-attention to produce the final latent embeddings.
For \textit{T-reg.}\thinspace FREUD, we additionally apply a Tanh function to the result and add a noise perturbation with $\sigma = 0.001$ to the latents. This small value is chosen to minimize adverse effects from perturbation and was not tuned or further ablated.

The FREUD decoder architecture is inspired by the Hourglass Diffusion Transformers (H-DiTs)~\cite{crowson2024Scalable}, which enable efficient diffusion in pixel space with a transformer architecture. The FREUD decoder employs downsampling, similar to the encoder, with a $1 \times 4 \times 4$ patching and a downsampling layer. The decoder downsampling layer uses spatio-temporally factorized~\cite{bertasius2021SpaceTime} neighborhood attention~\cite{hassani2023Neighborhood} with a $7 \times 7$ spatial and a 3-step temporal neighborhood. Therefore, each frame only attends to the immediate neighbor frames. In the bottleneck, we use 12 DiT blocks after concatenating the encoder latents channel-wise to the decoder feature maps. The upsampling layers of the decoder are built in parallel to the downsampling layers and use Token Split~\cite{shi2016RealTime} operations for upsampling.

\paragraph{Latent Space Model} The LSMs operate in the latent space of the FREUD first-stage. For improved convergence speed, we normalize latents to zero mean and unit variance with offsets determined on the training dataset. The LSMs use a standard DiT architecture~\cite{peebles2023Scalable, ma2024SiT} with $2 \times 2$ patching. The specific settings for all model sizes are provided in \cref{tab:sit_configuration}. We use 3D Axial RoPE and factorized spatio-temporal attention for all variants.

\begin{table}[t]
\centering
\caption{Small (S), Base (B), and Large (L) latent space model (LSM) configurations.} \label{tab:sit_configuration}
\footnotesize
\begin{tabular}{lcccc}
    \toprule
    Model & Layers & Hidden size & Attention Heads & Parameters \\
    \midrule
    S-LSM   & 12    & 384  & 6  & 44M   \\
    B-LSM   & 12    & 768  & 12 & 141M  \\
    L-LSM   & 24    & 1024 & 16 & 473M  \\
    \bottomrule
\end{tabular}
\end{table}

\subsection{Training}

\paragraph{Multi-stage Training}
Training transformers on high-resolution spatio-temporal nowcasting data is expensive~\cite{gao2022earthformer}. 
Exploiting RoPE’s sequence length generalization capabilities~\cite{su2023RoFormer, heo2024Rotary} and following current video training paradigms \cite{themoviegenteam@meta2024Movie}, we adopt a curriculum of training on progressively longer clips.
We first pre-train FREUD and LSMs on single frames, and then progressively increase the number of frames per clip, until we finally train on the full sequence.
During LSM image pre-training, we employ a standard unconditional diffusion training, which strengthens the model's spatial understanding before introducing the need to understand temporal dependencies.
FREUD gains only slightly from full-sequence training, showing good length generalization, whereas the LSMs improve substantially when training on the full sequence. We train the FREUD encoder for 250k iterations and the LSM for 350k iterations.

\paragraph{Additional Hyperparameters}
We train with the AdamW~\cite{loshchilov2019Decoupled} optimizer setting $\beta_1 = 0.9$ and $\beta_2 = 0.99$. We use bfloat16 precision in training for efficiency. Image pretraining uses a $160$ batch size on a single 40\,GB A100. Later stages are run with a global batch size of $128$. We use the Warmup-Stable-Decay (WSD) learning rate scheduler~\cite{hu2024MiniCPM}.

\paragraph{Outlier punishment}
Similar to MovieGen~\cite{themoviegenteam@meta2024Movie}, we initially find spot artifacts in reconstructions in our FREUD first stage. Therefore, we adopt the outlier punishment from MovieGen and penalize deviations of latent values from the latent mean if they exceed $r$ standard deviations. Following ~\cite{themoviegenteam@meta2024Movie} we use
\begin{align}\label{eq:opl-loss}
    &\mathcal{L}_{OPL}(\theta) =
    \nonumber \\
    &\frac{
    \sum_{i=1}^{H_l} \sum_{j=1}^{W_l} \max \Big [ ||\mathbf{z}_{i, j} - \text{mean}(\mathbf{z})||
    - r ||\text{Std}(\mathbf{z})||, 0 \Big]
    }{H_l \cdot W_l} \nonumber
\end{align}
\noindent
where $H_l$ and $W_l$ are the dimensions of the latent embeddings, and $\mathbf{z} \in \mathbb{R}^{C_l \times H_l \times W_l}$ are the frame-wise latents. In practice, we set $\lambda_{OPL} = 10^5$ to a large value and use $r=3$ as a common value in outlier detection. We observe $\mathcal{L}_{OPL}$ contributes only in the beginning of the training.

\subsection{SEVIR dataset}
The \textit{SEVIR} dataset~\cite{veillette2020SEVIR} is a weather observation dataset particularly well-suited for evaluating precipitation nowcasting methods because approximately 20\% of the 20,393 weather events are taken from NOAA's Storm Event Database~\cite{NOAA2020StormEvents} and, hence, represent extreme weather.
For the remaining 80\% of the dataset, random events are sampled while giving a higher probability to high precipitation events.
All events are drawn from the 2017--2019 time range and are recorded over the Continental United States. Each event covers a $384 \times 384$ km region over a 4\,h timespan.
To indicate precipitation, SEVIR uses the NEXRAD radar mosaic of Vertically Integrated Liquid (VIL).
VIL is recorded with 1\,km spatial and 5\,min temporal resolution.

We use the same test data as Earthformer and CasCast~\cite{gong2024cascast, gao2022earthformer} to calculate our metrics and use the remaining data for training.
Following previous work \cite{gong2024cascast, gao2022earthformer, she2024LLMDiff}, we use 13 frames (65\,min) as the conditioning to predict the next 12 frames (60\,min) of precipitation unless specified otherwise.
For the training data, we differ from previous work in taking exhaustive subsequences from each event.
As recommended by the dataset creators~\cite{veillette2020SEVIR}, we keep the non-linear encoding of VIL from SEVIR and normalize the encoded $[0, 255]$ values to the $[-1, 1]$ range.
At inference time, we can revert the normalization and apply the non-linear mapping
\begin{equation}
    x_{kg/m^2} =
    \begin{cases}
        0 \text{,} &\text{if }x_{pixel} \leq 5 \\
        (x_{pixel}-2)/90.66, &\text{if } 5 < x_{pixel} \leq 18 \\
        \exp ((x_{pixel} - 83.9) / 38.9) & 18 < x_{pixel}
    \end{cases} \nonumber
\end{equation}
to obtain the VIL in $kg/m^2$~\cite{veillette2020SEVIR}.

For consistency with previous work~\cite{gong2024cascast, gao2022earthformer, she2024LLMDiff} and alignment with the SEVIR data~\cite{veillette2020SEVIR}, we use $H=W=384$, $C=1$, $L^{in} = 13$, $L^{out} = 12$, and $T = 25$ unless specified otherwise.

\subsection{Metrics details}
To ensure comparability, we use the evaluation pipeline implemented by Gong et al. for the CasCast paper~\cite{gong2024cascast}.

\begin{figure}
    \centering
    \includegraphics[width=0.99\linewidth]{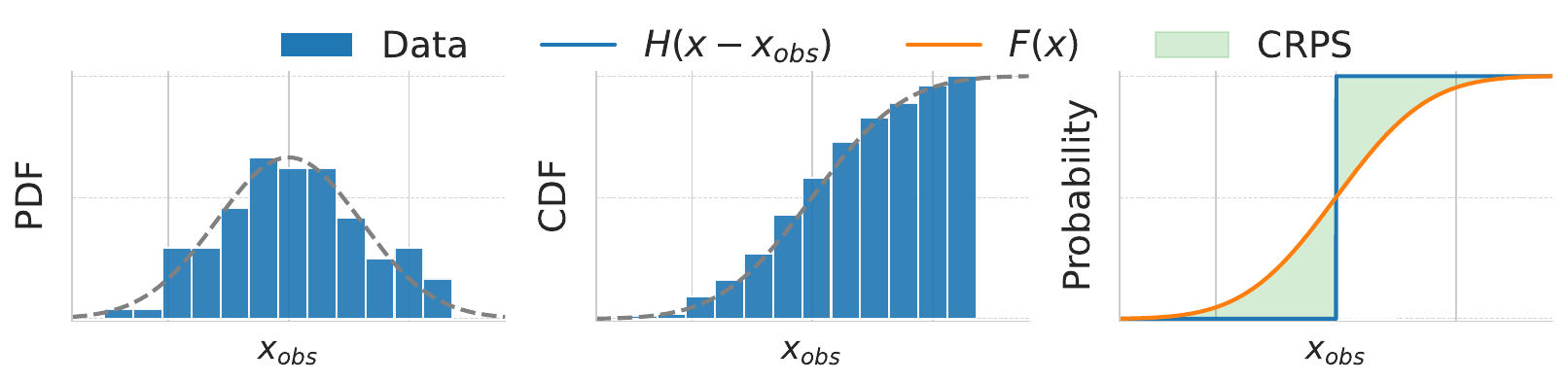}
    \caption{Schematic illustration of CRPS.}
    \label{fig:crps-explanation}
\end{figure}

\paragraph{CRPS}
We calculate the CRPS to measure the alignment of the predicted distribution with the ground truth distribution. CRPS is a generalization of the Mean Absolute Error (MAE) to probabilistic predictions. CRPS is calculated as
\begin{equation}
\begin{aligned}
    &CRPS(F, x) = \int_{-\infty}^{\infty} (F(y) - \mathbf 1_{y \geq x})^2 \text{ d}y \\
    &= \mathbb{E}_{X \sim F}\left [||X - x|| \right ] - \frac{1}{2} \mathbb{E}_{X, X' \sim F} \left [ X - X' \right] \\
    & \approx \frac{1}{N} \sum_{i = 1}^N |f_i - x| - \frac{1}{2 N^2} \sum_{i=1}^N \sum_{j=1}^N |f_i - f_j|, \nonumber
\end{aligned}
\end{equation}
where $x$ is the observed value, $F(y)$ is the cumulative distribution function of the forecast, and $\{ f_1, ..., f_N \}$ is the ensemble of forecasts. We show a schematic illustration of CRPS in \cref{fig:crps-explanation}.
Essentially, CRPS is the squared error between the ground truth cumulative distribution and the predicted cumulative distribution, which can be approximated with a finite ensemble using the MAE of the forecast and subtracting the MAE of ensemble members.

\paragraph{SSIM} Further, we use the Structural Similarity Index Measure (SSIM) to measure visual similarity between the ground truth precipitation and the forecast. SSIM for data with a single channel is calculated as
\begin{align}
 \text{SSIM}(x, y) =\frac{(2 \mu_x \mu_y + C_1)(2 \sigma_{xy} + C_2)}{(\mu_x^2 + \mu_y^2 + C_1)(\sigma_x^2 + \sigma_y^2 + C_2)},
 \nonumber
\end{align}
where $x$ is the ground truth and $y$ is the prediction, $\mu_x$ and $\mu_y$ are the average intensities, $\sigma_x^2$ and $\sigma_y^2$ are the variances of intensities, $\sigma_{xy}$ is the covariance between the images and $C_1$ and $C_2$ are two small constants for numerical stability. SSIM is calculated by averaging results from a sliding window across the image and using Gaussian weighting to calculate means and variances, whereby the largest weight is assigned to the central pixel.

\paragraph{HSS and CSI} We use the Heidke Skill Score (HSS) and Critical Success Index (CSI) computed on a per-pixel basis to identify positional inaccuracies.
HSS is calculated as
\begin{align}
    & \qquad \qquad \qquad \qquad \qquad HSS =
    \nonumber \\
 & \frac{
 2 (TP  \cdot TN - FP \cdot FN)
 }{(TP + FN)(FN + TN)+(TP + FP)(FP + TN)}. \nonumber
\end{align}
Here, the True Positives (TP), False Positives (FP), False Negatives (FN), and True Negatives (TN) are calculated with respect to some threshold. To obtain the reported HSS, we average HSS values calculated using six thresholds (16, 74, 133, 160, 181, 219). HSS indicates improvement of a prediction over random chance, where 0 indicates no forecasting skill beyond chance and 1 indicates a perfect forecast.

CSI or Threat Score quantifies the proportion of correctly predicted events while excluding true negatives, which makes it useful in scenarios where non-events (e.g., no precipitation) are more common than true events.
CSI is calculated as
\begin{equation}
    CSI = \frac{TP}{TP + FP + FN}. \nonumber
\end{equation}
Again, we report the CSI averaged over the six thresholds.

\section{Limitations}\label{sec:broad_impact}
Our method achieves superior distribution coverage and perceptual similarity as measured by CRPS and SSIM. However, especially the smaller models underperform for localization-centric metrics such as HSS and CSI, which aligns with previous work suggesting a trade-off between localization and distribution coverage~\cite{yu2024DiffCast}. While our method achieves a superior trade-off compared to prior work, future work could explore introducing different conditionings and priors that improve localization while keeping full distribution coverage. A possible solution might be to use noisy prior frames as the starting point for flow matching~\cite{lim2024Elucidating}, which, however, is non-trivial with non-autoregressive video modeling, as in RaMViD.

Our method yields more accurate and better-calibrated forecasts than the current state-of-the-art. Furthermore, FREUD can quantify reconstruction uncertainty at inference time, which was not possible with previous VAE-based compression stages. Sampling from the FREUD decoder achieves better calibration than samples from a VAE.
Still, our forecasts and FREUD reconstructions exhibit some overconfidence.
Therefore, future work should focus on developing calibration methods for generative flow models.
Related to this problem, our method, as well as prior methods, tend to underestimate precipitation. As generative models learn the distribution of the training data, this is expected because low precipitation is more common than extreme precipitation. 
Therefore, future work should explore options to better align the forecast distribution with the ground truth in rare extreme scenarios.

Previous work has solved this problem using cfg~\cite{ho2022ClassifierFree}; however, we find that the use of cfg is flawed in precipitation nowcasting, as increasing cfg leads to higher precipitation independent of the conditioning. This holds for our method as well as our strongest competitor, CasCast (see \cref{sec:cfg_ablation}). Therefore, the use of cfg does not yield improved alignment with conditions, but rather increases intensity.
Follow-up research should identify better guidance mechanisms that enable the generation of realistic forecasts that align with actual conditions.

\paragraph{}
As with all generative models, our approach inherits the statistical properties of its training data. Since high-quality weather datasets are primarily collected in technologically advanced regions with dense radar coverage, nowcasting methods for precipitation that rely on radar observations - including ours - are inherently limited to areas with such infrastructure, leaving large parts of the world (e.g., oceans and many developing regions) underserved. Future work should explore conditioning strategies that do not depend solely on radar. Cloud-top temperatures from geostationary satellites offer near-global coverage~\cite{theeuropeanspaceagency2020Types}, and decades of research have demonstrated their utility for estimating precipitation in radar-sparse regions~\cite{hong2004Precipitation, sadeghi2019PERSIANNCNN, japanmeteorologicalagency2014Users, milford1996Rainfall}. These alternative conditioning signals present a promising path toward globally applicable precipitation nowcasting.

\onecolumn
\twocolumn
\section{Weather Forecasting Literature Review}\label{sec:extensive_literature}
Weather nowcasting refers to short-term weather forecasting for periods of 30\,min to 12\,h, with high temporal and spatial resolution~\cite{gong2024cascast, espeholt22:MetNet2, leinonen2023ldcast, gao2023prediff}.
Traditionally, numerical weather prediction (NWP) systems, which simulate atmospheric evolution by numerically solving physical equations, have set the state-of-the-art for this task~\cite{sonderby20:MetNet, espeholt22:MetNet2, roberts2025High, dowell2022HighResolution}.
NWPs approximate solutions by averaging over spatial regions. For low-latency forecasts, higher downsampling is required, resulting in reduced resolution~\cite{li2024Latent}.
Further, the inherent non-linearity in atmospheric dynamics~\cite{navier1827Memoire, stokes2007Theories, landau1959Fluid} implies that minor perturbations or sensor inaccuracies will lead to exponential divergence over time~\cite{lorenz1963Deterministic, khalil2002Nonlinear, ott2000Chaos}. Therefore, NWPs are run multiple times with slightly altered inputs to estimate uncertainty, exploding computational costs.
As a computationally efficient alternative to NWPs, extrapolation-based techniques using optical flow are used~\cite{woo2017Operational, cheung2012Application, germann2002ScaleDependence, sakaino2013SpatioTemporal, woo2014Application}.
While these methods are much faster to compute, they are flawed because optical flow cannot model the formation and dissipation of weather patterns, limiting their predictive capability.

To overcome the limitations of traditional methods, recent work has turned to deep learning for efficient precipitation nowcasting. These approaches typically fall into two categories: \textit{deterministic} and \textit{probabilistic} models.
Deterministic nowcasting refers to models that make a single prediction based on the input and do not account for uncertainty in their predictions. 
In contrast, probabilistic nowcasting methods usually employ generative models to produce samples of possible outcomes, which allows for estimating uncertainty through repeated sampling.

\subsection{Deterministic}
\paragraph{Recurrent}
In an early work, Shi et al.~\cite{shi15:ConvLSTM} propose the ConvLSTM architecture, which extends the traditional LSTM~\cite{hochreiter1997Long} with convolutions to retain spatial structure. With this modification, they outperform a traditional optical flow-based nowcasting system~\cite{woo2014Application}. Later, they proposed the TrajGRU~\cite{shi17:TrajectoryGRU}, which improves the previous method by replacing fixed-size convolution with learned warping.
The PredRNN~\cite{wang2017predrnn} extends ConvLSTMs with a dual memory mechanism, integrating short-term cell and long-term memory, to better capture complex spatio-temporal dependencies.
PhyDNet~\cite{guen2020disentangling} outperforms the previous methods by learning a latent space with known dynamics, evolving the latent state according to these laws, and using a ConvLSTM to correct inaccuracies.

\paragraph{Convolutional}
Differing from the previous methods, Agrawal et al.~\cite{agrawal19:UNet} formulate nowcasting as an image-to-image translation problem and use a convolutional Unet architecture~\cite{ronneberger2015UNeta} with timesteps concatenated along the channel dimension. Their method outperforms optical flow and a low-latency NWP~\cite{dowell2022HighResolution}.
Trebing et al.~\cite{trebing21:SmaAt-UNet} improve computational efficiency by factorizing spatio-temporal attention and convolutions, reducing the number of parameters while retaining competitive performance.
In contrast, Fernàndez \& Mehrkanoon~\cite{fernandez21:Broad-UNet} explicitly model the temporal dimension using 3D convolutions instead of channel-wise concatenation and retain computational efficiency by factorizing 3D convolution into three axial convolutions. Additionally, convolutions with increasing dilation are used in the bottleneck. Their method outperforms the previous convolutional approaches, highlighting the advantage of temporal modeling.
Gao et al. propose "simpler yet better video prediction" (SimVP)~\cite{gao2022simvp}, which uses a frame-wise encoder and decoder and a convolution-based bottleneck model that captures temporal evolution. Their approach outperforms recurrent models~\cite{guen2020disentangling, shi15:ConvLSTM, wang2017predrnn}, underscoring the advantage of treating nowcasting as image translation.

\paragraph{Transformers}
Yang et al.~\cite{yang22:AA-TransUNet} propose a TransUnet~\cite{chen2024TransUNet} architecture for nowcasting where transformer blocks~\cite{vaswani2017Attention, dosovitskiy2020vit} are used in a Unet bottleneck. They further augment the Unet with attention and apply factorized convolutions, resulting in improved performance over a standard TransUnet and previous convolutional Unet architectures, revealing benefits of transformer-based modeling.
Discarding convolution entirely, Gao et al.~\cite{gao2022earthformer} introduce the Earthformer model with cuboid attention for efficient processing of high-resolution spatio-temporal data. Cuboid attention is applied to fixed-size spatio-temporal regions independently, and communication between cuboids is facilitated by global self-attention over class tokens. Their approach outperforms convolutional~\cite{agrawal19:UNet} and recurrent~\cite{shi15:ConvLSTM, wang2017predrnn, guen2020disentangling} architectures, cementing the advantage of attention-based modeling.
Similarly, Pathak et al.~\cite{pathak2022FourCastNet} also adopt a ViT-based~\cite{dosovitskiy2020vit} architecture but compute self-attention in the Fourier domain for efficiency. They outperform a state-of-the-art NWP~\cite{roberts2018Climate} for small-scale variables such as precipitation, demonstrating the superiority of deep learning for nowcasting.

\paragraph{Classification}
In contrast to the previous methods, Sonderby et al.~\cite{sonderby20:MetNet} treat nowcasting as classification and predict a SoftMax probability distribution over discrete intensity bins. Their approach, MetNet, conditions on a large spatial context of previous precipitation, cloud-top temperatures, and topology and uses a spatial encoder for efficient processing with a bottleneck ConvLSTM and Axial Self-Attention~\cite{ho2019Axial}.
MetNet-2~\cite{espeholt22:MetNet2} uses an even larger context, and an additional sequence of convolutions with increasing dilation to outperform a state-of-the-art NWP ensemble~\cite{molteni1996ECMWF} over the entire lead time range up to 12\,h, demonstrating the flexibility of deep learning-based nowcasting.

The major drawback of these deterministic methods is their lack of accurate uncertainty quantification. Methods trained with regression losses minimize the mean difference between predicted and observed values, where the minimizer is given as the expectation over the outcome, leading to mode averaging, which results in blurry predictions.
Models trained with a classification objective mitigate this problem, but the pixel-wise objective does not exploit spatial dependencies and cross-correlations. In addition, SoftMax probability distributions suffer from suboptimal calibration~\cite{hullermeier2021Aleatoric}.

\subsection{Probabilistic}
\paragraph{GAN nowcasting} As a solution, Ravuri et al.~\cite{ravuri21:GAN} propose to use a conditional Generative Adversarial Network (GAN)~\cite{mirza2014Conditional, goodfellow2014gan} with a temporal and spatial discriminator for nowcasting, where samples from the model are sharp and multiple samples can be produced to quantify uncertainty. Compared to deterministic methods such as MetNet~\cite{sonderby20:MetNet} and Unet~\cite{agrawal19:UNet}, and extrapolation (exemplified by PySteps~\cite{imhoff2023Scaledependent, pulkkinen2019Pysteps}), their method, Deep Generative Models of Radar (DGMR), achieves superior accuracy, calibration, and expert evaluation.
Ji et al.~\cite{ji22:CLGAN} propose a ConvLSTM-based GAN model that outperforms optical flow and ConvLSTM for heavy precipitation events, underscoring the importance of probabilistic modeling in critical extreme situations.
Liu et al.~\cite{liu2020MPLGAN} propose GAN-based forecasting with conditioning on a deterministic forecast, resulting in sharpened forecasts compared to the deterministic-only approach.
A similar approach is proposed by Price \& Rasp~\cite{price22:CorrectorGAN} who condition a GAN on a low-resolution NWP ensemble forecast to produce a high-resolution prediction. Their method is superior to simple interpolation and convolution-based upsampling and approaches the performance of a state-of-the-art NWP~\cite{molteni1996ECMWF} with drastically reduced complexity.
Finally, Zhang et al.~\cite{zhang2023skilful} propose NowCastNet, which combines a differentiable extrapolation with a learned residual and a stochastic GAN refiner. Their approach outperforms previous deterministic~\cite{wang2017predrnn}, probabilistic~\cite{ravuri21:GAN}, and extrapolation-based~\cite{imhoff2023Scaledependent, pulkkinen2019Pysteps} methods according to quantitative metrics and expert evaluations, highlighting the importance of sharp generative forecasts for downstream deployment.

While these works hint at the importance of probabilistic modeling for nowcasting, they rely on GANs, which are known to suffer from mode collapse~\cite{kossale2022Mode}. Therefore, the samples from these models will underestimate the true weather variance. Further, they might neglect additional smaller modes in the true weather distribution, which are particularly important in nowcasting rare extreme weather events. Therefore, these models fail to capture the full distribution of future weather evolution.

\paragraph{Diffusion-based nowcasting}
To tackle this problem, Leinonen et al.~\cite{leinonen2023ldcast} use a diffusion model with a solid mathematical foundation~\cite{sohl2015thermodynamics, ho2020ddpm} and empirically high sample variance~\cite{ho2020ddpm, nichol2021improved} for nowcasting. Due to the complexity of iterative sampling, they opt for a latent diffusion approach~\cite{rombach2022ldm}, and apply diffusion in the latent space of a VAE compression model. Their generative model is conditioned on a deterministic prediction, which is augmented by the diffusion model. This architecture outperforms GAN-based nowcasting~\cite{ravuri21:GAN}, especially for detecting extreme weather events.
Similarly, Gao et al.~\cite{gao2023prediff} use a latent diffusion approach for their PreDiff architecture, but discard the deterministic conditioning in favor of physics-based guidance and conditioning on previous precipitation maps. Their approach consistently outperforms a range of deterministic~\cite{agrawal19:UNet, shi15:ConvLSTM, wang2017predrnn, guen2020disentangling, gao2022earthformer} and GAN-based~\cite{ravuri21:GAN} approaches.

Since these early successes, architectural modifications have been proposed. Asperti et al.~\cite{asperti2023Precipitation} suggest a learned aggregation of ensemble members instead of the canonical mean, while She et al.~\cite{she2024LLMDiff} integrate a transformer block from a large vision language model to ease training. Nai et al.~\cite{nai2024Reliable} identify classifier-free guidance~\cite{ho2022ClassifierFree} as an option to improve forecasting skill. Moreover, Ling et al.~\cite{ling2024SRNDiff} raise concerns about latent diffusion, as the cascaded architecture can lead to error accumulation and unquantified uncertainty. As an alternative, they suggest training the model end-to-end while integrating conditioning by concatenating feature maps from an condition encoder model.

Diffusion has also been used for longer-range forecasts. Li et al.~\cite{li2024Latent} forecast precipitation over 16 days with a latent diffusion model conditioned on a deterministic forecast and outperform two NWP approaches~\cite{zhou2022Development, powers2017Weather}, cementing diffusion-based weather forecasting as a viable alternative to simulation.
Further, Stock et al.~\cite{stock2024DiffObs} show that by auto-regressively unrolling a $1$\,h ahead diffusion-based forecast, reasonable forecasts up to 60 days lead time can be obtained and seasonal patterns are captured in year-long rollouts, indicating a robust understanding of the weather system in diffusion models.
Similarly, Shi et al.~\cite{shi2024CoDiCast} outperform a range of deterministic baselines~\cite{chen2018neural, pathak2022FourCastNet, nguyen23:ClimaX, verma_climode_2024} with auto-regressive modeling conditioned on only two initial states.
Building on this success, Price et al.~\cite{price2024GenCast} introduce GenCast as a follow-up to GraphCast~\cite{lam2023Learning}, providing 15-day global forecasts of 25 variables with 27\,km spatial and 12\,h temporal resolution. The model is conditioned on two previous states, forecasts auto-regressively, and uses a Graph Transformer~\cite{yun2019Graph} backbone. They outperform a top-of-the-line ensemble NWP system~\cite{molteni1996ECMWF}, demonstrating the superiority of deep-learning-based generative forecasting with diffusion.
Recently, Alet et al.~\cite{alet2025skillful} proposed \textit{Functional Generative Networks} (FGN) as a follow-up to GenCast. They independently train an ensemble of predictors to account for epistemic uncertainty. To model aleatoric uncertainty, the authors turn the ensemble members themselves probabilistic by sampling a low-dimensional noise vector and injecting it via conditional normalization layers, effectively acting as a structured weight perturbation. The models are then trained directly using CRPS as the loss function, encouraging the models to make accurate and diverse predictions. The proposed model outperforms GenCast while providing substantially faster generation. Their results call the current dominance of diffusion-based approaches into question, although it remains unclear how well this approach transfers to domains where fine-scale stochasticity plays a larger role, such as precipitation nowcasting.

\subsection{Cascaded deterministic-probabilistic}
Early results show that conditioning a diffusion model on a deterministic forecast yields high-quality forecasts~\cite{li2023SEEDS, chen2023SwinRDM, leinonen2023ldcast, ling2024SRNDiff, li2024Latent, shi2024CoDiCast}.
Yu et al.~\cite{yu2024DiffCast} analyze these results and determine that probabilistic diffusion-only approaches show superior distribution coverage and sharpness than deterministic forecasts, but often suffer in terms of accurate localization. Building on physical knowledge, they suggest using a diffusion model to predict the residual between a deterministic forecast and the ground truth and find improved performance over deterministic-only~\cite{gao2022earthformer, gao2022simvp} or diffusion-only~\cite{gao2023prediff} approaches.
Similarly, CasCast~\cite{gong2024cascast} conditions a latent space diffusion transformer~\cite{peebles2023Scalable} on a deterministic forecast and applies sequence-wise diffusion transformer blocks after spatial encoding. Their model consistently outperforms deterministic~\cite{shi15:ConvLSTM, gao2023prediff, guen2020disentangling, gao2022simvp, gao2022earthformer}, GAN-based~\cite{zhang2018lpips}, and diffusion-only~\cite{gao2023prediff} nowcasting methods.
Inspired by these results, Pathak et al.~\cite{pathak2024KilometerScale} condition an auto-regressive diffusion model on a deterministic initial forecast, and find their model outperforms a low-latency NWP~\cite{dowell2022HighResolution}. However, they also indicate their method is overconfident and not well calibrated, hinting at a disadvantage of cascaded architectures.

We improve upon these prior works by compressing pixel-space data into a low-resolution latent space while quantifying the uncertainty from compression~\cite{ling2024SRNDiff}. Further, we avoid biasing the generation with a deterministic forecast~\cite{pathak2024KilometerScale} and aim to achieve similar localization by exploiting the advantages of transformer-based architectures that allow to exploit conditioning more effectively with self-attention.

\FloatBarrier
\newpage
\section{Uncurated Qualitative Results}
\label{sec:qualitative}
In the following we provide further qualitative samples.
\begin{figure*}[htbp]
    \centering \small
    \setlength\tabcolsep{1pt}

    \newcommand{\imagepng}[3]{
        \includegraphics[width=0.1\linewidth]{fig/freud/ensembles-blobs/T-seed_42/#1_blobs/#2_#3.png}
    }
    
    \newcommand{\rowpng}[1]{
        \imagepng{#1}{gt}{0} &
        \imagepng{#1}{ens}{0} &
        \imagepng{#1}{ens}{1} &
        \imagepng{#1}{ens}{2} &
        \imagepng{#1}{ens}{3} &
        \imagepng{#1}{ens}{4} &
        \imagepng{#1}{var}{0}
    }

    \begin{subfigure}{1\textwidth}
        \centering
        \begin{tabular}{ccccccccc}
        & GT & E1 & E2 & E3 & E4 & E5 & Variance & \multirow{3}{*}{%
          \begin{minipage}[c]{0.137\linewidth}
            \centering
            \vspace{6.8em}
            \includegraphics[width=\linewidth]{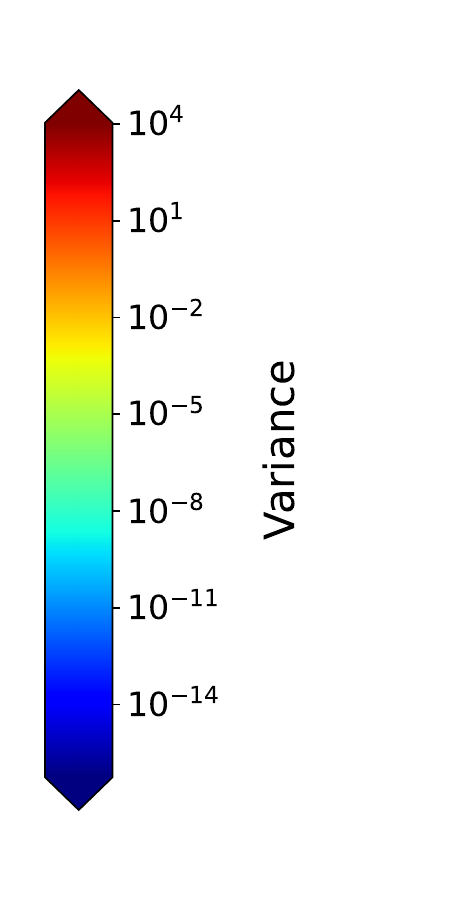}
          \end{minipage}%
        }\\
        \rotatebox{90}{\hspace{.5em} 0 Blobs} & \rowpng{0} \\
        \rotatebox{90}{\hspace{.5em} 5 Blobs} & \rowpng{5} \\
        \rotatebox{90}{\hspace{.5em} 10 Blobs} & \rowpng{10} \\
        \rotatebox{90}{\hspace{.25em} 100 Blobs} & \rowpng{100} \\
        \multicolumn{9}{c}{\includegraphics[width=0.4\linewidth]{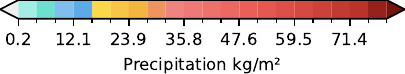}} \\
        \end{tabular}
    \end{subfigure}
    \caption{Qualitative results for blob experiment. The variance around blobs is high. Therefore, our first stage can localize abnormal features using ensemble variance. Best viewed zoomed in.}
    \label{fig:freud:qualitative_blobs_42}
\end{figure*}

\begin{figure*}[htbp]
    \centering \small
    \setlength\tabcolsep{1pt}

    \newcommand{\imagepng}[3]{
        \includegraphics[width=0.1\linewidth]{fig/freud/ensembles-blobs/T-seed_741/#1_blobs/#2_#3.png}
    }
    
    \newcommand{\rowpng}[1]{
        \imagepng{#1}{gt}{0} &
        \imagepng{#1}{ens}{0} &
        \imagepng{#1}{ens}{1} &
        \imagepng{#1}{ens}{2} &
        \imagepng{#1}{ens}{3} &
        \imagepng{#1}{ens}{4} &
        \imagepng{#1}{var}{0}
    }

    \begin{subfigure}{1\textwidth}
        \centering
        \begin{tabular}{ccccccccc}
        & GT & E1 & E2 & E3 & E4 & E5 & Variance & \multirow{3}{*}{%
          \begin{minipage}[c]{0.137\linewidth}
            \centering
            \vspace{6.8em}
            \includegraphics[width=\linewidth]{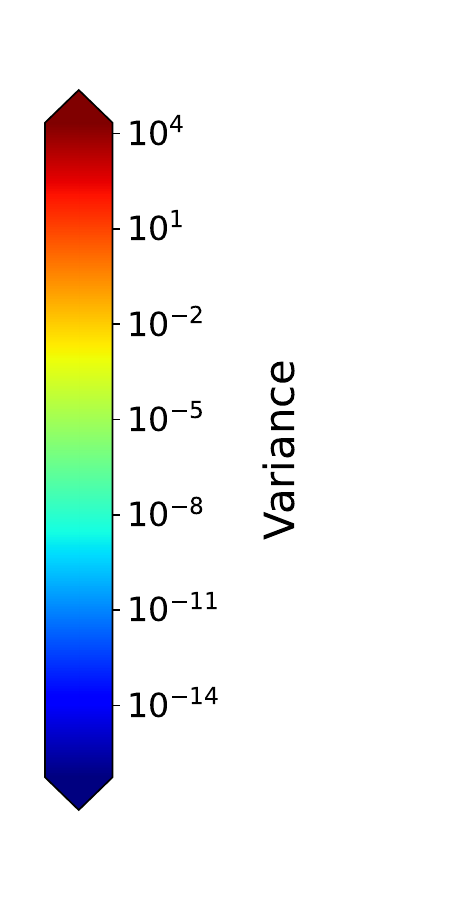}
          \end{minipage}%
        }\\
        \rotatebox{90}{\hspace{.5em} 0 Blobs} & \rowpng{0} \\
        \rotatebox{90}{\hspace{.5em} 5 Blobs} & \rowpng{5} \\
        \rotatebox{90}{\hspace{.5em} 10 Blobs} & \rowpng{10} \\
        \rotatebox{90}{\hspace{.25em} 100 Blobs} & \rowpng{100} \\
        \multicolumn{9}{c}{\includegraphics[width=0.4\linewidth]{fig/lsm/forecast/colorbar.pdf}} \\
        \end{tabular}
    \end{subfigure}
    \caption{Qualitative results for blob experiment. The variance around blobs is high. Therefore, our first stage can localize abnormal features using ensemble variance. Best viewed zoomed in.}
    \label{fig:freud:qualitative_blobs741}
\end{figure*}

\begin{figure*}[htbp]
    \centering \small
    \setlength\tabcolsep{1pt}

    \newcommand{\imagepng}[3]{
        \includegraphics[width=0.13\linewidth]{fig/lsm/forecast/cfg-T-B_seed0/#1_cfg#2_#3.png}
    }

    \newcommand{\rowpng}[2]{
        \imagepng{#1}{#2}{1} & %
        \imagepng{#1}{#2}{3} & %
        \imagepng{#1}{#2}{5} & %
        \imagepng{#1}{#2}{7} & %
        \imagepng{#1}{#2}{9} & %
        \imagepng{#1}{#2}{11} %
    }

    \begin{minipage}{.9\textwidth}
        \begin{subfigure}{1\textwidth}
        \centering
        \begin{tabular}{lcccccc}
            & $+10$\,min & $+20$\,min & $+30$\,min &  $+40$\,min &  $+50$\,min &  $+60$\,min\\
            \rotatebox{90}{\hspace{0.em} $cfg=0.0$} & \rowpng{ens}{0} \\
            \rotatebox{90}{\hspace{0.em} $cfg=1.0$} & \rowpng{ens}{1} \\
            \rotatebox{90}{\hspace{0.em} $cfg=2.0$} & \rowpng{ens}{2} \\
            \rotatebox{90}{\hspace{0.em} $cfg=3.0$} & \rowpng{ens}{3} \\
            \rotatebox{90}{\hspace{0.em} $cfg=4.0$} & \rowpng{ens}{4} \\
            \rotatebox{90}{\hspace{0.em} $cfg=5.0$} & \rowpng{ens}{5} \\
            \arrayrulecolor{gray!50!white} \cmidrule{1-7} \arrayrulecolor{black} \\
            \rotatebox{90}{\hspace{1.5em} GT} & \rowpng{gt}{0} \\
        \end{tabular}
    \end{subfigure}
    \end{minipage}
    \hspace{-0.8cm}
    \begin{minipage}{0.09\textwidth}
        \centering
        \includegraphics[width=\linewidth]{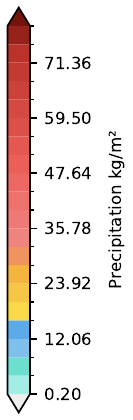}
    \end{minipage}
    \caption{Qualitative results obtained with our B-LSM in the \textit{T-reg.}\thinspace latent space for different guidance scales~\cite{ho2022ClassifierFree}. Guidance 0.0 indicates unconditional sampling, while guidance 1.0 indicates conditional sampling without guidance. Best viewed zoomed in.}
    \label{fig:lsm:cfg_qual_1}
\end{figure*}

\begin{figure*}[htbp]
    \centering \small
    \setlength\tabcolsep{1pt}

    \newcommand{\imagepng}[3]{
        \includegraphics[width=0.13\linewidth]{fig/lsm/forecast/cfg-T-B_seed17/#1_cfg#2_#3.png}
    }

    \newcommand{\rowpng}[2]{
        \imagepng{#1}{#2}{1} & %
        \imagepng{#1}{#2}{3} & %
        \imagepng{#1}{#2}{5} & %
        \imagepng{#1}{#2}{7} & %
        \imagepng{#1}{#2}{9} & %
        \imagepng{#1}{#2}{11} %
    }

    \begin{minipage}{.9\textwidth}
        \begin{subfigure}{1\textwidth}
        \centering
        \begin{tabular}{lcccccc}
            & $+10$\,min & $+20$\,min & $+30$\,min &  $+40$\,min &  $+50$\,min &  $+60$\,min\\
            \rotatebox{90}{\hspace{0.em} $cfg=0.0$} & \rowpng{ens}{0} \\
            \rotatebox{90}{\hspace{0.em} $cfg=1.0$} & \rowpng{ens}{1} \\
            \rotatebox{90}{\hspace{0.em} $cfg=2.0$} & \rowpng{ens}{2} \\
            \rotatebox{90}{\hspace{0.em} $cfg=3.0$} & \rowpng{ens}{3} \\
            \rotatebox{90}{\hspace{0.em} $cfg=4.0$} & \rowpng{ens}{4} \\
            \rotatebox{90}{\hspace{0.em} $cfg=5.0$} & \rowpng{ens}{5} \\
            \arrayrulecolor{gray!50!white} \cmidrule{1-7} \arrayrulecolor{black} \\
            \rotatebox{90}{\hspace{1.5em} GT} & \rowpng{gt}{0} \\
        \end{tabular}
    \end{subfigure}
    \end{minipage}
    \hspace{-0.8cm}
    \begin{minipage}{0.09\textwidth}
        \centering
        \includegraphics[width=\linewidth]{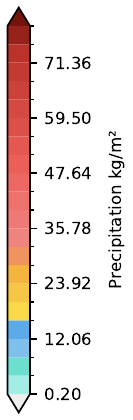}
    \end{minipage}
    \caption{Qualitative results obtained with our B-LSM in the \textit{T-reg.}\thinspace latent space for different guidance scales~\cite{ho2022ClassifierFree}. Guidance 0.0 indicates unconditional sampling, while guidance 1.0 indicates conditional sampling without guidance. Best viewed zoomed in.}
    \label{fig:lsm:cfg_qual_2}
\end{figure*}

\begin{figure*}[tbh]
    \centering \small
    \setlength\tabcolsep{1pt}

    \newcommand{\imagepng}[3]{
        \includegraphics[width=0.12\linewidth]{fig/lsm/forecast/CasCast_seed0/#1_cfg#2_#3.png}
    }

    \newcommand{\rowpng}[2]{
        \imagepng{#1}{#2}{1} & %
        \imagepng{#1}{#2}{3} & %
        \imagepng{#1}{#2}{5} & %
        \imagepng{#1}{#2}{7} & %
        \imagepng{#1}{#2}{9} & %
        \imagepng{#1}{#2}{11} %
    }

    \begin{minipage}{.9\textwidth}
        \begin{subfigure}{1\textwidth}
        \centering
        \begin{tabular}{lcccccc}
            & $+10$\,min & $+20$\,min & $+30$\,min &  $+40$\,min &  $+50$\,min &  $+60$\,min\\
            \rotatebox{90}{\hspace{0.em} $cfg=0.0$} & \rowpng{ens}{0} \\
            \rotatebox{90}{\hspace{0.em} $cfg=1.0$} & \rowpng{ens}{1} \\
            \rotatebox{90}{\hspace{0.em} $cfg=2.0$} & \rowpng{ens}{2} \\
            \rotatebox{90}{\hspace{0.em} $cfg=4.0$} & \rowpng{ens}{4} \\
            \rotatebox{90}{\hspace{0.em} $cfg=8.0$} & \rowpng{ens}{8} \\
            \rotatebox{90}{\hspace{0.em} $cfg=12.0$} & \rowpng{ens}{12} \\
            \arrayrulecolor{gray!50!white} \cmidrule{1-7} \arrayrulecolor{black} \\
            \rotatebox{90}{\hspace{1.5em} GT} & \rowpng{gt}{0} \\
        \end{tabular}
    \end{subfigure}
    \end{minipage}
    \hspace{-0.8cm}
    \begin{minipage}{0.09\textwidth}
        \centering
        \includegraphics[width=\linewidth]{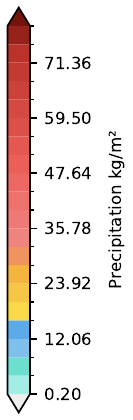}
    \end{minipage}
    \caption{Qualitative results obtained with \textbf{CasCast}~\cite{gong2024cascast} for different guidance scales~\cite{ho2022ClassifierFree}. Guidance 0.0 indicates unconditional sampling, while guidance 1.0 indicates conditional sampling without guidance. Best viewed zoomed in.}
    \label{fig:cascast:qual_1}
\end{figure*}

\begin{figure*}[tbh]
    \centering \small
    \setlength\tabcolsep{1pt}

    \newcommand{\imagepng}[3]{
        \includegraphics[width=0.13\linewidth]{fig/lsm/forecast/CasCast_seed2025/#1_cfg#2_#3.png}
    }

    \newcommand{\rowpng}[2]{
        \imagepng{#1}{#2}{1} & %
        \imagepng{#1}{#2}{3} & %
        \imagepng{#1}{#2}{5} & %
        \imagepng{#1}{#2}{7} & %
        \imagepng{#1}{#2}{9} & %
        \imagepng{#1}{#2}{11} %
    }

    \begin{minipage}{.9\textwidth}
        \begin{subfigure}{1\textwidth}
        \centering
        \begin{tabular}{lcccccc}
            & $+10$\,min & $+20$\,min & $+30$\,min &  $+40$\,min &  $+50$\,min &  $+60$\,min\\
            \rotatebox{90}{\hspace{0.em} $cfg=0.0$} & \rowpng{ens}{0} \\
            \rotatebox{90}{\hspace{0.em} $cfg=1.0$} & \rowpng{ens}{1} \\
            \rotatebox{90}{\hspace{0.em} $cfg=2.0$} & \rowpng{ens}{2} \\
            \rotatebox{90}{\hspace{0.em} $cfg=4.0$} & \rowpng{ens}{4} \\
            \rotatebox{90}{\hspace{0.em} $cfg=8.0$} & \rowpng{ens}{8} \\
            \rotatebox{90}{\hspace{0.em} $cfg=12.0$} & \rowpng{ens}{12} \\
            \arrayrulecolor{gray!50!white} \cmidrule{1-7} \arrayrulecolor{black} \\
            \rotatebox{90}{\hspace{1.5em} GT} & \rowpng{gt}{0} \\
        \end{tabular}
    \end{subfigure}
    \end{minipage}
    \hspace{-0.8cm}
    \begin{minipage}{0.09\textwidth}
        \centering
        \includegraphics[width=\linewidth]{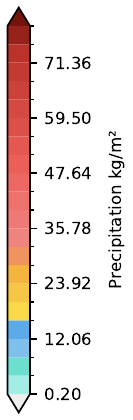}
    \end{minipage}
    \caption{Qualitative results obtained with \textbf{CasCast}~\cite{gong2024cascast} for different guidance scales~\cite{ho2022ClassifierFree}. Guidance 0.0 indicates unconditional sampling, while guidance 1.0 indicates conditional sampling without guidance. Best viewed zoomed in.}
    \label{fig:cascast:qual_2}
\end{figure*}

\begin{figure*}[t]
    \centering \small
    \setlength\tabcolsep{1pt}

    \newcommand{\imagepng}[3]{
        \includegraphics[width=0.12\linewidth]{fig/lsm/forecast/uncond_T-B/#1_#2_#3.png}
    }

    \newcommand{\rowpng}[2]{
        \imagepng{#1}{#2}{1} & %
        \imagepng{#1}{#2}{3} & %
        \imagepng{#1}{#2}{5} & %
        \imagepng{#1}{#2}{7} & %
        \imagepng{#1}{#2}{9} & %
        \imagepng{#1}{#2}{11} %
    }

    \begin{minipage}{.9\textwidth}
        \begin{subfigure}{1\textwidth}
        \centering
        \begin{tabular}{lcccccc}
            & $+10$\,min & $+20$\,min & $+30$\,min & $+40$\,min & $+50$\,min & $+60$\,min\\
            \rotatebox{90}{\hspace{1.5em} S1} & \rowpng{ens}{0} \\
            \rotatebox{90}{\hspace{1.5em} S2} & \rowpng{ens}{1} \\
            \rotatebox{90}{\hspace{1.5em} S3} & \rowpng{ens}{3} \\
        \end{tabular}
    \end{subfigure}
    \end{minipage}
    \hspace{-0.8cm}
    \begin{minipage}{0.09\textwidth}
        \centering
        \includegraphics[width=\linewidth]{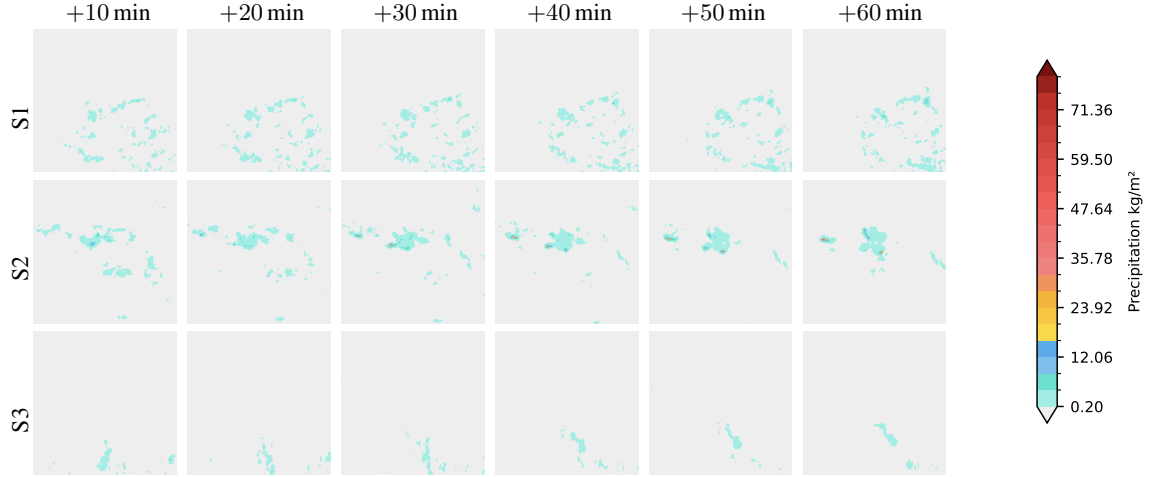}
    \end{minipage}
    \caption{\textit{Unconditional} qualitative samples from the B-LSM in \textit{T-reg.}\thinspace latent space. We observe that unconditional samples are temporally consistent, and sharp, but tend to contain low precipitation.}
    \label{fig:lsm:unconditional_qualitative}
\end{figure*}

\begin{figure*}[htbp]
    \centering \small
    \setlength\tabcolsep{1pt}

    \newcommand{\imagepng}[3]{
        \includegraphics[width=0.115\linewidth]{fig/freud/ensembles-framewise/T-seed_42/#1_precip/#2_#3.png}
    }
    
    \newcommand{\rowpng}[1]{
        \imagepng{#1}{gt}{0} &
        \imagepng{#1}{ens}{0} &
        \imagepng{#1}{ens}{1} &
        \imagepng{#1}{ens}{2} &
        \imagepng{#1}{ens}{3} &
        \imagepng{#1}{ens}{4} &
        \imagepng{#1}{var}{0}
    }

    \begin{subfigure}{1\textwidth}
        \centering
        \begin{tabular}{ccccccccc}
        & GT & E1 & E2 & E3 & E4 & E5 & Variance & \multirow{3}{*}{%
          \begin{minipage}[c]{0.137\linewidth}
            \centering
            \vspace{.5em}
            \includegraphics[width=\linewidth]{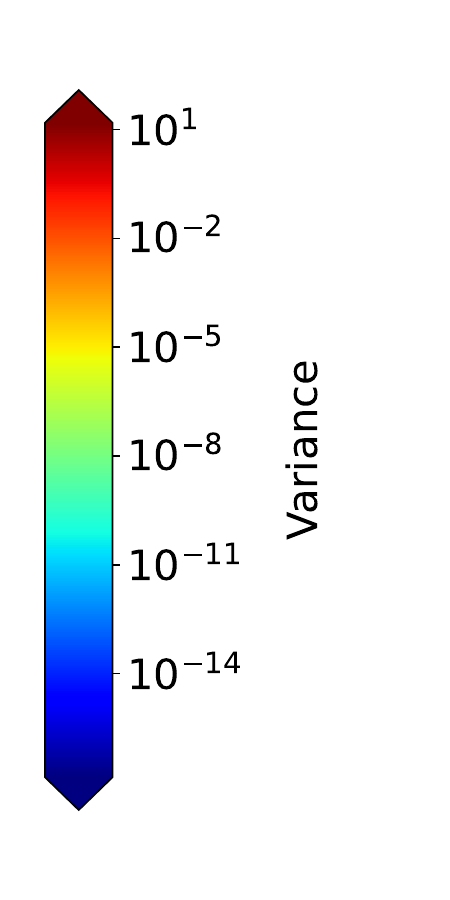}
          \end{minipage}%
        }\\
        \rotatebox{90}{\hspace{.5em} Normal} & \rowpng{low} \\
        \rotatebox{90}{\hspace{.5em} Extreme} & \rowpng{high} \\
        \multicolumn{9}{c}{\includegraphics[width=0.4\linewidth]{fig/lsm/forecast/colorbar.pdf}} \\
        \end{tabular}
    \end{subfigure}
    \caption{Qualitative reconstruction sampled with the frame-wise DiffAE. Qualitatively, the frame-wise reconstructions match the ground truth well, validating the strong reconstruction performance of the generative decoder.}
    \label{fig:framewise:qual_42}
\end{figure*}

\begin{figure*}[htbp]
    \centering \small
    \setlength\tabcolsep{1pt}

    \newcommand{\imagepng}[3]{
        \includegraphics[width=0.115\linewidth]{fig/freud/ensembles-framewise/T-seed_51/#1_precip/#2_#3.png}
    }
    
    \newcommand{\rowpng}[1]{
        \imagepng{#1}{gt}{0} &
        \imagepng{#1}{ens}{0} &
        \imagepng{#1}{ens}{1} &
        \imagepng{#1}{ens}{2} &
        \imagepng{#1}{ens}{3} &
        \imagepng{#1}{ens}{4} &
        \imagepng{#1}{var}{0}
    }

    \begin{subfigure}{1\textwidth}
        \centering
        \begin{tabular}{ccccccccc}
        & GT & E1 & E2 & E3 & E4 & E5 & Variance & \multirow{3}{*}{%
          \begin{minipage}[c]{0.137\linewidth}
            \centering
            \vspace{.5em}
            \includegraphics[width=\linewidth]{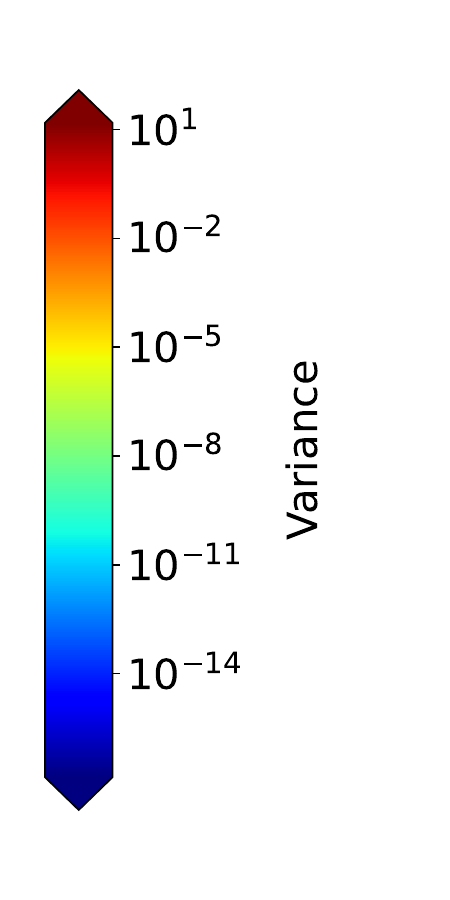}
          \end{minipage}%
        }\\
        \rotatebox{90}{\hspace{.5em} Normal} & \rowpng{low} \\
        \rotatebox{90}{\hspace{.5em} Extreme} & \rowpng{high} \\
        \multicolumn{9}{c}{\includegraphics[width=0.4\linewidth]{fig/lsm/forecast/colorbar.pdf}} \\
        \end{tabular}
    \end{subfigure}
    \caption{Qualitative reconstruction sampled with the frame-wise DiffAE. Qualitatively, the frame-wise reconstructions match the ground truth well, validating the strong reconstruction performance of the generative decoder.}
    \label{fig:framewise:qual_51}
\end{figure*}

\begin{figure*}[htbp]
    \centering \small
    \setlength\tabcolsep{1pt}

    \newcommand{\imagepng}[3]{
        \includegraphics[width=0.11\linewidth]{fig/freud/ensembles/T-seed_45/#1_precip/#2_#3.png}
    }
    
    \newcommand{\rowpng}[1]{
        \imagepng{#1}{gt}{0} &
        \imagepng{#1}{ens}{0} &
        \imagepng{#1}{ens}{1} &
        \imagepng{#1}{ens}{2} &
        \imagepng{#1}{ens}{3} &
        \imagepng{#1}{ens}{4} &
        \imagepng{#1}{var}{0}
    }

    \begin{subfigure}{1\textwidth}
        \centering
        \begin{tabular}{ccccccccc}
        & GT & E1 & E2 & E3 & E4 & E5 & Variance & \multirow{3}{*}{%
          \begin{minipage}[c]{0.137\linewidth}
            \centering
            \includegraphics[width=\linewidth]{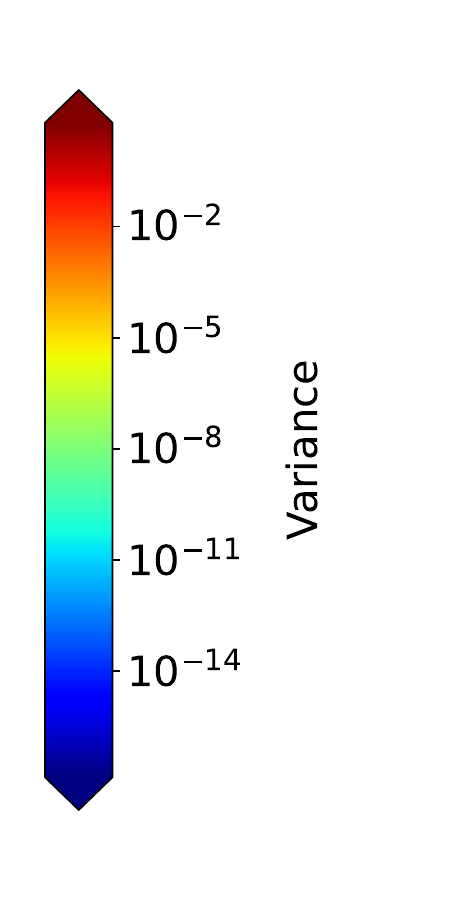}
          \end{minipage}%
        }\\
        \rotatebox{90}{\hspace{.75em} Normal} & \rowpng{low} \\
        \rotatebox{90}{\hspace{.75em} Extreme} & \rowpng{high} \\
        \multicolumn{9}{c}{\includegraphics[width=0.4\linewidth]{fig/lsm/forecast/colorbar.pdf}} \\
        \end{tabular}
    \end{subfigure}
    \caption{Qualitative results for reconstruction of one normal and one extreme weather event. Variance map uses log scale. Areas with precipitation show higher variance. Areas with extreme precipitation exhibit extreme variance. Best viewed zoomed in.}
    \label{fig:freud:qualitative_45}
\end{figure*}

\begin{figure*}[htbp]
    \centering \small
    \setlength\tabcolsep{1pt}

    \newcommand{\imagepng}[3]{
        \includegraphics[width=0.11\linewidth]{fig/freud/ensembles/T-seed_2025/#1_precip/#2_#3.png}
    }
    
    \newcommand{\rowpng}[1]{
        \imagepng{#1}{gt}{0} &
        \imagepng{#1}{ens}{0} &
        \imagepng{#1}{ens}{1} &
        \imagepng{#1}{ens}{2} &
        \imagepng{#1}{ens}{3} &
        \imagepng{#1}{ens}{4} &
        \imagepng{#1}{var}{0}
    }

    \begin{subfigure}{1\textwidth}
        \centering
        \begin{tabular}{ccccccccc}
        & GT & E1 & E2 & E3 & E4 & E5 & Variance & \multirow{3}{*}{%
          \begin{minipage}[c]{0.137\linewidth}
            \centering
            \includegraphics[width=\linewidth]{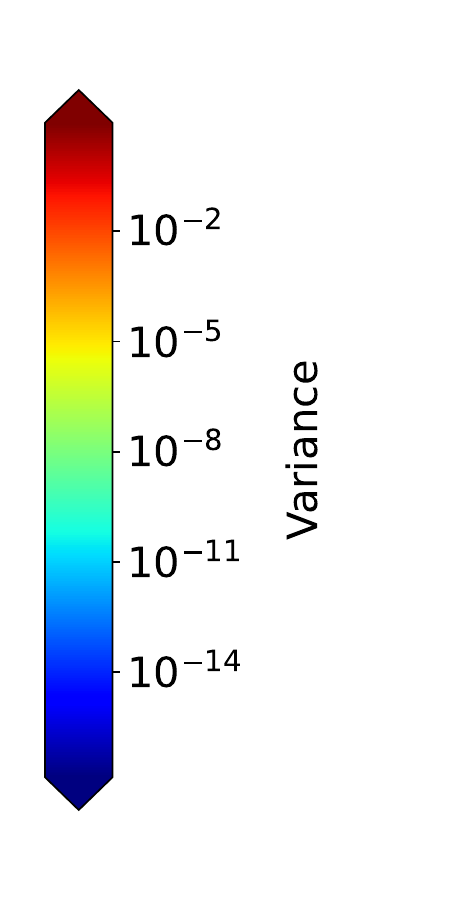}
          \end{minipage}%
        }\\
        \rotatebox{90}{\hspace{.75em} Normal} & \rowpng{low} \\
        \rotatebox{90}{\hspace{.75em} Extreme} & \rowpng{high} \\
        \multicolumn{9}{c}{\includegraphics[width=0.4\linewidth]{fig/lsm/forecast/colorbar.pdf}} \\
        \end{tabular}
    \end{subfigure}
    \caption{Qualitative results for reconstruction of one normal and one extreme weather event. Variance map uses log scale. Areas with precipitation show higher variance. Areas with extreme precipitation exhibit extreme variance. Best viewed zoomed in.}
    \label{fig:freud:qualitative_2025}
\end{figure*}

\begin{figure*}[htbp]
    \centering \small
    \setlength\tabcolsep{1pt}

    \newcommand{\imagepng}[3]{
        \includegraphics[width=0.11\linewidth]{fig/freud/ensembles/T-seed_2026/#1_precip/#2_#3.png}
    }
    
    \newcommand{\rowpng}[1]{
        \imagepng{#1}{gt}{0} &
        \imagepng{#1}{ens}{0} &
        \imagepng{#1}{ens}{1} &
        \imagepng{#1}{ens}{2} &
        \imagepng{#1}{ens}{3} &
        \imagepng{#1}{ens}{4} &
        \imagepng{#1}{var}{0}
    }

    \begin{subfigure}{1\textwidth}
        \centering
        \begin{tabular}{ccccccccc}
        & GT & E1 & E2 & E3 & E4 & E5 & Variance & \multirow{3}{*}{%
          \begin{minipage}[c]{0.137\linewidth}
            \centering
            \includegraphics[width=\linewidth]{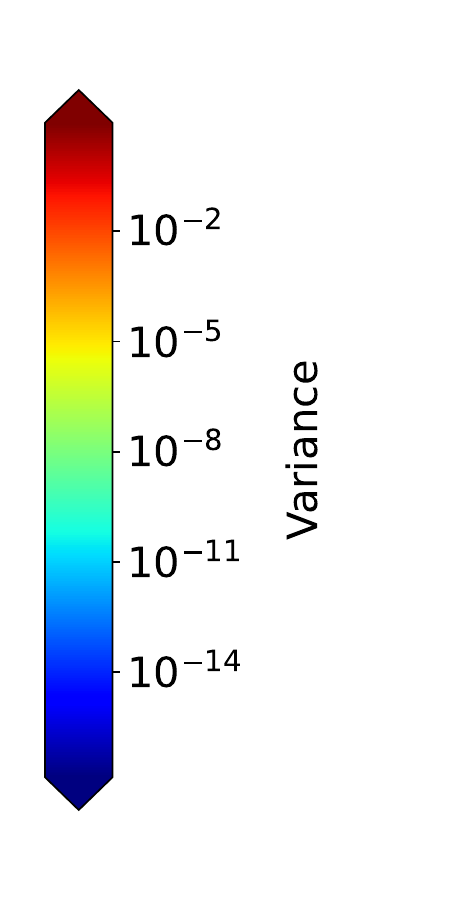}
          \end{minipage}%
        }\\
        \rotatebox{90}{\hspace{.75em} Normal} & \rowpng{low} \\
        \rotatebox{90}{\hspace{.75em} Extreme} & \rowpng{high} \\
        \multicolumn{9}{c}{\includegraphics[width=0.4\linewidth]{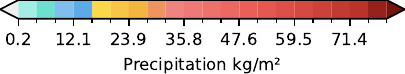}} \\
        \end{tabular}
    \end{subfigure}
    \caption{Qualitative results for reconstruction of one normal and one extreme weather event. Variance map uses log scale. Areas with precipitation show higher variance. Areas with extreme precipitation exhibit extreme variance. Best viewed zoomed in.}
    \label{fig:freud:qualitative_2026}
\end{figure*}

\begin{figure*}[htbp]
    \centering \small
    \setlength\tabcolsep{1pt}

    \newcommand{\imagepng}[3]{
        \includegraphics[width=0.115\linewidth]{fig/freud/ensembles/T-seed_2027/#1_precip/#2_#3.png}
    }
    
    \newcommand{\rowpng}[1]{
        \imagepng{#1}{gt}{0} &
        \imagepng{#1}{ens}{0} &
        \imagepng{#1}{ens}{1} &
        \imagepng{#1}{ens}{2} &
        \imagepng{#1}{ens}{3} &
        \imagepng{#1}{ens}{4} &
        \imagepng{#1}{var}{0}
    }

    \begin{subfigure}{1\textwidth}
        \centering
        \begin{tabular}{ccccccccc}
        & GT & E1 & E2 & E3 & E4 & E5 & Variance & \multirow{3}{*}{%
          \begin{minipage}[c]{0.137\linewidth}
            \centering
            \includegraphics[width=\linewidth]{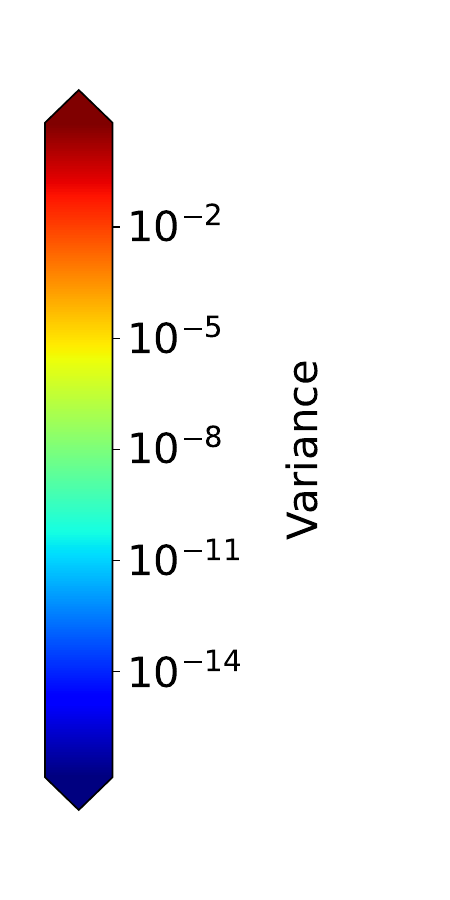}
          \end{minipage}%
        }\\
        \rotatebox{90}{\hspace{.75em} Normal} & \rowpng{low} \\
        \rotatebox{90}{\hspace{.75em} Extreme} & \rowpng{high} \\
        \multicolumn{9}{c}{\includegraphics[width=0.4\linewidth]{fig/lsm/forecast/colorbar.pdf}} \\
        \end{tabular}
    \end{subfigure}
    \caption{Qualitative results for reconstruction of one normal and one extreme weather event. Variance map uses log scale. Areas with precipitation show higher variance. Areas with extreme precipitation exhibit extreme variance. Best viewed zoomed in.}
    \label{fig:freud:qualitative_2027}
\end{figure*}

\begin{figure*}[htbp]
    \centering \small
    \setlength\tabcolsep{1pt}

    \newcommand{\imagepng}[3]{
        \includegraphics[width=0.11\linewidth]{fig/freud/ensembles/T-seed_2030/#1_precip/#2_#3.png}
    }
    
    \newcommand{\rowpng}[1]{
        \imagepng{#1}{gt}{0} &
        \imagepng{#1}{ens}{0} &
        \imagepng{#1}{ens}{1} &
        \imagepng{#1}{ens}{2} &
        \imagepng{#1}{ens}{3} &
        \imagepng{#1}{ens}{4} &
        \imagepng{#1}{var}{0}
    }

    \begin{subfigure}{1\textwidth}
        \centering
        \begin{tabular}{ccccccccc}
        & GT & E1 & E2 & E3 & E4 & E5 & Variance & \multirow{3}{*}{%
          \begin{minipage}[c]{0.137\linewidth}
            \centering
            \includegraphics[width=\linewidth]{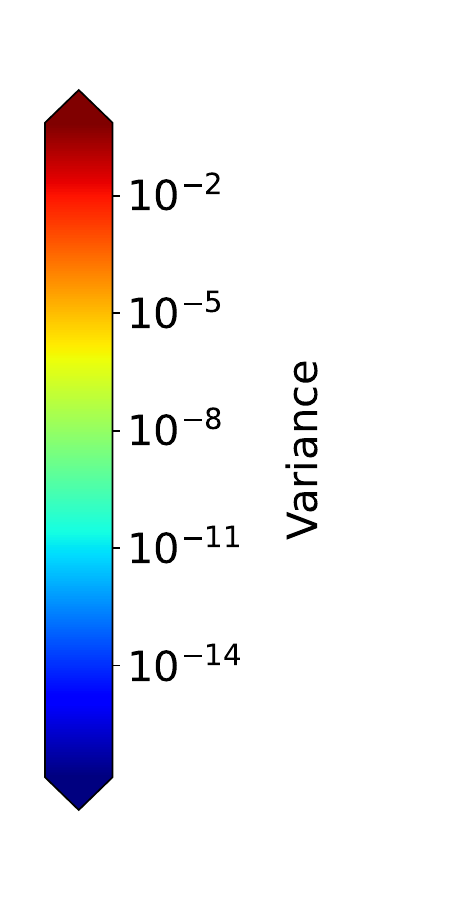}
          \end{minipage}%
        }\\
        \rotatebox{90}{\hspace{.75em} Normal} & \rowpng{low} \\
        \rotatebox{90}{\hspace{.75em} Extreme} & \rowpng{high} \\
        \multicolumn{9}{c}{\includegraphics[width=0.4\linewidth]{fig/lsm/forecast/colorbar.pdf}} \\
        \end{tabular}
    \end{subfigure}
    \caption{Qualitative results for reconstruction of one normal and one extreme weather event. Variance map uses log scale. Areas with precipitation show higher variance. Areas with extreme precipitation exhibit extreme variance. Best viewed zoomed in.}
    \label{fig:freud:qualitative_2030}
\end{figure*}

\begin{figure*}[htbp]
    \centering \small
    \setlength\tabcolsep{1pt}

    \newcommand{\imagepng}[3]{
        \includegraphics[width=0.115\linewidth]{fig/freud/ensembles/T-seed_49/#1_precip/#2_#3.png}
    }
    
    \newcommand{\rowpng}[1]{
        \imagepng{#1}{gt}{0} &
        \imagepng{#1}{ens}{0} &
        \imagepng{#1}{ens}{1} &
        \imagepng{#1}{ens}{2} &
        \imagepng{#1}{ens}{3} &
        \imagepng{#1}{ens}{4} &
        \imagepng{#1}{var}{0}
    }

    \begin{subfigure}{1\textwidth}
        \centering
        \begin{tabular}{ccccccccc}
        & GT & E1 & E2 & E3 & E4 & E5 & Variance & \multirow{3}{*}{%
          \begin{minipage}[c]{0.137\linewidth}
            \centering
            \includegraphics[width=\linewidth]{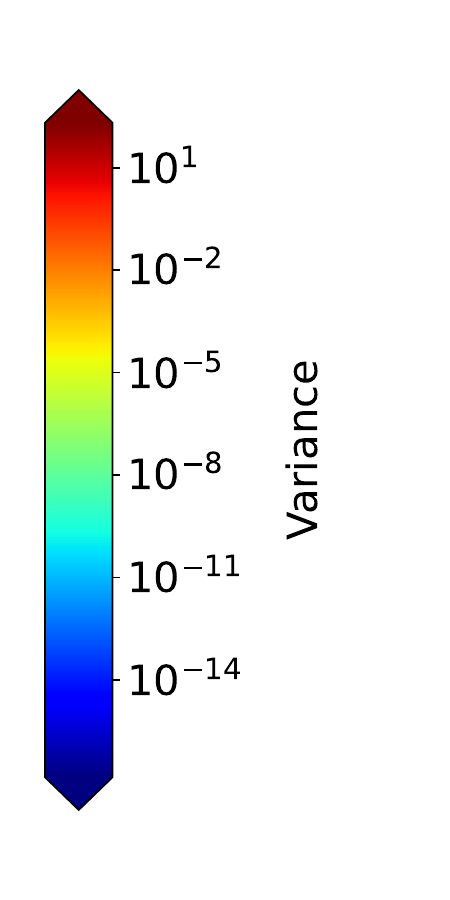}
          \end{minipage}%
        }\\
        \rotatebox{90}{\hspace{.75em} Normal} & \rowpng{low} \\
        \rotatebox{90}{\hspace{.75em} Extreme} & \rowpng{high} \\
        \multicolumn{9}{c}{\includegraphics[width=0.4\linewidth]{fig/lsm/forecast/colorbar.pdf}} \\
        \end{tabular}
    \end{subfigure}
    \caption{Qualitative results for reconstruction of one normal and one extreme weather event. Variance map uses log scale. Areas with precipitation show higher variance. Areas with extreme precipitation exhibit extreme variance. Best viewed zoomed in.}
    \label{fig:freud:qualitative_49}
\end{figure*}

\begin{figure*}[h]
    \centering \small
    \setlength\tabcolsep{1pt}

    \newcommand{\imagepng}[3]{%
        \includegraphics[width=0.23\linewidth]{fig/lsm/forecast/comparison-T-L-seed_0/#1_#2_#3.png}%
    }

    \newcommand{\rowpng}[1]{%
        \imagepng{gt}{0}{#1}
        & \imagepng{mean_our}{0}{#1}
        & \imagepng{var_our}{0}{#1}
        & \imagepng{mean_cascast}{0}{#1}
        & \imagepng{var_cascast}{0}{#1}
        & \imagepng{earthformer}{0}{#1}
    }

    \begin{minipage}{0.5\textwidth}
        \begin{subfigure}{1\textwidth}
        \centering
        \begin{tabular}{c c@{\hspace{7pt}}c@{\hspace{0pt}}c@{\hspace{7pt}}c@{\hspace{0pt}}c@{\hspace{7pt}}c}
            & GT & Ours (M) & Ours (Var) & CC (M) & CC (Var) & EF \\
            \rotatebox{90}{\hspace{.25em} $+10$\,min} &  \rowpng{1} \\
            \rotatebox{90}{\hspace{.25em} $+20$\,min} &  \rowpng{3} \\
            \rotatebox{90}{\hspace{.25em} $+30$\,min} &  \rowpng{5} \\
            \rotatebox{90}{\hspace{.25em} $+40$\,min} &  \rowpng{7} \\
            \rotatebox{90}{\hspace{.25em} $+50$\,min} &  \rowpng{9} \\
            \rotatebox{90}{\hspace{.25em} $+60$\,min} &  \rowpng{11} \\
            \multicolumn{7}{c}{\includegraphics[width=0.7\linewidth]{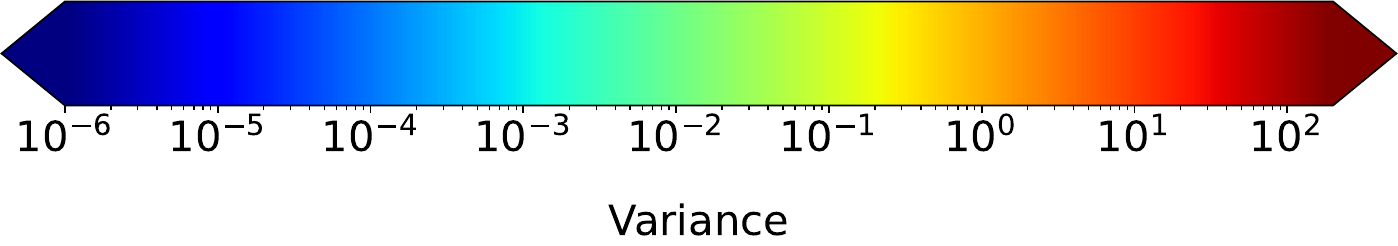}} \\
        \end{tabular}
    \end{subfigure}
    \end{minipage}
    \hspace{4.7cm}
    \begin{minipage}{0.1\textwidth}
        \centering
        \includegraphics[width=\linewidth]{fig/lsm/forecast/colorbar_vertical.pdf}
    \end{minipage}
    \caption{
        Comparison of forecasts made with our method, CasCast (CC)~\cite{gong2024cascast} and Earthformer (EF)~\cite{gao2022earthformer}. For probabilistic methods, we show the mean (M) forecast and variance (Var) of a 10-member ensemble. Our forecast aligns more closely with the ground truth and variance, while CasCast overestimates and Earthformer underestimates precipitation. Further, the ensemble variance in our method focuses on high precipitation regions in the ground truth. Best viewed zoomed in.
    }
    \label{fig:lsm:qualitative_comp_0}
\end{figure*}

\begin{figure*}[h]
    \centering \small
    \setlength\tabcolsep{1pt}

    \newcommand{\imagepng}[3]{%
        \includegraphics[width=0.22\linewidth]{fig/lsm/forecast/comparison-T-L-seed_37/#1_#2_#3.png}%
    }

    \newcommand{\rowpng}[1]{%
        \imagepng{gt}{0}{#1}
        & \imagepng{mean_our}{0}{#1}
        & \imagepng{var_our}{0}{#1}
        & \imagepng{mean_cascast}{0}{#1}
        & \imagepng{var_cascast}{0}{#1}
        & \imagepng{earthformer}{0}{#1}
    }

    \begin{minipage}{0.5\textwidth}
        \begin{subfigure}{1\textwidth}
        \centering
        \begin{tabular}{c c@{\hspace{7pt}}c@{\hspace{0pt}}c@{\hspace{7pt}}c@{\hspace{0pt}}c@{\hspace{7pt}}c}
            & GT & Ours (M) & Ours (Var) & CC (M) & CC (Var) & EF \\
            \rotatebox{90}{\hspace{.25em} $+10$\,min} &  \rowpng{1} \\
            \rotatebox{90}{\hspace{.25em} $+20$\,min} &  \rowpng{3} \\
            \rotatebox{90}{\hspace{.25em} $+30$\,min} &  \rowpng{5} \\
            \rotatebox{90}{\hspace{.25em} $+40$\,min} &  \rowpng{7} \\
            \rotatebox{90}{\hspace{.25em} $+50$\,min} &  \rowpng{9} \\
            \rotatebox{90}{\hspace{.25em} $+60$\,min} &  \rowpng{11} \\
            \multicolumn{7}{c}{\includegraphics[width=0.7\linewidth]{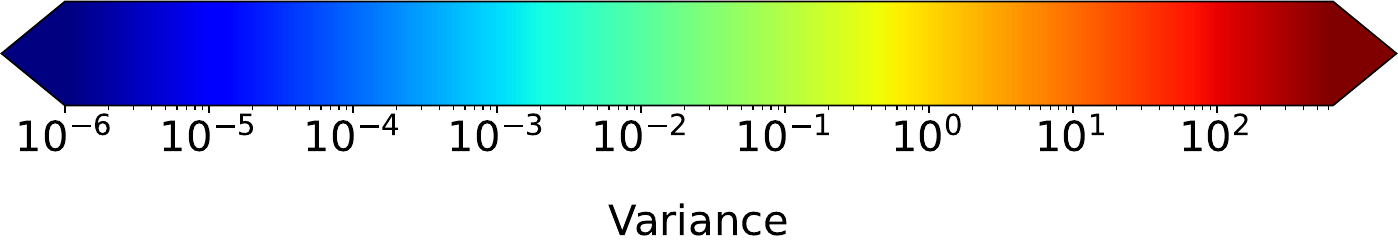}} \\
        \end{tabular}
    \end{subfigure}
    \end{minipage}
    \hspace{4.5cm}
    \begin{minipage}{0.1\textwidth}
        \centering
        \includegraphics[width=\linewidth]{fig/lsm/forecast/colorbar_vertical.pdf}
    \end{minipage}
    \caption{
        Comparison of forecasts made with our method, CasCast (CC)~\cite{gong2024cascast} and Earthformer (EF)~\cite{gao2022earthformer}. For probabilistic methods, we show the mean (M) forecast and variance (Var) of a 10-member ensemble. Our forecast shows substantially improved temporal consistency compared to CasCast, where a strong precipitation event appears for a single frame. Further, the shape of our forecast aligns more closely with the ground truth. Best viewed zoomed in.
    }
    \label{fig:lsm:qualitative_comp_37}
\end{figure*}

\begin{figure*}[h]
    \centering \small
    \setlength\tabcolsep{1pt}

    \newcommand{\imagepng}[3]{
        \includegraphics[width=0.12\linewidth]{fig/lsm/forecast/T-L-seed_42/#1_#2_#3.png}
    }

    \newcommand{\rowpng}[2]{
        \imagepng{#1}{#2}{1} & %
        \imagepng{#1}{#2}{3} & %
        \imagepng{#1}{#2}{5} & %
        \imagepng{#1}{#2}{7} & %
        \imagepng{#1}{#2}{9} & %
        \imagepng{#1}{#2}{11} %
    }

    \begin{minipage}{0.9\textwidth}
         \begin{subfigure}{1\textwidth}
            \centering
            \begin{tabular}{lcccccc}
                & $+10$\,min & $+20$\,min & $+30$\,min & $+40$\,min & $+50$\,min & $+60$\,min \\
                \rotatebox{90}{\hspace{1.8em} E1} & \rowpng{ens}{0} \\
                \rotatebox{90}{\hspace{1.8em} E2} & \rowpng{ens}{1} \\
                \rotatebox{90}{\hspace{1.8em} E3} & \rowpng{ens}{2} \\
                \arrayrulecolor{gray!50!white} \cmidrule{1-7} \arrayrulecolor{black} \\
                \rotatebox{90}{\hspace{1.75em} GT} & \rowpng{gt}{0} \\
            \end{tabular}
        \end{subfigure}
    \end{minipage}
    \hspace{-0.8cm}
    \begin{minipage}{0.09\textwidth}
        \centering
        \includegraphics[width=\linewidth]{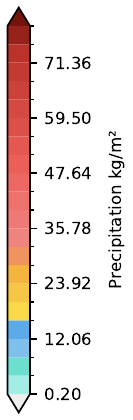}
    \end{minipage}
    \caption{Qualitative result for our L-LSM with \textit{T-reg.}\thinspace first-stage. The red rectangle highlights a punch-in for better visualization of ensemble differences in details. Best viewed zoomed in.}
    \label{fig:lsm:qualitative_42}
\end{figure*}

\begin{figure*}[h]
    \centering \small
    \setlength\tabcolsep{1pt}

    \newcommand{\imagepng}[3]{
        \includegraphics[width=0.12\linewidth]{fig/lsm/forecast/T-L-seed_37/#1_#2_#3.png}
    }

    \newcommand{\rowpng}[2]{
        \imagepng{#1}{#2}{1} & %
        \imagepng{#1}{#2}{3} & %
        \imagepng{#1}{#2}{5} & %
        \imagepng{#1}{#2}{7} & %
        \imagepng{#1}{#2}{9} & %
        \imagepng{#1}{#2}{11} %
    }

    \begin{minipage}{0.9\textwidth}
         \begin{subfigure}{1\textwidth}
            \centering
            \begin{tabular}{lcccccc}
                & $+10$\,min & $+20$\,min & $+30$\,min & $+40$\,min & $+50$\,min & $+60$\,min \\
                \rotatebox{90}{\hspace{1.8em} E1} & \rowpng{ens}{0} \\
                \rotatebox{90}{\hspace{1.8em} E2} & \rowpng{ens}{1} \\
                \rotatebox{90}{\hspace{1.8em} E3} & \rowpng{ens}{2} \\
                \arrayrulecolor{gray!50!white} \cmidrule{1-7} \arrayrulecolor{black} \\
                \rotatebox{90}{\hspace{1.75em} GT} & \rowpng{gt}{0} \\
            \end{tabular}
        \end{subfigure}
    \end{minipage}
    \hspace{-0.8cm}
    \begin{minipage}{0.09\textwidth}
        \centering
        \includegraphics[width=\linewidth]{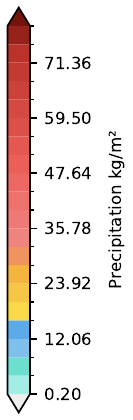}
    \end{minipage}
    \caption{Qualitative result for our L-LSM with \textit{T-reg.}\thinspace first-stage. The red rectangle highlights a punch-in for better visualization of ensemble differences in details. Best viewed zoomed in.}
    \label{fig:lsm:qualitative_37}
\end{figure*}

\begin{figure*}[h]
    \centering \small
    \setlength\tabcolsep{1pt}

    \newcommand{\imagepng}[3]{
        \includegraphics[width=0.12\linewidth]{fig/lsm/forecast/T-L-seed_2024/#1_#2_#3.png}
    }

    \newcommand{\rowpng}[2]{
        \imagepng{#1}{#2}{1} & %
        \imagepng{#1}{#2}{3} & %
        \imagepng{#1}{#2}{5} & %
        \imagepng{#1}{#2}{7} & %
        \imagepng{#1}{#2}{9} & %
        \imagepng{#1}{#2}{11} %
    }

    \begin{minipage}{0.9\textwidth}
         \begin{subfigure}{1\textwidth}
            \centering
            \begin{tabular}{lcccccc}
                & $+10$\,min & $+20$\,min & $+30$\,min & $+40$\,min & $+50$\,min & $+60$\,min \\
                \rotatebox{90}{\hspace{1.8em} E1} & \rowpng{ens}{0} \\
                \rotatebox{90}{\hspace{1.8em} E2} & \rowpng{ens}{1} \\
                \rotatebox{90}{\hspace{1.8em} E3} & \rowpng{ens}{2} \\
                \arrayrulecolor{gray!50!white} \cmidrule{1-7} \arrayrulecolor{black} \\
                \rotatebox{90}{\hspace{1.75em} GT} & \rowpng{gt}{0} \\
            \end{tabular}
        \end{subfigure}
    \end{minipage}
    \hspace{-0.8cm}
    \begin{minipage}{0.09\textwidth}
        \centering
        \includegraphics[width=\linewidth]{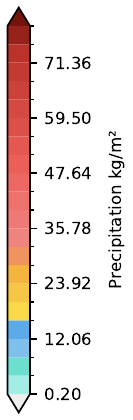}
    \end{minipage}
    \caption{Qualitative result for our L-LSM with \textit{T-reg.}\thinspace first-stage. The red rectangle highlights a punch-in for better visualization of ensemble differences in details. Best viewed zoomed in.}
    \label{fig:lsm:qualitative_2024}
\end{figure*}

\begin{figure*}[h]
    \centering \small
    \setlength\tabcolsep{1pt}

    \newcommand{\imagepng}[3]{
        \includegraphics[width=0.12\linewidth]{fig/lsm/forecast/T-L-seed_20555/#1_#2_#3.png}
    }

    \newcommand{\rowpng}[2]{
        \imagepng{#1}{#2}{1} & %
        \imagepng{#1}{#2}{3} & %
        \imagepng{#1}{#2}{5} & %
        \imagepng{#1}{#2}{7} & %
        \imagepng{#1}{#2}{9} & %
        \imagepng{#1}{#2}{11} %
    }

    \begin{minipage}{0.9\textwidth}
         \begin{subfigure}{1\textwidth}
            \centering
            \begin{tabular}{lcccccc}
                & $+10$\,min & $+20$\,min & $+30$\,min & $+40$\,min & $+50$\,min & $+60$\,min \\
                \rotatebox{90}{\hspace{1.8em} E1} & \rowpng{ens}{0} \\
                \rotatebox{90}{\hspace{1.8em} E2} & \rowpng{ens}{1} \\
                \rotatebox{90}{\hspace{1.8em} E3} & \rowpng{ens}{2} \\
                \arrayrulecolor{gray!50!white} \cmidrule{1-7} \arrayrulecolor{black} \\
                \rotatebox{90}{\hspace{1.75em} GT} & \rowpng{gt}{0} \\
            \end{tabular}
        \end{subfigure}
    \end{minipage}
    \hspace{-0.8cm}
    \begin{minipage}{0.09\textwidth}
        \centering
        \includegraphics[width=\linewidth]{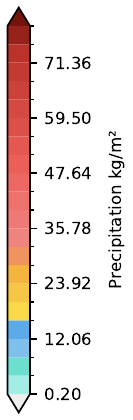}
    \end{minipage}
    \caption{Qualitative result for our L-LSM with \textit{T-reg.}\thinspace first-stage. The red rectangle highlights a punch-in for better visualization of ensemble differences in details. Best viewed zoomed in.}
    \label{fig:lsm:qualitative_20555}
\end{figure*}

\begin{figure*}[h]
    \centering \small
    \setlength\tabcolsep{1pt}

    \newcommand{\imagepng}[3]{
        \includegraphics[width=0.12\linewidth]{fig/lsm/forecast/T-L-seed_23019/#1_#2_#3.png}
    }

    \newcommand{\rowpng}[2]{
        \imagepng{#1}{#2}{1} & %
        \imagepng{#1}{#2}{3} & %
        \imagepng{#1}{#2}{5} & %
        \imagepng{#1}{#2}{7} & %
        \imagepng{#1}{#2}{9} & %
        \imagepng{#1}{#2}{11} %
    }

    \begin{minipage}{0.9\textwidth}
         \begin{subfigure}{1\textwidth}
            \centering
            \begin{tabular}{lcccccc}
                & $+10$\,min & $+20$\,min & $+30$\,min & $+40$\,min & $+50$\,min & $+60$\,min \\
                \rotatebox{90}{\hspace{1.8em} E1} & \rowpng{ens}{0} \\
                \rotatebox{90}{\hspace{1.8em} E2} & \rowpng{ens}{1} \\
                \rotatebox{90}{\hspace{1.8em} E3} & \rowpng{ens}{2} \\
                \arrayrulecolor{gray!50!white} \cmidrule{1-7} \arrayrulecolor{black} \\
                \rotatebox{90}{\hspace{1.75em} GT} & \rowpng{gt}{0} \\
            \end{tabular}
        \end{subfigure}
    \end{minipage}
    \hspace{-0.8cm}
    \begin{minipage}{0.09\textwidth}
        \centering
        \includegraphics[width=\linewidth]{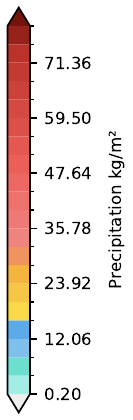}
    \end{minipage}
    \caption{Qualitative result for our L-LSM with \textit{T-reg.}\thinspace first-stage. The red rectangle highlights a punch-in for better visualization of ensemble differences in details. Best viewed zoomed in.}
    \label{fig:lsm:qualitative_23019}
\end{figure*}

\begin{figure*}[h]
    \centering \small
    \setlength\tabcolsep{1pt}

    \newcommand{\imagepng}[3]{
        \includegraphics[width=0.12\linewidth]{fig/lsm/forecast/T-L-seed_27849/#1_#2_#3.png}
    }

    \newcommand{\rowpng}[2]{
        \imagepng{#1}{#2}{1} & %
        \imagepng{#1}{#2}{3} & %
        \imagepng{#1}{#2}{5} & %
        \imagepng{#1}{#2}{7} & %
        \imagepng{#1}{#2}{9} & %
        \imagepng{#1}{#2}{11} %
    }

    \begin{minipage}{0.9\textwidth}
         \begin{subfigure}{1\textwidth}
            \centering
            \begin{tabular}{lcccccc}
                & $+10$\,min & $+20$\,min & $+30$\,min & $+40$\,min & $+50$\,min & $+60$\,min \\
                \rotatebox{90}{\hspace{1.8em} E1} & \rowpng{ens}{0} \\
                \rotatebox{90}{\hspace{1.8em} E2} & \rowpng{ens}{1} \\
                \rotatebox{90}{\hspace{1.8em} E3} & \rowpng{ens}{2} \\
                \arrayrulecolor{gray!50!white} \cmidrule{1-7} \arrayrulecolor{black} \\
                \rotatebox{90}{\hspace{1.75em} GT} & \rowpng{gt}{0} \\
            \end{tabular}
        \end{subfigure}
    \end{minipage}
    \hspace{-0.8cm}
    \begin{minipage}{0.09\textwidth}
        \centering
        \includegraphics[width=\linewidth]{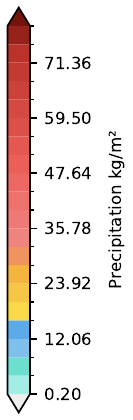}
    \end{minipage}
    \caption{Qualitative result for our L-LSM with \textit{T-reg.}\thinspace first-stage. The red rectangle highlights a punch-in for better visualization of ensemble differences in details. Best viewed zoomed in.}
    \label{fig:lsm:qualitative_27849}
\end{figure*}

\begin{figure*}[h]
    \centering \small
    \setlength\tabcolsep{1pt}

    \newcommand{\imagepng}[3]{
        \includegraphics[width=0.12\linewidth]{fig/lsm/forecast/T-L-seed_741852/#1_#2_#3.png}
    }

    \newcommand{\rowpng}[2]{
        \imagepng{#1}{#2}{1} & %
        \imagepng{#1}{#2}{3} & %
        \imagepng{#1}{#2}{5} & %
        \imagepng{#1}{#2}{7} & %
        \imagepng{#1}{#2}{9} & %
        \imagepng{#1}{#2}{11} %
    }

    \begin{minipage}{0.9\textwidth}
         \begin{subfigure}{1\textwidth}
            \centering
            \begin{tabular}{lcccccc}
                & $+10$\,min & $+20$\,min & $+30$\,min & $+40$\,min & $+50$\,min & $+60$\,min \\
                \rotatebox{90}{\hspace{1.8em} E1} & \rowpng{ens}{0} \\
                \rotatebox{90}{\hspace{1.8em} E2} & \rowpng{ens}{1} \\
                \rotatebox{90}{\hspace{1.8em} E3} & \rowpng{ens}{2} \\
                \arrayrulecolor{gray!50!white} \cmidrule{1-7} \arrayrulecolor{black} \\
                \rotatebox{90}{\hspace{1.75em} GT} & \rowpng{gt}{0} \\
            \end{tabular}
        \end{subfigure}
    \end{minipage}
    \hspace{-0.8cm}
    \begin{minipage}{0.09\textwidth}
        \centering
        \includegraphics[width=\linewidth]{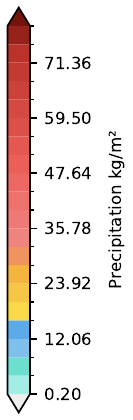}
    \end{minipage}
    \caption{Qualitative result for our L-LSM with \textit{T-reg.}\thinspace first-stage. The red rectangle highlights a punch-in for better visualization of ensemble differences in details. Best viewed zoomed in.}
    \label{fig:lsm:qualitative_741852}
\end{figure*}

\begin{figure*}[h]
    \centering \small
    \setlength\tabcolsep{1pt}

    \newcommand{\imagepng}[3]{
        \includegraphics[width=0.12\linewidth]{fig/lsm/forecast/T-L-seed_654/#1_#2_#3.png}
    }

    \newcommand{\rowpng}[2]{
        \imagepng{#1}{#2}{1} & %
        \imagepng{#1}{#2}{3} & %
        \imagepng{#1}{#2}{5} & %
        \imagepng{#1}{#2}{7} & %
        \imagepng{#1}{#2}{9} & %
        \imagepng{#1}{#2}{11} %
    }

    \begin{minipage}{0.9\textwidth}
         \begin{subfigure}{1\textwidth}
            \centering
            \begin{tabular}{lcccccc}
                & $+10$\,min & $+20$\,min & $+30$\,min & $+40$\,min & $+50$\,min & $+60$\,min \\
                \rotatebox{90}{\hspace{1.8em} E1} & \rowpng{ens}{0} \\
                \rotatebox{90}{\hspace{1.8em} E2} & \rowpng{ens}{1} \\
                \rotatebox{90}{\hspace{1.8em} E3} & \rowpng{ens}{2} \\
                \arrayrulecolor{gray!50!white} \cmidrule{1-7} \arrayrulecolor{black} \\
                \rotatebox{90}{\hspace{1.75em} GT} & \rowpng{gt}{0} \\
            \end{tabular}
        \end{subfigure}
    \end{minipage}
    \hspace{-0.8cm}
    \begin{minipage}{0.09\textwidth}
        \centering
        \includegraphics[width=\linewidth]{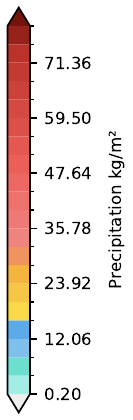}
    \end{minipage}
    \caption{Qualitative result for our L-LSM with \textit{T-reg.}\thinspace first-stage. The red rectangle highlights a punch-in for better visualization of ensemble differences in details. Best viewed zoomed in.}
    \label{fig:lsm:qualitative_654}
\end{figure*}

\end{document}